\documentclass{article} 
\usepackage{iclr2027_conference,times}

\usepackage{amsmath,amsfonts,bm}

\def\eqref#1{equation~\ref{#1}}

\def\1{\bm{1}}

\DeclareMathAlphabet{\mathsfit}{\encodingdefault}{\sfdefault}{m}{sl}
\SetMathAlphabet{\mathsfit}{bold}{\encodingdefault}{\sfdefault}{bx}{n}

\usepackage[T1]{fontenc}

\usepackage[utf8]{inputenc}

\usepackage{microtype}

\usepackage{inconsolata}

\usepackage{algorithm}
\usepackage{algpseudocode}
\usepackage{amsfonts}
\usepackage{amsmath}
\usepackage{amssymb}
\usepackage{array}
\usepackage{bm}
\usepackage{bbm}
\usepackage{booktabs}
\usepackage{colortbl}
\usepackage{dsfont}
\usepackage{enumitem}
\usepackage{graphicx}
\usepackage{hyperref}
\usepackage{latexsym}
\usepackage{mathtools}
\usepackage{multirow}
\usepackage{multicol}
\usepackage{nicefrac}
\usepackage{tabularx}
\usepackage{times}
\usepackage{pgf}
\usepackage{pifont}
\usepackage{tcolorbox}
\usepackage{tikz}
\usepackage{url}
\usepackage{wrapfig}
\usepackage{xcolor}
\usepackage{xspace}

\title{Consistency-Driven Co-Evolution for Self-Supervised Cross-Representation Learning}

\makeatletter
\newcommand{\blfootnote}[1]{%
  \begingroup
    \renewcommand{\thefootnote}{}%
    \renewcommand{\@makefntext}[1]{\noindent ##1}%
    \footnotetext{#1}%
  \endgroup
}
\makeatother

\author{%
  \makebox[\textwidth][c]{%
    Xuehang Guo$^{1}$\quad
    Pengyuan Li$^{2}$\quad
    Tom Hope$^{3}$\quad
    Tirthankar Ghosal$^{4}$\quad
    Manling Li$^{5}$\quad
    Qingyun Wang$^{1}$%
  }\\[0.45em]
  \makebox[\textwidth][c]{%
    \normalfont
    $^{1}$William \& Mary\quad
    $^{2}$IBM\quad
    $^{3}$Allen Institute for AI\quad
    $^{4}$Oak Ridge National Laboratory\quad
    $^{5}$Northwestern University%
  }%
}

\definecolor{ourscolor}{HTML}{4C7CA4}
\definecolor{oursttcolor}{HTML}{375E76}

\definecolor{mycitecolor}{HTML}{6FA9DF}

\definecolor{bboxedge}{HTML}{86A8C8}
\definecolor{bboxface}{HTML}{E7EDF4}
\definecolor{bboxtag}{HTML}{4C7CA4}

\definecolor{notunecolor}{HTML}{888888}

\definecolor{sandboxcolor}{HTML}{EF8537}
\definecolor{teachergreen}{HTML}{5CB86F}

\definecolor{BASELINE_BG}{HTML}{E6E6E6}
\definecolor{STUDENT_BG}{HTML}{E0EFF8}
\definecolor{STUDENT_AUG_BG}{HTML}{E0EFF8}
\definecolor{TEACHER_BG}{HTML}{D6F1E0}
\definecolor{TEACHER_AUG_BG}{HTML}{D6F1E0}

\definecolor{deltagreen}{HTML}{B1D3A8}
\definecolor{deltared}{HTML}{EA8C8C}
\definecolor{deltaredbg}{HTML}{F4AAAA}
\colorlet{deltacellcolor}{white} 

\definecolor{checkgreen}{HTML}{31B331}
\definecolor{crossred}{HTML}{E6716E}
\definecolor{checkgreen_bg}{HTML}{DFF6DF}
\definecolor{crossred_bg}{HTML}{FDE6E6}

\newcommand{\cmark}{\textcolor{checkgreen}{\textbf{\ding{51}}}}

\newcommand{\tcmark}{\cellcolor{checkgreen_bg}\textcolor{checkgreen}{\textbf{\ding{51}}}}
\newcommand{\txmark}{\cellcolor{crossred_bg}\textcolor{crossred}{\textbf{\ding{55}}}}
\newcommand{\tcmarkn}[1]{\cellcolor{checkgreen_bg}\textcolor{checkgreen}{\textbf{\ding{51}~(#1)}}}

\hypersetup{
    colorlinks=true,
    linkcolor=mycitecolor,
    citecolor=mycitecolor,
    urlcolor=mycitecolor
}

\newcommand{\ours}{\textcolor{ourscolor}{\textbf{CoCoEvolve}}\xspace}

\newcommand{\ourstrain}{\textcolor{oursttcolor}{\textbf{CoCoEvolve@Train}}\xspace}

\newcommand{\ourstest}{\textcolor{oursttcolor}{\textbf{CoCoEvolve@Test}}\xspace}

\newcommand{\ourseval}{\textcolor{oursttcolor}{\textbf{CoCoEvolve@Eval}}\xspace}

\tcbuselibrary{skins,breakable}

\newtcolorbox{hlbox}[1]{
  enhanced, breakable,
  colback=bboxface, colframe=bboxface, boxrule=0pt, arc=6pt,
  left=8pt, right=8pt, top=12pt, bottom=6pt,
  title=#1,
  fonttitle=\footnotesize\bfseries, coltitle=white,
  attach boxed title to top left={xshift=10pt, yshift=-\tcboxedtitleheight/2},
  boxed title style={colback=bboxtag, colframe=bboxtag, boxrule=0pt, arc=3pt,
                     left=6pt, right=6pt, top=2pt, bottom=2pt},
}

\newcommand{\circnum}[1]{%
  \begin{tikzpicture}[baseline=-0.6ex]
    \node[circle, fill=bboxedge, inner sep=0.6pt,
          minimum size=6pt, text=white,
          font=\bfseries\small] {#1};
  \end{tikzpicture}%
}

\newlist{oursenumerate}{enumerate}{1}

\setlist[oursenumerate,1]{
  label=\protect\circnum{\arabic*},
  leftmargin=1.5em,
  itemsep=0pt,
  topsep=0pt,
  partopsep=0pt,
  parsep=0pt,
}

\algrenewcommand{\algorithmiccomment}[1]{\hfill\textcolor{gray}{\(\triangleright\) #1}}

\newcommand{\applydeltacellcolor}[1]{%
  \pgfmathsetmacro{\myintensity}{int(min(100, abs(#1) * 6))}%
  \ifdim#1pt>0pt%
    \edef\tmpcolor{deltagreen!\myintensity!white}%
  \else%
    \edef\tmpcolor{deltaredbg!\myintensity!white}%
  \fi%
  \expandafter\cellcolor\expandafter{\tmpcolor}%
}

\newcommand{\myul}[1]{\leavevmode\vtop{\vbox{\hbox{#1}}\hrule height 0.4pt}}

\newcommand{\testvaldeltabgpct}[2]{%
  \pgfmathsetmacro{\deltares}{#2-#1}%
  \pgfmathtruncatemacro{\valwhole}{int(#2)}%
  \pgfmathtruncatemacro{\valdec}{int(round((#2-\valwhole)*100))}%
  \ifnum\valdec<10\edef\valdecstr{0\valdec}\else\edef\valdecstr{\valdec}\fi%
  \ifdim\deltares pt>0pt%
    \pgfmathsetmacro{\deltaint}{int(min(100, abs(\deltares) * 3))}%
    \edef\deltacolor{deltagreen!\deltaint!white}%
    \expandafter\cellcolor\expandafter{\deltacolor}\small\rmfamily\valwhole.\valdecstr%
  \else%
    \pgfmathsetmacro{\deltaint}{int(min(100, abs(\deltares) * 2))}%
    \edef\deltacolor{deltared!\deltaint!white}%
    \expandafter\cellcolor\expandafter{\deltacolor}\small\rmfamily\valwhole.\valdecstr%
  \fi%
}

\newcommand{\testvaldeltabgpctbf}[2]{%
  \pgfmathsetmacro{\deltares}{#2-#1}%
  \pgfmathtruncatemacro{\valwhole}{int(#2)}%
  \pgfmathtruncatemacro{\valdec}{int(round((#2-\valwhole)*100))}%
  \ifnum\valdec<10\edef\valdecstr{0\valdec}\else\edef\valdecstr{\valdec}\fi%
  \ifdim\deltares pt>0pt%
    \pgfmathsetmacro{\deltaint}{int(min(100, abs(\deltares) * 3))}%
    \edef\deltacolor{deltagreen!\deltaint!white}%
    \begingroup\rmfamily\bfseries\small\expandafter\cellcolor\expandafter{\deltacolor}\valwhole.\valdecstr\endgroup%
  \else%
    \pgfmathsetmacro{\deltaint}{int(min(100, abs(\deltares) * 2))}%
    \edef\deltacolor{deltared!\deltaint!white}%
    \begingroup\rmfamily\bfseries\small\expandafter\cellcolor\expandafter{\deltacolor}\valwhole.\valdecstr\endgroup%
  \fi%
}

\newcommand{\testvaldeltabgpctul}[2]{%
  \pgfmathsetmacro{\deltares}{#2-#1}%
  \pgfmathtruncatemacro{\valwhole}{int(#2)}%
  \pgfmathtruncatemacro{\valdec}{int(round((#2-\valwhole)*100))}%
  \ifnum\valdec<10\edef\valdecstr{0\valdec}\else\edef\valdecstr{\valdec}\fi%
  \ifdim\deltares pt>0pt%
    \pgfmathsetmacro{\deltaint}{int(min(100, abs(\deltares) * 3))}%
    \edef\deltacolor{deltagreen!\deltaint!white}%
    \expandafter\cellcolor\expandafter{\deltacolor}\small\rmfamily\myul{\valwhole.\valdecstr}%
  \else%
    \pgfmathsetmacro{\deltaint}{int(min(100, abs(\deltares) * 2))}%
    \edef\deltacolor{deltared!\deltaint!white}%
    \expandafter\cellcolor\expandafter{\deltacolor}\small\rmfamily\myul{\valwhole.\valdecstr}%
  \fi%
}

\newcommand{\testvaldeltabgpctbful}[2]{%
  \pgfmathsetmacro{\deltares}{#2-#1}%
  \pgfmathtruncatemacro{\valwhole}{int(#2)}%
  \pgfmathtruncatemacro{\valdec}{int(round((#2-\valwhole)*100))}%
  \ifnum\valdec<10\edef\valdecstr{0\valdec}\else\edef\valdecstr{\valdec}\fi%
  \ifdim\deltares pt>0pt%
    \pgfmathsetmacro{\deltaint}{int(min(100, abs(\deltares) * 3))}%
    \edef\deltacolor{deltagreen!\deltaint!white}%
    \begingroup\rmfamily\bfseries\small\expandafter\cellcolor\expandafter{\deltacolor}\myul{\valwhole.\valdecstr}\endgroup%
  \else%
    \pgfmathsetmacro{\deltaint}{int(min(100, abs(\deltares) * 2))}%
    \edef\deltacolor{deltared!\deltaint!white}%
    \begingroup\rmfamily\bfseries\small\expandafter\cellcolor\expandafter{\deltacolor}\myul{\valwhole.\valdecstr}\endgroup%
  \fi%
}

\newcommand{\trainvaldeltabgpct}[2]{%
  \pgfmathsetmacro{\deltares}{#2-#1}%
  \pgfmathtruncatemacro{\valwhole}{int(#2)}%
  \pgfmathtruncatemacro{\valdec}{int(round((#2-\valwhole)*100))}%
  \ifnum\valdec<10\edef\valdecstr{0\valdec}\else\edef\valdecstr{\valdec}\fi%
  \ifdim\deltares pt>0pt%
    \pgfmathsetmacro{\deltaint}{int(min(100, abs(\deltares) * 20))}%
    \edef\deltacolor{deltagreen!\deltaint!white}%
    \expandafter\cellcolor\expandafter{\deltacolor}\small\rmfamily\valwhole.\valdecstr%
  \else%
    \pgfmathsetmacro{\deltaint}{int(min(100, abs(\deltares) * 6))}%
    \edef\deltacolor{deltared!\deltaint!white}%
    \expandafter\cellcolor\expandafter{\deltacolor}\small\rmfamily\valwhole.\valdecstr%
  \fi%
}

\newcommand{\trainvaldeltabgpctbf}[2]{%
  \pgfmathsetmacro{\deltares}{#2-#1}%
  \pgfmathtruncatemacro{\valwhole}{int(#2)}%
  \pgfmathtruncatemacro{\valdec}{int(round((#2-\valwhole)*100))}%
  \ifnum\valdec<10\edef\valdecstr{0\valdec}\else\edef\valdecstr{\valdec}\fi%
  \ifdim\deltares pt>0pt%
    \pgfmathsetmacro{\deltaint}{int(min(100, abs(\deltares) * 20))}%
    \edef\deltacolor{deltagreen!\deltaint!white}%
    \begingroup\rmfamily\bfseries\small\expandafter\cellcolor\expandafter{\deltacolor}\valwhole.\valdecstr\endgroup%
  \else%
    \pgfmathsetmacro{\deltaint}{int(min(100, abs(\deltares) * 6))}%
    \edef\deltacolor{deltared!\deltaint!white}%
    \begingroup\rmfamily\bfseries\small\expandafter\cellcolor\expandafter{\deltacolor}\valwhole.\valdecstr\endgroup%
  \fi%
}

\newcommand{\trainvaldeltabgpctul}[2]{%
  \pgfmathsetmacro{\deltares}{#2-#1}%
  \pgfmathtruncatemacro{\valwhole}{int(#2)}%
  \pgfmathtruncatemacro{\valdec}{int(round((#2-\valwhole)*100))}%
  \ifnum\valdec<10\edef\valdecstr{0\valdec}\else\edef\valdecstr{\valdec}\fi%
  \ifdim\deltares pt>0pt%
    \pgfmathsetmacro{\deltaint}{int(min(100, abs(\deltares) * 20))}%
    \edef\deltacolor{deltagreen!\deltaint!white}%
    \expandafter\cellcolor\expandafter{\deltacolor}\small\rmfamily\myul{\valwhole.\valdecstr}%
  \else%
    \pgfmathsetmacro{\deltaint}{int(min(100, abs(\deltares) * 6))}%
    \edef\deltacolor{deltared!\deltaint!white}%
    \expandafter\cellcolor\expandafter{\deltacolor}\small\rmfamily\myul{\valwhole.\valdecstr}%
  \fi%
}

\newcommand{\trainvaldeltabgpctbful}[2]{%
  \pgfmathsetmacro{\deltares}{#2-#1}%
  \pgfmathtruncatemacro{\valwhole}{int(#2)}%
  \pgfmathtruncatemacro{\valdec}{int(round((#2-\valwhole)*100))}%
  \ifnum\valdec<10\edef\valdecstr{0\valdec}\else\edef\valdecstr{\valdec}\fi%
  \ifdim\deltares pt>0pt%
    \pgfmathsetmacro{\deltaint}{int(min(100, abs(\deltares) * 20))}%
    \edef\deltacolor{deltagreen!\deltaint!white}%
    \begingroup\rmfamily\bfseries\small\expandafter\cellcolor\expandafter{\deltacolor}\myul{\valwhole.\valdecstr}\endgroup%
  \else%
    \pgfmathsetmacro{\deltaint}{int(min(100, abs(\deltares) * 6))}%
    \edef\deltacolor{deltared!\deltaint!white}%
    \begingroup\rmfamily\bfseries\small\expandafter\cellcolor\expandafter{\deltacolor}\myul{\valwhole.\valdecstr}\endgroup%
  \fi%
}

\newcommand{\chartdeltabgpct}[2]{%
  \pgfmathsetmacro{\deltares}{#2-#1}%
  \pgfmathtruncatemacro{\valwhole}{int(#2)}%
  \pgfmathtruncatemacro{\valdec}{int(round((#2-\valwhole)*100))}%
  \ifnum\valdec<10\edef\valdecstr{0\valdec}\else\edef\valdecstr{\valdec}\fi%
  \ifdim\deltares pt>0pt%
    \pgfmathsetmacro{\deltaint}{int(min(100, abs(\deltares) * 1.8))}%
    \edef\deltacolor{deltagreen!\deltaint!white}%
    \expandafter\cellcolor\expandafter{\deltacolor}\small\rmfamily\valwhole.\valdecstr%
  \else%
    \pgfmathsetmacro{\deltaint}{int(min(100, abs(\deltares) * 2))}%
    \edef\deltacolor{deltared!\deltaint!white}%
    \expandafter\cellcolor\expandafter{\deltacolor}\small\rmfamily\valwhole.\valdecstr%
  \fi%
}

\newcommand{\chartdeltabgpctbf}[2]{%
  \pgfmathsetmacro{\deltares}{#2-#1}%
  \pgfmathtruncatemacro{\valwhole}{int(#2)}%
  \pgfmathtruncatemacro{\valdec}{int(round((#2-\valwhole)*100))}%
  \ifnum\valdec<10\edef\valdecstr{0\valdec}\else\edef\valdecstr{\valdec}\fi%
  \ifdim\deltares pt>0pt%
    \pgfmathsetmacro{\deltaint}{int(min(100, abs(\deltares) * 1.8))}%
    \edef\deltacolor{deltagreen!\deltaint!white}%
    \begingroup\rmfamily\bfseries\small\expandafter\cellcolor\expandafter{\deltacolor}\valwhole.\valdecstr\endgroup%
  \else%
    \pgfmathsetmacro{\deltaint}{int(min(100, abs(\deltares) * 2))}%
    \edef\deltacolor{deltared!\deltaint!white}%
    \begingroup\rmfamily\bfseries\small\expandafter\cellcolor\expandafter{\deltacolor}\valwhole.\valdecstr\endgroup%
  \fi%
}

\newcommand{\chartdeltabgpctul}[2]{%
  \pgfmathsetmacro{\deltares}{#2-#1}%
  \pgfmathtruncatemacro{\valwhole}{int(#2)}%
  \pgfmathtruncatemacro{\valdec}{int(round((#2-\valwhole)*100))}%
  \ifnum\valdec<10\edef\valdecstr{0\valdec}\else\edef\valdecstr{\valdec}\fi%
  \ifdim\deltares pt>0pt%
    \pgfmathsetmacro{\deltaint}{int(min(100, abs(\deltares) * 1.8))}%
    \edef\deltacolor{deltagreen!\deltaint!white}%
    \expandafter\cellcolor\expandafter{\deltacolor}\small\rmfamily\myul{\valwhole.\valdecstr}%
  \else%
    \pgfmathsetmacro{\deltaint}{int(min(100, abs(\deltares) * 2))}%
    \edef\deltacolor{deltared!\deltaint!white}%
    \expandafter\cellcolor\expandafter{\deltacolor}\small\rmfamily\myul{\valwhole.\valdecstr}%
  \fi%
}

\newcommand{\chartdeltabgpctbful}[2]{%
  \pgfmathsetmacro{\deltares}{#2-#1}%
  \pgfmathtruncatemacro{\valwhole}{int(#2)}%
  \pgfmathtruncatemacro{\valdec}{int(round((#2-\valwhole)*100))}%
  \ifnum\valdec<10\edef\valdecstr{0\valdec}\else\edef\valdecstr{\valdec}\fi%
  \ifdim\deltares pt>0pt%
    \pgfmathsetmacro{\deltaint}{int(min(100, abs(\deltares) * 1.8))}%
    \edef\deltacolor{deltagreen!\deltaint!white}%
    \begingroup\rmfamily\bfseries\small\expandafter\cellcolor\expandafter{\deltacolor}\myul{\valwhole.\valdecstr}\endgroup%
  \else%
    \pgfmathsetmacro{\deltaint}{int(min(100, abs(\deltares) * 2))}%
    \edef\deltacolor{deltared!\deltaint!white}%
    \begingroup\rmfamily\bfseries\small\expandafter\cellcolor\expandafter{\deltacolor}\myul{\valwhole.\valdecstr}\endgroup%
  \fi%
}

\newcommand{\charttrainvaldeltabgpct}[2]{%
  \pgfmathsetmacro{\deltares}{#2-#1}%
  \pgfmathtruncatemacro{\valwhole}{int(#2)}%
  \pgfmathtruncatemacro{\valdec}{int(round((#2-\valwhole)*100))}%
  \ifnum\valdec<10\edef\valdecstr{0\valdec}\else\edef\valdecstr{\valdec}\fi%
  \ifdim\deltares pt>0pt%
    \pgfmathsetmacro{\deltaint}{int(min(100, abs(\deltares) * 5.2))}%
    \edef\deltacolor{deltagreen!\deltaint!white}%
    \expandafter\cellcolor\expandafter{\deltacolor}\small\rmfamily\valwhole.\valdecstr%
  \else%
    \pgfmathsetmacro{\deltaint}{int(min(100, abs(\deltares) * 2))}%
    \edef\deltacolor{deltared!\deltaint!white}%
    \expandafter\cellcolor\expandafter{\deltacolor}\small\rmfamily\valwhole.\valdecstr%
  \fi%
}

\newcommand{\charttrainvaldeltabgpctbf}[2]{%
  \pgfmathsetmacro{\deltares}{#2-#1}%
  \pgfmathtruncatemacro{\valwhole}{int(#2)}%
  \pgfmathtruncatemacro{\valdec}{int(round((#2-\valwhole)*100))}%
  \ifnum\valdec<10\edef\valdecstr{0\valdec}\else\edef\valdecstr{\valdec}\fi%
  \ifdim\deltares pt>0pt%
    \pgfmathsetmacro{\deltaint}{int(min(100, abs(\deltares) * 5.2))}%
    \edef\deltacolor{deltagreen!\deltaint!white}%
    \begingroup\rmfamily\bfseries\small\expandafter\cellcolor\expandafter{\deltacolor}\valwhole.\valdecstr\endgroup%
  \else%
    \pgfmathsetmacro{\deltaint}{int(min(100, abs(\deltares) * 2))}%
    \edef\deltacolor{deltared!\deltaint!white}%
    \begingroup\rmfamily\bfseries\small\expandafter\cellcolor\expandafter{\deltacolor}\valwhole.\valdecstr\endgroup%
  \fi%
}

\newcommand{\charttrainvaldeltabgpctul}[2]{%
  \pgfmathsetmacro{\deltares}{#2-#1}%
  \pgfmathtruncatemacro{\valwhole}{int(#2)}%
  \pgfmathtruncatemacro{\valdec}{int(round((#2-\valwhole)*100))}%
  \ifnum\valdec<10\edef\valdecstr{0\valdec}\else\edef\valdecstr{\valdec}\fi%
  \ifdim\deltares pt>0pt%
    \pgfmathsetmacro{\deltaint}{int(min(100, abs(\deltares) * 5.2))}%
    \edef\deltacolor{deltagreen!\deltaint!white}%
    \expandafter\cellcolor\expandafter{\deltacolor}\small\rmfamily\myul{\valwhole.\valdecstr}%
  \else%
    \pgfmathsetmacro{\deltaint}{int(min(100, abs(\deltares) * 2))}%
    \edef\deltacolor{deltared!\deltaint!white}%
    \expandafter\cellcolor\expandafter{\deltacolor}\small\rmfamily\myul{\valwhole.\valdecstr}%
  \fi%
}

\newcommand{\charttrainvaldeltabgpctbful}[2]{%
  \pgfmathsetmacro{\deltares}{#2-#1}%
  \pgfmathtruncatemacro{\valwhole}{int(#2)}%
  \pgfmathtruncatemacro{\valdec}{int(round((#2-\valwhole)*100))}%
  \ifnum\valdec<10\edef\valdecstr{0\valdec}\else\edef\valdecstr{\valdec}\fi%
  \ifdim\deltares pt>0pt%
    \pgfmathsetmacro{\deltaint}{int(min(100, abs(\deltares) * 5.2))}%
    \edef\deltacolor{deltagreen!\deltaint!white}%
    \begingroup\rmfamily\bfseries\small\expandafter\cellcolor\expandafter{\deltacolor}\myul{\valwhole.\valdecstr}\endgroup%
  \else%
    \pgfmathsetmacro{\deltaint}{int(min(100, abs(\deltares) * 2))}%
    \edef\deltacolor{deltared!\deltaint!white}%
    \begingroup\rmfamily\bfseries\small\expandafter\cellcolor\expandafter{\deltacolor}\myul{\valwhole.\valdecstr}\endgroup%
  \fi%
}

\iclrfinalcopy 
\begin{document}

\maketitle

\blfootnote{Correspondence to: Xuehang Guo
  \texttt{\textless \url{xguo15@wm.edu}\textgreater}, Qingyun Wang
  \texttt{\textless \url{qwang16@wm.edu}\textgreater}.}

\begin{abstract}
As chart images, tabular data, and visualization code play increasingly important roles across diverse domains, cross-representation understanding across these modalities poses fundamental challenges for AI systems: the relationships across representations are inherently \textit{one-to-many}, supervision is ambiguous and costly, and model optimization lacks a principled signal that is both direction-adaptive and representation-generalizable beyond task-specific objectives. 
We introduce \ours to improve consistency across chart, table, and code representations. Instead of treating cross-representation mapping as a one-to-many problem, we define explicit one-to-one correspondences and optimize models using agreement between representations, without additional annotations. During training, \ourstrain performs co-evolution across the chart-table-code cycle, while \ourstest applies the same consistency objective at inference time for test-time co-optimization. We also present \ourseval, an evaluation suite covering all six cross-representation tasks. Across four benchmarks, \ours improves performance in both training-time and test-time settings. Our project page: \url{https://xhguo7.github.io/CoCoEvolve/}.
\end{abstract}

\section{Introduction}
\label{sec:intro}

\begin{wrapfigure}{r}{0.5\textwidth}
    \vspace{-42pt}
    \small
    \centering
    \includegraphics[width=\linewidth]{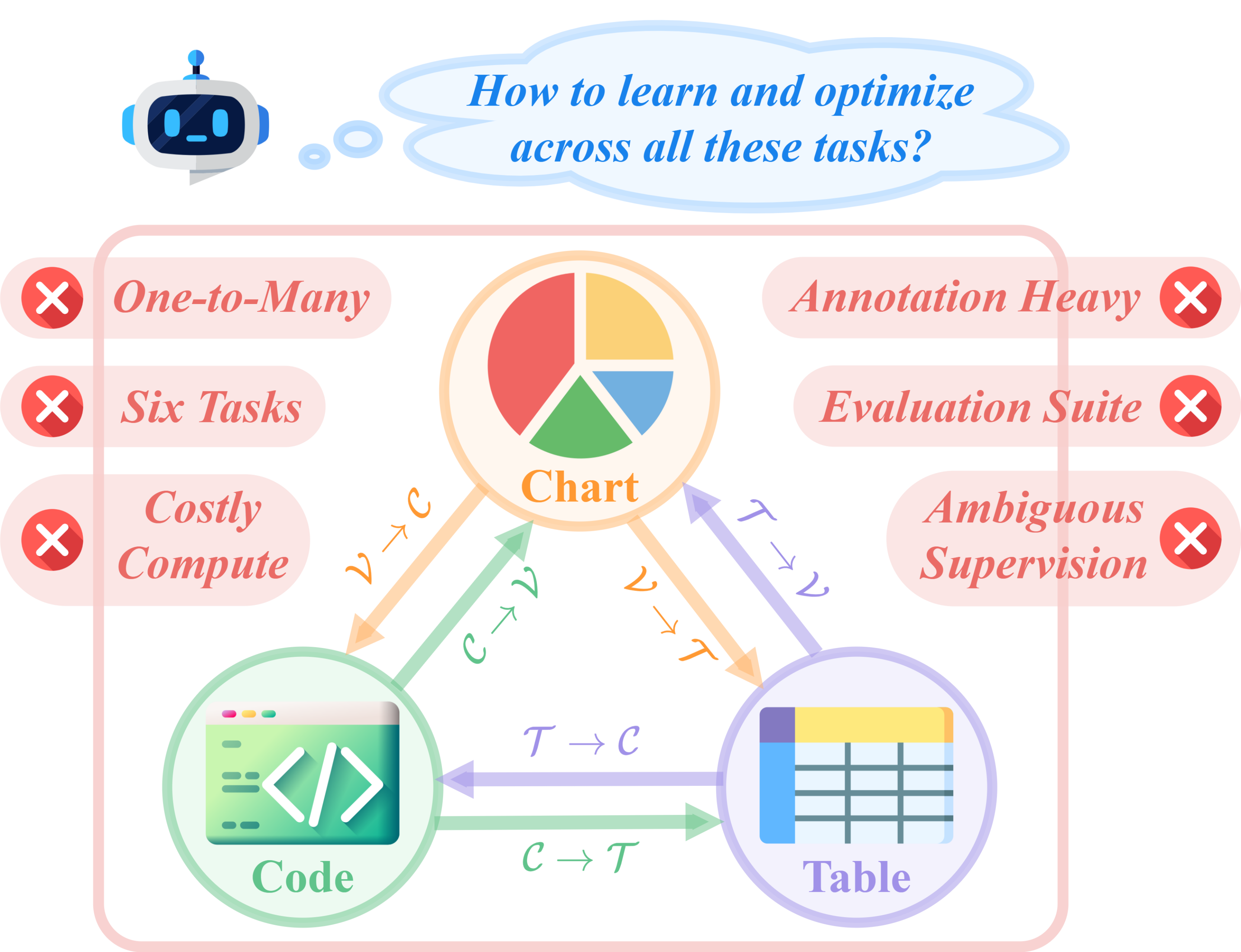}

    \vspace{-3pt}
    
    \caption{\textbf{Chart-Table-Code Representation Cycle.} Cross-representation understanding across six tasks poses great challenges for model learning and optimization due to its inherent one-to-many nature, supervision ambiguity, and high computational cost (\S\ref{sec:intro}).}
    \label{fig:teaser}

    \vspace{-12pt}

\end{wrapfigure}

Chart images are ubiquitous across scientific publications~\citep{wang2024charxiv,chartmimic2025benchmark,guo2026anagent}, financial reports~\citep{shu2025finchartbench}, data analyses~\citep{huang2024pixelsinsightssurvey}, etc. The structured information they convey can be expressed in multiple forms~\citep{chartcoder2025benchmark,chart2code2025benchmark,chartcoder2025benchmark}: a \textit{chart image} encodes data visually through axes, marks, and layouts; a \textit{data table} represents the core information in structured tabular form; and \textit{visualization code} captures the declarative logic that transforms data into visual output. Understanding and reasoning across different domains and representations poses great challenges for AI systems, as it demands not merely visual perception but precise \textit{structural cross-modal understanding}: the ability to reason about data semantics, encode logic, and establish representational correspondence across modalities.

A central bottleneck in cross-representation understanding is the \textbf{ambiguity and cost of supervision}. The relationships across chart images, tabular data, and rendering code are inherently 
\textit{one-to-many}: a single chart image may correspond to multiple valid tabular representations, and equally, multiple valid rendering programs. This means each \textit{chart-table-code} instance requires annotating a large space of potential correspondences, making labeled supervision not only expensive to produce at scale, but fundamentally ill-addressed by existing benchmarks~\citep{2026chartnet} that \textit{unconstrained one-to-one} ground truths (\S\ref{appendix:subsec:preliminary:Q1:one_to_one}). A \textbf{principled constraint definition} is needed to ground such \textit{one-to-many} mappings to precise \textit{one-to-one} correspondences (\S\ref{subsec:problem_formulation}). Chart images are abundant, yet \textit{accurately labeled cross-representation correspondences} remain scarce and poorly defined.

Compounding this, even with principled constraint definitions in place, it remains challenging to \textbf{establish a principled optimization signal} that is both direction-adaptive and representation-generalizable. A model trained on fixed chart-to-table or chart-to-code pairs can exhibit degraded performance not only on the trained tasks (\textit{chart-to-table} \& \textit{chart-to-code}), but also on tasks in the reversed direction (\textit{e.g.,} \textit{table-to-chart}) or with unseen representation combinations (\textit{e.g.,} \textit{table-to-code}) (\S\ref{appendix:subsec:preliminary:Q2:supervision_signal}). Such task-specific objectives operate locally on fixed representation pairs and cannot enforce global semantic correctness or generalize to unseen directions and representation combinations. This calls for a principled supervision paradigm: \textit{one that operates agnostically across tasks, directions, and representations, enforcing global semantic correctness without relying on fixed ground-truth labels}.

\begin{hlbox}{Key Insight}
With a principled constraint definition (\S\ref{subsec:problem_formulation}), \textit{chart}, \textit{table}, and \textit{visualization code} are placed on equal footing as \textit{multiple representations of the same underlying semantics}. If each representation is correct, they should mutually agree --- this agreement can serve as a \textit{principled optimization signal} requiring no annotation and remaining agnostic to tasks, directions, and representations (\S\ref{appendix:subsec:preliminary:Q3:failure_analysis}).

\end{hlbox}

\vspace{6pt}

Building on these insights, we introduce \ours (\S\ref{sec:method}), a \underline{co}nsistency-driven \underline{co}-\underline{evolve} framework that jointly addresses both challenges. By proposing a \textit{principled constraint definition} (\S\ref{subsec:problem_formulation}), \ours unambiguously grounds \textit{one-to-many} mappings to precise \textit{one-to-one} correspondences. Also, rather than relying on labeled correspondences and task-specific learning, \ours leverages cross-representation agreement as a \textit{principled, annotation-free optimization signal} that is agnostic to tasks, directions, and representations, training models through a \textit{representation cycle} that enforces global semantic correctness at scale.
%
%
%
To sum up, our main contributions are:

\begin{oursenumerate}
    \item We introduce a \textit{principled constraint definition} (\S\ref{subsec:problem_formulation}) that explicitly addresses the inherent \textit{one-to-many} ambiguity overlooked by existing benchmarks and methods.
    
    \item We present \ours (\S\ref{sec:method}), a co-evolution framework that turns \textit{cross-representation consistency} into a \textit{principled self-supervision signal}, enabling annotation-free optimization at both train (\S\ref{subsec:ours_at_train_time}) and test time (\S\ref{subsec:ours_at_test_time}).

    \item We propose a systematic and unified evaluation suite (\S\ref{subsec:method:evaluation_metrics}) for all tasks in the cycle (\S\ref{subsec:problem_formulation}), addressing the key limitations of existing LLM- and MLLM-based evaluation approaches.
  
    \item Experiments demonstrate that \ours effectively enhances model cross-representation understanding abilities, yielding up to $\uparrow37.91\%$ gains on the non-overlapping test set, and is generalizable to \textit{out-of-domain} settings with improvements of up to $\uparrow46.88\%$ (\S\ref{sec:experiments}).
\end{oursenumerate}

\section{Related Work}
\label{sec:related_work}

\noindent\textbf{Cross-Representation Learning.}
Recent work shows promising progress on individual edges of the \textit{chart-table-code} representation cycle (\S\ref{subsec:problem_formulation}).
\textit{Chart-to-table extraction} focuses on recovering structured tabular data from chart images~\citep{meng2024chartassistant,liu2023deplot,2026chartnet}, while \textit{chart-to-code generation} targets the visualization program underlying a chart~\citep{chartmimic2025benchmark,chart2code2025benchmark,chartcoder2025benchmark,2026chartnet}, leveraging code-capable multimodal models to reproduce the rendering logic.
However, these lines of work share a common limitation: \textit{they treat each edge as an independent supervised task, requiring costly labeled correspondences and ignoring the natural semantic redundancy across three representations}. This reveals the critical cross-representation learning gap that our work aims to bridge via co-evolution.

\noindent\textbf{Self-Supervised Cycle Learning.}
Self-supervised learning shows its strengths in annotation-free learning across various domains~\citep{chen2021selfsupervisedvisioninhistopathology,end2endcontrastiveselfsupervised2022,li20232024}.
Cycle consistency, as a self-supervision signal, is introduced in the vision domain by CycleGAN~\citep{CycleGAN2017}, revealing that unpaired cross-domain translation can be learned by enforcing round-trip reconstruction. This principle has since been extended to language~\citep{ shen2025cycleinstruct} and vision settings~\citep{adrian2024cyclecorrespondencelosslearningdense}. However, \textit{existing approaches either rely on single representations that are poorly suited for structured cross-modal reasoning or are limited to specific domains}. This motivates our consistency-driven co-evolution framework that unifies cycle consistency and co-training into a single annotation-free paradigm for multimodal cross-representation learning.

\section{Method}
\label{sec:method}

\begin{figure*}[!t]
    \vspace{0pt}

    \small
    \centering
    \includegraphics[width=1.0\textwidth]{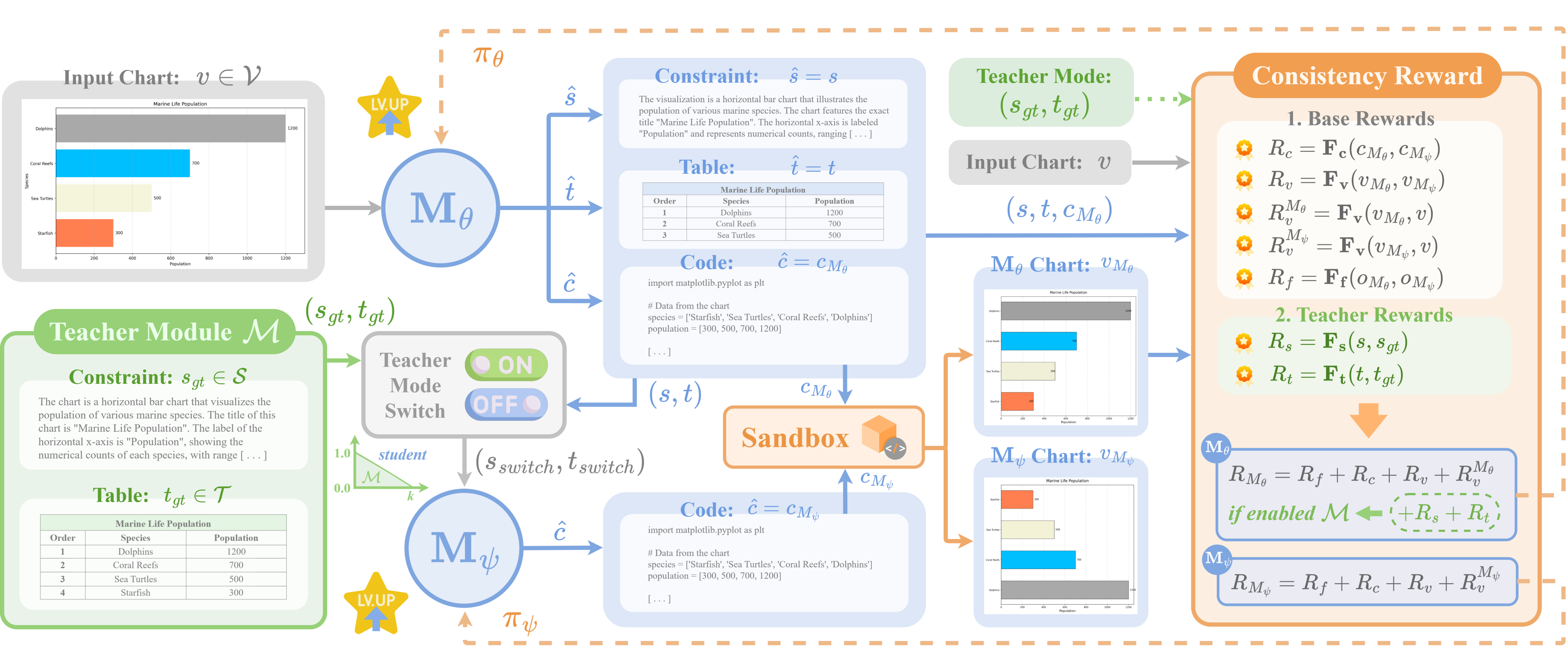}
    
    \vspace{0pt}
    
    \caption{\textbf{\ours Overview.} We introduce the consistency-driven co-evolution framework. Dashed arrows (\textcolor{orange}{- - -}) denote gradient updates at training time; models are frozen at test time.}
    \label{fig:cocoevolve_methodology}

    \vspace{6pt}

\end{figure*}

\subsection{Cross-Representation Learning}
\label{subsec:problem_formulation}

\textbf{Problem Formulation.}
Let $\mathcal{V}$, $\mathcal{T}$, and $\mathcal{C}$ denote the spaces of chart images, tabular data, and visualization code, respectively. We study the problem of cross-representation understanding as \textit{six} tasks in a \textit{representation cycle} across these three spaces: $\mathcal{V} \Leftrightarrow \mathcal{T}$, $\mathcal{T} \Leftrightarrow \mathcal{C}$, and $\mathcal{V} \Leftrightarrow \mathcal{C}$.

\noindent\textbf{Representation Cycle.}
We define three directional mappings that together constitute the 
chart-table-code cycle:
\begin{equation}
    f_{\theta} : \mathcal{V} \rightarrow \mathcal{T}, \quad
    g_{\psi} : \mathcal{T} \rightarrow \mathcal{C}, \quad
    h : \mathcal{C} \rightarrow \mathcal{V}
    \label{eq:cycle}
\end{equation}
where $f_{\theta}$ performs \textit{chart-to-table} decoding of a chart image $v \in \mathcal{V}$ into a tabular representation $t \in \mathcal{T}$, $g_{\psi}$ performs \textit{table-to-code} generation from tabular data $t \in \mathcal{T}$ to visualization sandbox $c \in \mathcal{C}$, and $h$ denotes a deterministic 
code executor that renders $c$ back into a chart image. Together, these 
form a full cycle $h \circ g_{\psi} \circ f_{\theta} : \mathcal{V} 
\rightarrow \mathcal{T} \rightarrow \mathcal{C} \rightarrow \mathcal{V}$.

\noindent\textbf{Principled Constraint Definition.}
Existing work~\citep{2026chartnet} fails to reasonably account for the inherent one-to-many nature of cross-representation mappings (\S\ref{sec:intro}).
To mitigate this ambiguity, we augment $g_{\psi}$ with auxiliary constraints $s \in \mathcal{S}$, where $\mathcal{S}$ denotes the space of descriptive constraints, and each $s_i$ conditions the alignment of $t_i$ and $c_i$ with respect to $v_i$.
\begin{equation}
    g_{\psi}(\cdot \mid s) : \mathcal{T} \rightarrow \mathcal{C}
    \label{eq:constrained_table2code}
\end{equation}
where $g_\psi(\cdot \mid s)$ represents a family of conditioned \textit{table-to-code} mappings ${g_{\psi}(\cdot \mid s) : s \in \mathcal{S}}$.



\subsection{\ourstrain: Train-Time Consistency-Driven Co-Evolution}
\label{subsec:ours_at_train_time}

By sharing a \textit{chart-table-code representation cycle} (\S\ref{subsec:problem_formulation}), \ours enables the co-evolution of two models, $M_{\theta}$ and $M_{\psi}$, jointly over unannotated chart images $\mathcal{V}=\{v\}$, where $\pi_{\theta}$ and $\pi_{\psi}$ (Eq.\ref{eq:cocoevolve_objective}) denote the respective policies parameterizing $f_{\theta}: \mathcal{V} \rightarrow \mathcal{T}$ (Eq.\ref{eq:cycle}) and $g_{\psi}: \mathcal{T} \rightarrow \mathcal{C}$ (Eq.\ref{eq:constrained_table2code}).
Leveraging hierarchical consistency-driven training signals below, \ours proposes a novel co-evolution objective (Eq.\ref{eq:cocoevolve_objective}) that serves as a drop-in training signal for existing RL algorithms, such as GRPO~\citep{shao2024grpo}, DAPO~\citep{yu2025dapo}, and GSPO~\citep{zheng2025gspo}.

\noindent\textbf{Code Consistency Reward.}
$\mathbf{F}_c$ measures semantic agreement between $\hat{c}_{\theta}$ and $\hat{c}_{\psi}$ through execution success and embedding similarity:
\begin{equation}
    \mathbf{F}_c^{(\pi)}(\hat{c}_{\theta}, \hat{c}_{\psi}) =
        \omega_e^{(\pi)} \cdot \mathds{1}[h(\hat{c}_{\theta}) \wedge
        h(\hat{c}_{\psi})]
        + \omega_s^{(\pi)} \cdot
        \texttt{sim}_{\text{c}}(\hat{c}_{\theta}, \hat{c}_{\psi})
    \label{eq:code_consistency}
\end{equation}
where $\pi \in \{\theta, \psi\}$ indexes the model, and $\omega_e^{(\pi)}, \omega_s^{(\pi)}$ are per-model sub-weights.
where $\texttt{sim}_{\text{c}}$ is a normalized cosine similarity:
\begin{equation}
    \texttt{sim}_{\text{c}}(c_a, c_b) = 
    \biggl[
        \scalebox{0.92}{$\dfrac{\cos(e_a, e_b) - \cos(e_a, e_{\textit{base}})}
              {1 - \cos(e_a, e_{\textit{base}})}$}
    \biggr]_{0}^{1}
    \label{eq:unixcoder}
\end{equation}
where $e_a$ and $e_b$ are embeddings of $c_a$ and $c_b$, $e_{\textit{base}}$ is the baseline embedding anchoring similarity above generic code patterns, and $[\cdot]_0^1$ denotes clamping to $[0,1]$.

\noindent\textbf{Visual Consistency Reward.}
$\mathbf{F}_v$ measures visual agreement between two rendered chart images through four complementary metrics, gated on code execution success:
\begin{equation}
    \mathbf{F}_v^{(\pi)}(v_a, v_b) = 
        \sum_{m} \omega_{m}^{(\pi)} \cdot F_{m}(v_a, v_b)
    \label{eq:chart_consistency}
\end{equation}
where $m \in \{\textit{clip}, \textit{ssim}, \textit{ocr}, \textit{dino}\}$, and $F_m \in \{F_{\textit{clip}}, F_{\textit{dino}}, F_{\textit{ssim}}, F_{\textit{ocr}}\}$ (Eq.\ref{eq:rule_as_judge:chart:ssim},\ref{eq:rule_as_judge:chart:clip},\ref{eq:rule_as_judge:chart:dino},\ref{eq:rule_as_judge:chart:ocr}) covers four complementary dimensions of chart evaluation (\S\ref{subsubsec:appendix:ours_eval_suite:rule_judge:chart}), respectively.

\noindent\textbf{Format Reward.}
$\mathbf{F}_f$ enforces well-formed structured outputs as a binary reward:
\begin{equation}
    \mathbf{F}_f(c, \pi) = \mathds{1}\left[\texttt{format}_{\pi}(c)\right]
    \label{eq:format}
\end{equation}
where $\texttt{format}_{\pi}(c)$ is a model-specific boolean predicate that verifies if $c$ conforms to the output format required by model $\pi \in \{\pi_{\theta}, \pi_{\psi}\}$.

\noindent\textbf{Hierarchical Reward.}
The full hierarchical reward for each model aggregates the above components:
\begin{equation}
    R_{\pi} = \lambda_{f}^{(\pi)} \mathbf{F}_f(\hat{c}_{\pi})
               + \lambda_{c}^{(\pi)} \mathbf{F}_c^{(\pi)}(\hat{c}_{\theta}, \hat{c}_{\psi})
               + \lambda_{\pi} \mathbf{F}_v^{(\pi)}(\hat{v}_{\pi}, v)
               + \lambda_{v}^{(\pi)} \mathbf{F}_v^{(\pi)}(\hat{v}_{\theta}, \hat{v}_{\psi})
    \label{eq:hierarchical_reward}
\end{equation}
where $\pi \in \{\theta, \psi\}$, and $\hat{v}_{\pi} = h(\hat{c}_{\pi})$ denotes the rendered chart image of model $\pi$ via $h$ (Eq.\ref{eq:cycle}). The shared reward terms $\mathbf{F}_c$ (Eq.\ref{eq:code_consistency}) and $\mathbf{F}_v(\hat{v}_{\theta}, \hat{v}_{\psi})$ (Eq.\ref{eq:chart_consistency}) enforce mutual agreement between two models, while $\mathbf{F}_v(\hat{v}_{\theta}, v)$ and $\mathbf{F}_v(\hat{v}_{\psi}, v)$ (Eq.\ref{eq:chart_consistency}) independently ground each model against the original $v$, preventing degenerate solutions where both models collude to produce mutually consistent but semantically incorrect outputs.

\noindent\textbf{Co-Evolve Training Objective.}
Unlike independent multi-task learning, $\pi_{\theta}$ and $\pi_{\psi}$ are coupled through both sampling and reward: $\pi_{\psi}$ conditions on $\pi_{\theta}$'s output $(\hat{s}^{(i)}, \hat{t}^{(i)})$, and each model's advantage is derived from a reward functional that depends on the other model's output. For each chart image $v \in \mathcal{V}$, $\pi_{\theta}$ draws $K_{\theta}$ rollouts $o_{\theta}^{(i)} \sim \pi_{\theta,\text{old}}(\cdot \mid v)$; and for each rollout $o_{\theta}^{(i)} = (\hat{s}^{(i)}, \hat{t}^{(i)}, \hat{c}_{\theta}^{(i)})$, $\pi_{\psi}$ draws $K_{\psi}$ conditioned rollouts $o_{\psi}^{(i,k)} \sim \pi_{\psi,\text{old}}(\cdot \mid \hat{s}^{(i)}, \hat{t}^{(i)})$. The joint objective is (\S\ref{appendix:sec:cocoevolve_at_train_details}):
\begin{equation}
\mathcal{J}(\theta, \psi) = \mathbb{E}_{v}\,
\mathbb{E}_{\substack{o_{\theta}^{(i)} \sim \pi_{\theta,\text{old}}(\cdot\mid v) \\
o_{\psi}^{(i,k)} \sim \pi_{\psi,\text{old}}(\cdot\mid \hat{s}^{(i)}, \hat{t}^{(i)})}}\!\Bigg[
    \underbrace{\frac{1}{K_{\theta}}\sum_{i=1}^{K_{\theta}}
        \mathcal{L}_{\theta}^{(i)}}_{\pi_{\theta} \text{ surrogate}}
    +
    \underbrace{\frac{1}{K_{\theta}K_{\psi}}\sum_{i=1}^{K_{\theta}}
        \sum_{k=1}^{K_{\psi}} \mathcal{L}_{\psi}^{(i,k)}}_{\pi_{\psi}
        \text{ surrogate}}
\Bigg]
\label{eq:cocoevolve_objective}
\end{equation}

\subsection{\ourstest: Test-Time Consistency-Driven Co-Optimization}
\label{subsec:ours_at_test_time}

Mirroring co-training, \ours further optimizes $M_\theta$ and $M_\psi$ at test time.
Given an input chart image $v$, \ours produces the best outputs from both models:
\begin{equation}
    c_{\psi}^{(k^*)} = \operatorname*{arg\,max}_{k \in [K_{\psi}]} 
    R_{\psi}\!\left(\hat{c}_{\psi}^{(k)} \mid 
        \operatorname*{arg\,max}_{i \in [K_{\theta}]} 
        R_{\theta}\!\left(o_{\theta}^{(i)} \mid v\right)
    \right)
    \label{eq:tto}
\end{equation}
where $o_{\theta}^{(i)} = (\hat{s}^{(i)}, \hat{t}^{(i)}, \hat{c}_{\theta}^{(i)}) \sim f_{\theta}(\cdot \mid v)$ are $K_{\theta}$ candidate rollouts from $f_{\theta}$, $c_{\psi}^{(k)} \sim g_{\psi}(\cdot \mid \hat{s}^{(i^*)}, 
\hat{t}^{(i^*)})$ are $K_{\psi}$ candidate rollouts from $g_{\psi}$ conditioned on the best $f_{\theta}$ output $o_{\theta}^{(i^*)}$. $R_{\theta}$ and $R_{\psi}$ (Eq.\ref{eq:hierarchical_reward}) serve as annotation-free signals for test-time optimization. The final outputs $(\hat{t}^{(i^*)}, \hat{c}_{\theta}^{(i^*)}, \hat{c}_{\psi}^{(k^*)})$ cover all three representations of the cycle across six tasks (\S\ref{subsec:problem_formulation}).

\subsection{Seamless Extension to Supervised Consistency-Driven Co-Learning}
\label{subsec:teacher_module}


When high-quality annotations are available, \ours is also seamlessly extensible to supervised settings at both train and test time via an optional \textbf{teacher module} $\bm{\mathcal{M}}$.

\noindent\textbf{Teacher-Guided Grounding.}
For a fraction $\tau$ of samples per batch, $\mathbf{\mathcal{M}}$ replaces $\pi_{\theta}$'s outputs $(\hat{s}^{(i)}, \hat{t}^{(i)})$ with ground-truth $(s_{\textit{gt}}, t_{\textit{gt}})$ as $\pi_{\psi}$'s conditioning context, decoupling $\pi_{\psi}$ from $\pi_{\theta}$'s potentially corrupted outputs on those samples. Formally, for each sample, the conditioning context is:
%
%
%
%
\vspace{-6pt}
\begin{equation}
    (\hat{s}^{(i)}_{\psi}, \hat{t}^{(i)}_{\psi}) = 
    \begin{cases}
        (s_{\textit{gt}}, t_{\textit{gt}}) & \quad p = \alpha \\
        (\hat{s}^{(i)}, \hat{t}^{(i)}) & \quad p = 1 - \alpha
    \end{cases}
    \label{eq:teacher_forcing}
\end{equation}
where $p$ denotes sampling probability.
In practice, we implement $\mathcal{M}$ via \textit{stratified sampling with randomized rounding} where exactly $\lfloor \alpha B \rfloor$ or $\lfloor \alpha B \rfloor + 1$ samples per batch of size $B$ are teacher-guided, ensuring the minimum-variance unbiased estimator of the target fraction $\alpha$ under a fixed batch size.

\noindent\textbf{Teacher Reward.}
When $\mathcal{M}$ is enabled, $\pi_{\theta}$ receives two additional grounding rewards:
\begin{align}
    R_{\textit{s}} &= \mathbf{F}_{\textit{s}}^{(\theta)}(\hat{s}, s_{\textit{gt}}) = \cos(\phi(\hat{s}), \phi(s_{\textit{gt}}))
    \label{eq:constraint_reward} \\
    R_{\textit{t}} &= \mathbf{F}_{\textit{t}}^{(\theta)}(\hat{t}, t_{\textit{gt}}) = \sum_{m} \omega_{m}^{(\theta)} \cdot F_{m}(\hat{t}, t_{\textit{gt}})
    \label{eq:table_reward}
\end{align}
where $m \in \{\textit{schema}, \textit{value}\}$, $\phi(\cdot)$ is a sentence embedding function via SentenceBERT~\citep{reimers2019sentenceBERT}, and $F_m \in \{F_{\textit{schema}}, F_{\textit{value}}\}$ denote column-level schema F1 and cell-level value F1 over all $(\textit{row}, \textit{column}, \textit{value})$ triples, respectively. The full $\pi_{\theta}$ reward under \textit{teacher} mode extends from Eq.\ref{eq:hierarchical_reward} to become:
\begin{equation}
    R_{\theta}^{+} = \lambda_{f}^{(\theta)} \mathbf{F}_f(\hat{c}_{\theta})
               + \lambda_{c}^{(\theta)} \mathbf{F}_c^{(\theta)}(\hat{c}_{\theta}, \hat{c}_{\psi})
               + \lambda_{\theta} \mathbf{F}_v^{(\theta)}(\hat{v}_{\theta}, v)
               + \lambda_{v}^{(\theta)} \mathbf{F}_v^{(\theta)}(\hat{v}_{\theta}, \hat{v}_{\psi})
               + \lambda_{\textit{s}}^{(\theta)} R_{\textit{s}}
               + \lambda_{\textit{t}}^{(\theta)} R_{\textit{t}}
    \label{eq:reward_mllm_teacher}
\end{equation}

\noindent\textbf{Linear Annealing.}
For robust co-evolution, \ourstrain incorporate a teacher-guidance fraction $\alpha$ linearly annealed from $\alpha_{\text{start}}$ to $\alpha_{\text{end}}$ over $N_{\textit{train}}$ training steps:
\begin{align}
    \alpha(t) = \alpha_{\textit{start}} + 
        \frac{t}{N_{\textit{train}}}\left(\alpha_{\textit{end}} - \alpha_{\textit{start}}\right)
    \label{eq:annealing}
\end{align}
where $t$ denotes current training step. This allows \ourstrain to transition gradually from teacher-guided co-training toward fully self-supervised co-evolution, as $\pi_{\theta}$ becomes increasingly reliable over the course of training.

\subsection{\ourseval{}uation Suite for Multidimensional Assessment}
\label{subsec:method:evaluation_metrics}

In addition to the lack of a systematic and generalizable evaluation suite, existing LLM-as-Judge and MLLM-as-Judge approaches for chart-to-code assessment suffer from ill-grained judgment, typically relying on LLMs and MLLMs to generate the final score by collapsing complex visual and structural fidelity into a single 0-100 score~\citep{chart2code2025benchmark,chartmimic2025benchmark}, a scale at which LLMs and MLLMs are known to be inconsistent and unreliable~\citep{wang2024largelanguagemodelsfair,stureborg2024largelanguagemodelsinconsistent}. This makes them ill-suited for the accurate, multidimensional assessment that charts and code demand. To address both limitations, we introduce \ourseval, a unified evaluation suite that, for the first time, systematically covers all six cross-representation tasks in the cycle.
Rather than collapsing assessment into wide-range scores or delegating final scoring to LLMs and MLLMs, our evaluation suite decomposes evaluation into interpretable dimensions to obtain multidimensional means. As such, \ourseval yields multidimensional measurements generalizable across benchmarks and evaluation scenarios.

\noindent\textbf{Rule-as-Judge.}
Our rule-as-jugde for chart, code, table, and constraint evaluation leverages deterministic rubric metrics, covering structural, semantic, perceptual, textual, and stylistic accuracy for charts ($F_{\textit{chart}}$, \S\ref{subsubsec:appendix:ours_eval_suite:rule_judge:chart}); executability, quality, structural, lexical, contextual, and semantic accuracy for code ($F_{\textit{code}}$, \S\ref{subsubsec:appendix:ours_eval_suite:rule_judge:code}); schema and value accuracy for tables ($F_{\textit{table}}$, \S\ref{subsubsec:appendix:ours_eval_suite:rule_judge:table}); and semantic and lexical accuracy for constraints ($F_{\textit{constraint}}$, \S\ref{subsubsec:appendix:ours_eval_suite:rule_judge:constraint}). Full metric definitions are provided in \S\ref{subsec:appendix:ours_eval_suite:rule_judge}.

\noindent\textbf{LLM-as-Judge.}
To complement rule-based metrics with semantic reasoning beyond surface-level code similarity, we introduce an LLM-as-judge evaluator that assesses visualization code quality across five fine-grained dimensions: data correctness, chart type accuracy, structural fidelity, visual accuracy, and style accuracy (\S\ref{subsec:appendix:ours_eval_suite:llm_judge}). Rather than collapsing each assessment into a wide-range score where LLMs are known to be unreliable (\S\ref{subsec:method:evaluation_metrics}), each dimension is scored independently and weighted into a final score $J_{\textit{code}}$.

\noindent\textbf{MLLM-as-Judge.}
To directly assess visual fidelity between predicted and reference outputs, we introduce MLLM-as-judge for charts, tables, and constraints. Each representation is evaluated across five fine-grained dimensions targeting distinct aspects of prediction quality, weighted into final scores $J_{\textit{chart}}$ (\S\ref{subsubsec:appendix:ours_eval_suite:mllm_judge:chart}), $J_{\textit{table}}$ (\S\ref{subsubsec:appendix:ours_eval_suite:mllm_judge:table}), and $J_{\textit{constraint}}$ (\S\ref{subsubsec:appendix:ours_eval_suite:mllm_judge:constraint}).


\begin{table*}[t!]

\vspace{-6pt}

\small
\centering
\renewcommand{\arraystretch}{1.1}
\setlength{\tabcolsep}{0pt}

\begin{tabularx}{\textwidth}{c@{\hspace{4pt}}*{13}{>{\centering\arraybackslash}X}}

\toprule

\multirow{2}{*}{\textbf{Model}} & \multicolumn{1}{c}{\textbf{Exe. (\%)}} & \multicolumn{6}{c}{\textbf{Rule-as-Judge (\%)}} & \multicolumn{6}{c}{\textbf{LLM-as-Judge (\%)}} \\
\cmidrule(lr){2-2} \cmidrule(lr){3-8} \cmidrule(lr){9-14}
& \textbf{m@4} & \textbf{bleu} & \textbf{str.} & \textbf{lex.} & \textbf{ctx.} & \textbf{sem.} & \textbf{m@4} & \textbf{type} & \textbf{data} & \textbf{str.} & \textbf{visual} & \textbf{style} & \textbf{m@4} \\

\midrule


\multicolumn{14}{c}{\cellcolor{BASELINE_BG}\textbf{\textcolor{notunecolor}{Baselines}}} \\
\addlinespace[3pt]

\textcolor{notunecolor}{\textbf{Qwen3-VL-2B}} & \textcolor{gray}{58.56} & \textcolor{gray}{29.92} & \textcolor{gray}{32.62} & \textcolor{gray}{34.39} & \textcolor{gray}{92.99} & \textcolor{gray}{76.12} & \textcolor{gray}{53.21} & \textcolor{gray}{55.37} & \textcolor{gray}{24.26} & \textcolor{gray}{30.79} & \textcolor{gray}{16.16} & \textcolor{gray}{24.03} & \textcolor{gray}{30.12} \\

\textcolor{notunecolor}{\textbf{Qwen3-VL-4B}} & \textcolor{gray}{78.47} & \textcolor{gray}{30.52} & \textcolor{gray}{33.84} & \textcolor{gray}{37.14} & \textcolor{gray}{94.02} & \textcolor{gray}{77.88} & \textcolor{gray}{54.68} & \textcolor{gray}{74.35} & \textcolor{gray}{34.03} & \textcolor{gray}{39.31} & \textcolor{gray}{23.94} & \textcolor{gray}{33.38} & \textcolor{gray}{41.00} \\

\textcolor{notunecolor}{\textbf{InternVL3.5-4B}} & \textcolor{gray}{59.72} & \textcolor{gray}{27.91} & \textcolor{gray}{35.99} & \textcolor{gray}{32.02} & \textcolor{gray}{92.29} & \textcolor{gray}{75.28} & \textcolor{gray}{52.70} & \textcolor{gray}{58.10} & \textcolor{gray}{28.33} & \textcolor{gray}{33.61} & \textcolor{gray}{18.89} & \textcolor{gray}{23.75} & \textcolor{gray}{32.54} \\

\arrayrulecolor{gray!50}\cmidrule(l){1-14}\arrayrulecolor{black}

\textcolor{notunecolor}{\textbf{Llama3.2-3B}} & \textcolor{gray}{22.51} & \textcolor{gray}{29.15} & \textcolor{gray}{10.79} & \textcolor{gray}{21.50} & \textcolor{gray}{84.32} & \textcolor{gray}{61.97} & \textcolor{gray}{41.55} & \textcolor{gray}{43.69} & \textcolor{gray}{30.42} & \textcolor{gray}{23.08} & \textcolor{gray}{13.60} & \textcolor{gray}{19.05} & \textcolor{gray}{25.97} \\

\textcolor{notunecolor}{\textbf{DeepSeek-1.3B}} & \textcolor{gray}{47.69} & \textcolor{gray}{26.89} & \textcolor{gray}{27.89} & \textcolor{gray}{26.44} & \textcolor{gray}{91.97} & \textcolor{gray}{72.05} & \textcolor{gray}{49.05} & \textcolor{gray}{45.73} & \textcolor{gray}{29.51} & \textcolor{gray}{25.08} & \textcolor{gray}{16.49} & \textcolor{gray}{16.47} & \textcolor{gray}{26.66} \\

\textcolor{notunecolor}{\textbf{DeepSeek-6.7B}} & \textcolor{gray}{68.00} & \textcolor{gray}{28.71} & \textcolor{gray}{29.81} & \textcolor{gray}{30.54} & \textcolor{gray}{92.26} & \textcolor{gray}{74.91} & \textcolor{gray}{51.25} & \textcolor{gray}{59.27} & \textcolor{gray}{34.93} & \textcolor{gray}{30.34} & \textcolor{gray}{21.05} & \textcolor{gray}{24.65} & \textcolor{gray}{34.05} \\

\textcolor{notunecolor}{\textbf{Qwen3-1.7B}} & \textcolor{gray}{57.47} & \textcolor{gray}{27.99} & \textcolor{gray}{30.03} & \textcolor{gray}{31.90} & \textcolor{gray}{93.32} & \textcolor{gray}{77.86} & \textcolor{gray}{52.22} & \textcolor{gray}{63.78} & \textcolor{gray}{33.63} & \textcolor{gray}{32.56} & \textcolor{gray}{19.26} & \textcolor{gray}{26.56} & \textcolor{gray}{35.16} \\

\textcolor{notunecolor}{\textbf{Qwen3-4B}} & \textcolor{gray}{77.60} & \textcolor{gray}{31.56} & \textcolor{gray}{26.08} & \textcolor{gray}{33.02} & \textcolor{gray}{91.94} & \textcolor{gray}{71.72} & \textcolor{gray}{50.86} & \textcolor{gray}{76.66} & \textcolor{gray}{35.75} & \textcolor{gray}{34.31} & \textcolor{gray}{23.03} & \textcolor{gray}{28.08} & \textcolor{gray}{39.57} \\

\textcolor{notunecolor}{\textbf{Qwen3-8B}} & \textcolor{gray}{82.75} & \textcolor{gray}{30.32} & \textcolor{gray}{29.89} & \textcolor{gray}{32.19} & \textcolor{gray}{93.65} & \textcolor{gray}{75.28} & \textcolor{gray}{52.27} & \textcolor{gray}{80.83} & \textcolor{gray}{37.95} & \textcolor{gray}{37.73} & \textcolor{gray}{26.34} & \textcolor{gray}{32.52} & \textcolor{gray}{43.07} \\[2pt]


\multicolumn{14}{c}{\cellcolor{STUDENT_BG}\textbf{\ours (\textcolor{oursttcolor}{@Test})}} \\
\addlinespace[3pt]

\textcolor{ourscolor}{\textbf{Qwen3-VL-2B}} & \testvaldeltabgpct{58.56}{87.04} & \testvaldeltabgpct{29.92}{31.54} & \testvaldeltabgpct{32.62}{38.36} & \testvaldeltabgpct{34.39}{37.78} & \testvaldeltabgpct{92.99}{93.22} & \testvaldeltabgpct{76.12}{77.37} & \testvaldeltabgpct{53.21}{55.65} & \testvaldeltabgpct{55.37}{76.85} & \testvaldeltabgpct{24.26}{35.56} & \testvaldeltabgpct{30.79}{41.67} & \testvaldeltabgpct{16.16}{24.44} & \testvaldeltabgpct{24.03}{35.00} & \testvaldeltabgpct{30.12}{42.70} \\

\textcolor{ourscolor}{\textbf{Qwen3-VL-4B}} & \testvaldeltabgpct{78.47}{93.52} & \testvaldeltabgpct{30.52}{33.09} & \testvaldeltabgpct{33.84}{41.21} & \testvaldeltabgpct{37.14}{41.91} & \testvaldeltabgpct{94.02}{96.27} & \testvaldeltabgpct{77.88}{81.18} & \testvaldeltabgpct{54.68}{58.73} & \testvaldeltabgpct{74.35}{86.30} & \testvaldeltabgpct{34.03}{44.26} & \testvaldeltabgpct{39.31}{48.70} & \testvaldeltabgpct{23.94}{34.81} & \testvaldeltabgpct{33.38}{44.81} & \testvaldeltabgpct{41.00}{51.78} \\

\textcolor{ourscolor}{\textbf{InternVL3.5-4B}} & \testvaldeltabgpct{59.72}{92.59} & \testvaldeltabgpct{27.91}{31.78} & \testvaldeltabgpct{35.99}{45.87} & \testvaldeltabgpct{32.02}{38.23} & \testvaldeltabgpct{92.29}{95.86} & \testvaldeltabgpct{75.28}{80.67} & \testvaldeltabgpct{52.70}{58.48} & \testvaldeltabgpct{58.10}{75.19} & \testvaldeltabgpct{28.33}{41.30} & \testvaldeltabgpct{33.61}{44.81} & \testvaldeltabgpct{18.89}{28.33} & \testvaldeltabgpct{23.75}{34.63} & \testvaldeltabgpct{32.54}{44.85} \\

\arrayrulecolor{gray!50}\cmidrule(l){1-14}\arrayrulecolor{black}

\textcolor{ourscolor}{\textbf{Llama3.2-3B}} & \testvaldeltabgpct{22.51}{85.19} & \testvaldeltabgpct{29.15}{34.87} & \testvaldeltabgpct{10.79}{29.95} & \testvaldeltabgpct{21.50}{32.06} & \testvaldeltabgpct{84.32}{94.73} & \testvaldeltabgpct{61.97}{76.70} & \testvaldeltabgpct{41.55}{53.66} & \testvaldeltabgpct{43.69}{59.26} & \testvaldeltabgpct{30.42}{48.52} & \testvaldeltabgpct{23.08}{37.78} & \testvaldeltabgpct{13.60}{28.15} & \testvaldeltabgpct{19.05}{29.07} & \testvaldeltabgpct{25.97}{40.56} \\

\textcolor{ourscolor}{\textbf{DeepSeek-1.3B}} & \testvaldeltabgpct{47.69}{96.30} & \testvaldeltabgpct{26.89}{32.23} & \testvaldeltabgpct{27.89}{42.37} & \testvaldeltabgpct{26.44}{34.08} & \testvaldeltabgpct{91.97}{96.47} & \testvaldeltabgpct{72.05}{81.01} & \testvaldeltabgpct{49.05}{57.23} & \testvaldeltabgpct{45.73}{69.44} & \testvaldeltabgpct{29.51}{46.33} & \testvaldeltabgpct{25.08}{44.63} & \testvaldeltabgpct{16.49}{34.63} & \testvaldeltabgpct{16.47}{29.81} & \testvaldeltabgpct{26.66}{44.97} \\

\textcolor{ourscolor}{\textbf{DeepSeek-6.7B}} & \testvaldeltabgpct{68.00}{99.07} & \testvaldeltabgpct{28.71}{34.23} & \testvaldeltabgpct{29.81}{45.48} & \testvaldeltabgpct{30.54}{40.81} & \testvaldeltabgpct{92.26}{97.50} & \testvaldeltabgpct{74.91}{83.18} & \testvaldeltabgpct{51.25}{60.24} & \testvaldeltabgpct{59.27}{88.33} & \testvaldeltabgpct{34.93}{48.89} & \testvaldeltabgpct{30.34}{50.19} & \testvaldeltabgpct{21.05}{41.48} & \testvaldeltabgpct{24.65}{42.96} & \testvaldeltabgpct{34.05}{54.37} \\

\textcolor{ourscolor}{\textbf{Qwen3-1.7B}} & \testvaldeltabgpct{57.47}{97.22} & \testvaldeltabgpct{27.99}{33.32} & \testvaldeltabgpct{30.03}{35.34} & \testvaldeltabgpct{31.90}{38.55} & \testvaldeltabgpct{93.32}{92.65} & \testvaldeltabgpct{77.86}{74.09} & \testvaldeltabgpct{52.22}{54.79} & \testvaldeltabgpct{63.78}{88.52} & \testvaldeltabgpct{33.63}{48.67} & \testvaldeltabgpct{32.56}{45.07} & \testvaldeltabgpct{19.26}{39.44} & \testvaldeltabgpct{26.56}{42.52} & \testvaldeltabgpct{35.16}{52.84} \\

\textcolor{ourscolor}{\textbf{Qwen3-4B}} & \testvaldeltabgpct{77.60}{99.07} & \testvaldeltabgpct{31.56}{34.94} & \testvaldeltabgpct{26.08}{36.87} & \testvaldeltabgpct{33.02}{39.50} & \testvaldeltabgpct{91.94}{95.93} & \testvaldeltabgpct{71.72}{79.64} & \testvaldeltabgpct{50.86}{57.38} & \testvaldeltabgpct{76.66}{93.33} & \testvaldeltabgpct{35.75}{50.74} & \testvaldeltabgpct{34.31}{48.89} & \testvaldeltabgpct{23.03}{41.85} & \testvaldeltabgpct{28.08}{44.26} & \testvaldeltabgpct{39.57}{55.81} \\

\textcolor{ourscolor}{\textbf{Qwen3-8B}} & \testvaldeltabgpct{82.75}{100.00} & \testvaldeltabgpct{30.32}{35.56} & \testvaldeltabgpct{29.89}{44.92} & \testvaldeltabgpct{32.19}{41.70} & \testvaldeltabgpct{93.65}{97.43} & \testvaldeltabgpct{75.28}{83.04} & \testvaldeltabgpct{52.27}{60.53} & \testvaldeltabgpct{80.83}{93.89} & \testvaldeltabgpct{37.95}{52.96} & \testvaldeltabgpct{37.73}{53.33} & \testvaldeltabgpct{26.34}{46.11} & \testvaldeltabgpct{32.52}{49.44} & \testvaldeltabgpct{43.07}{59.15} \\[2pt]


\multicolumn{14}{c}{\cellcolor{STUDENT_BG}\textbf{\ours (\textcolor{oursttcolor}{@Train})}} \\
\addlinespace[3pt]

\textcolor{ourscolor}{\textbf{Qwen3-VL-2B}} & \testvaldeltabgpct{58.56}{85.42} & \trainvaldeltabgpct{29.92}{30.84} & \trainvaldeltabgpct{32.62}{34.58} & \trainvaldeltabgpct{34.39}{36.79} & \trainvaldeltabgpct{92.99}{94.73} & \trainvaldeltabgpct{76.12}{79.37} & \trainvaldeltabgpct{53.21}{55.26} & \trainvaldeltabgpct{55.37}{57.68} & \trainvaldeltabgpct{24.26}{27.37} & \trainvaldeltabgpct{30.79}{33.24} & \trainvaldeltabgpct{16.16}{18.79} & \trainvaldeltabgpct{24.03}{26.39} & \trainvaldeltabgpct{30.12}{32.69} \\

\textcolor{ourscolor}{\textbf{Qwen3-VL-4B}} & \testvaldeltabgpct{78.47}{92.59} & \trainvaldeltabgpct{30.52}{31.36} & \trainvaldeltabgpct{33.84}{35.06} & \trainvaldeltabgpct{37.14}{39.91} & \trainvaldeltabgpct{94.02}{95.22} & \trainvaldeltabgpct{77.88}{80.92} & \trainvaldeltabgpct{54.68}{56.49} & \trainvaldeltabgpct{74.35}{75.70} & \trainvaldeltabgpct{34.03}{36.15} & \trainvaldeltabgpct{39.31}{40.80} & \trainvaldeltabgpct{23.94}{26.11} & \trainvaldeltabgpct{33.38}{35.28} & \trainvaldeltabgpct{41.00}{42.81} \\

\arrayrulecolor{gray!50}\cmidrule(l){1-14}\arrayrulecolor{black}

\textcolor{ourscolor}{\textbf{Llama3.2-3B}} & \testvaldeltabgpct{22.51}{83.10} & \testvaldeltabgpct{29.15}{29.32} & \testvaldeltabgpct{10.79}{29.56} & \testvaldeltabgpct{21.50}{32.53} & \testvaldeltabgpct{84.32}{94.30} & \testvaldeltabgpct{61.97}{78.74} & \testvaldeltabgpct{41.55}{52.89} & \trainvaldeltabgpct{43.69}{42.80} & \trainvaldeltabgpct{30.42}{32.40} & \trainvaldeltabgpct{23.08}{27.16} & \trainvaldeltabgpct{13.60}{16.60} & \trainvaldeltabgpct{19.05}{20.83} & \trainvaldeltabgpct{25.97}{27.96} \\

\textcolor{ourscolor}{\textbf{DeepSeek-1.3B}} & \testvaldeltabgpct{47.69}{93.00} & \testvaldeltabgpct{26.89}{27.75} & \testvaldeltabgpct{27.89}{34.65} & \testvaldeltabgpct{26.44}{29.39} & \testvaldeltabgpct{91.97}{92.70} & \testvaldeltabgpct{72.05}{76.77} & \testvaldeltabgpct{49.05}{52.25} & \testvaldeltabgpct{45.73}{49.99} & \testvaldeltabgpct{29.51}{32.73} & \testvaldeltabgpct{25.08}{33.43} & \testvaldeltabgpct{16.49}{21.74} & \testvaldeltabgpct{16.47}{17.24} & \testvaldeltabgpct{26.66}{31.03} \\

\textcolor{ourscolor}{\textbf{Qwen3-1.7B}} & \testvaldeltabgpct{57.47}{76.68} & \trainvaldeltabgpct{27.99}{29.02} & \trainvaldeltabgpct{30.03}{32.28} & \trainvaldeltabgpct{31.90}{34.93} & \trainvaldeltabgpct{93.32}{94.98} & \trainvaldeltabgpct{77.86}{79.38} & \trainvaldeltabgpct{52.22}{54.12} & \trainvaldeltabgpct{63.78}{64.83} & \trainvaldeltabgpct{33.63}{36.01} & \trainvaldeltabgpct{32.56}{33.54} & \trainvaldeltabgpct{19.26}{22.56} & \trainvaldeltabgpct{26.56}{25.20} & \trainvaldeltabgpct{35.16}{36.43} \\

\textcolor{ourscolor}{\textbf{Qwen3-4B}} & \testvaldeltabgpct{77.60}{95.25} & \trainvaldeltabgpct{31.56}{32.80} & \testvaldeltabgpct{26.08}{33.47} & \testvaldeltabgpct{33.02}{41.32} & \testvaldeltabgpct{91.94}{96.37} & \testvaldeltabgpct{71.72}{80.41} & \testvaldeltabgpct{50.86}{56.87} & \trainvaldeltabgpct{76.66}{77.86} & \trainvaldeltabgpct{35.75}{37.16} & \trainvaldeltabgpct{34.31}{35.72} & \trainvaldeltabgpct{23.03}{25.40} & \trainvaldeltabgpct{28.08}{30.59} & \trainvaldeltabgpct{39.57}{41.35} \\

\textcolor{ourscolor}{\textbf{Qwen3-8B}} & \testvaldeltabgpct{82.75}{98.15} & \testvaldeltabgpct{30.32}{33.89} & \testvaldeltabgpct{29.89}{36.02} & \testvaldeltabgpct{32.19}{42.19} & \testvaldeltabgpct{93.65}{97.65} & \testvaldeltabgpct{75.28}{82.28} & \testvaldeltabgpct{52.27}{58.41} & \trainvaldeltabgpct{80.83}{83.28} & \trainvaldeltabgpct{37.95}{39.24} & \trainvaldeltabgpct{37.73}{39.12} & \trainvaldeltabgpct{26.34}{28.74} & \trainvaldeltabgpct{32.52}{34.27} & \trainvaldeltabgpct{43.07}{44.93} \\[2pt]


\multicolumn{14}{c}{\cellcolor{STUDENT_AUG_BG}\textbf{\ours (\textcolor{oursttcolor}{@Train + @Test})}} \\
\addlinespace[3pt]

\textcolor{ourscolor}{\textbf{Qwen3-VL-2B}} & \testvaldeltabgpct{58.56}{93.36} & \testvaldeltabgpct{29.92}{33.90} & \testvaldeltabgpct{32.62}{40.44} & \testvaldeltabgpct{34.39}{39.14} & \testvaldeltabgpct{92.99}{94.76} & \testvaldeltabgpct{76.12}{79.42} & \testvaldeltabgpct{53.21}{57.53} & \testvaldeltabgpct{55.37}{79.32} & \testvaldeltabgpct{24.26}{37.37} & \testvaldeltabgpct{30.79}{42.83} & \testvaldeltabgpct{16.16}{25.69} & \testvaldeltabgpct{24.03}{35.88} & \testvaldeltabgpct{30.12}{44.22} \\

\textcolor{ourscolor}{\textbf{Qwen3-VL-4B}} & \testvaldeltabgpct{78.47}{97.22} & \testvaldeltabgpct{30.52}{36.31} & \testvaldeltabgpct{33.84}{41.36} & \testvaldeltabgpct{37.14}{43.87} & \testvaldeltabgpct{94.02}{96.26} & \testvaldeltabgpct{77.88}{82.32} & \testvaldeltabgpct{54.68}{60.02} & \testvaldeltabgpct{74.35}{87.96} & \testvaldeltabgpct{34.03}{46.48} & \testvaldeltabgpct{39.31}{51.11} & \testvaldeltabgpct{23.94}{38.34} & \testvaldeltabgpct{33.38}{43.33} & \testvaldeltabgpct{41.00}{53.44} \\

\arrayrulecolor{gray!50}\cmidrule(l){1-14}\arrayrulecolor{black}

\textcolor{ourscolor}{\textbf{Llama3.2-3B}} & \testvaldeltabgpct{22.51}{100.00} & \testvaldeltabgpct{29.15}{36.02} & \testvaldeltabgpct{10.79}{41.34} & \testvaldeltabgpct{21.50}{38.31} & \testvaldeltabgpct{84.32}{97.19} & \testvaldeltabgpct{61.97}{82.30} & \testvaldeltabgpct{41.55}{59.03} & \testvaldeltabgpct{43.69}{78.70} & \testvaldeltabgpct{30.42}{50.74} & \testvaldeltabgpct{23.08}{42.41} & \testvaldeltabgpct{13.60}{34.63} & \testvaldeltabgpct{19.05}{37.59} & \testvaldeltabgpct{25.97}{48.81} \\

\textcolor{ourscolor}{\textbf{DeepSeek-1.3B}} & \testvaldeltabgpct{47.69}{99.07} & \testvaldeltabgpct{26.89}{33.08} & \testvaldeltabgpct{27.89}{44.93} & \testvaldeltabgpct{26.44}{37.48} & \testvaldeltabgpct{91.97}{97.28} & \testvaldeltabgpct{72.05}{81.43} & \testvaldeltabgpct{49.05}{58.84} & \testvaldeltabgpct{45.73}{80.19} & \testvaldeltabgpct{29.51}{48.48} & \testvaldeltabgpct{25.08}{46.85} & \testvaldeltabgpct{16.49}{35.37} & \testvaldeltabgpct{16.47}{35.93} & \testvaldeltabgpct{26.66}{49.36} \\

\textcolor{ourscolor}{\textbf{Qwen3-1.7B}} & \testvaldeltabgpct{57.47}{100.00} & \testvaldeltabgpct{27.99}{34.32} & \testvaldeltabgpct{30.03}{49.47} & \testvaldeltabgpct{31.90}{43.75} & \testvaldeltabgpct{93.32}{97.72} & \testvaldeltabgpct{77.86}{84.52} & \testvaldeltabgpct{52.22}{61.96} & \testvaldeltabgpct{63.78}{88.85} & \testvaldeltabgpct{33.63}{50.74} & \testvaldeltabgpct{32.56}{49.81} & \testvaldeltabgpct{19.26}{39.63} & \testvaldeltabgpct{26.56}{42.22} & \testvaldeltabgpct{35.16}{54.25} \\

\textcolor{ourscolor}{\textbf{Qwen3-4B}} & \testvaldeltabgpct{77.60}{100.00} & \testvaldeltabgpct{31.56}{36.47} & \testvaldeltabgpct{26.08}{42.73} & \testvaldeltabgpct{33.02}{41.57} & \testvaldeltabgpct{91.94}{97.42} & \testvaldeltabgpct{71.72}{84.03} & \testvaldeltabgpct{50.86}{60.44} & \testvaldeltabgpct{76.66}{93.48} & \testvaldeltabgpct{35.75}{51.33} & \testvaldeltabgpct{34.31}{49.85} & \testvaldeltabgpct{23.03}{42.81} & \testvaldeltabgpct{28.08}{45.33} & \testvaldeltabgpct{39.57}{56.56} \\

\textcolor{ourscolor}{\textbf{Qwen3-8B}} & \testvaldeltabgpct{82.75}{100.00} & \testvaldeltabgpct{30.32}{39.65} & \testvaldeltabgpct{29.89}{50.37} & \testvaldeltabgpct{32.19}{43.68} & \testvaldeltabgpct{93.65}{98.88} & \testvaldeltabgpct{75.28}{88.23} & \testvaldeltabgpct{52.27}{64.16} & \testvaldeltabgpct{80.83}{94.62} & \testvaldeltabgpct{37.95}{54.77} & \testvaldeltabgpct{37.73}{55.38} & \testvaldeltabgpct{26.34}{47.52} & \testvaldeltabgpct{32.52}{50.79} & \testvaldeltabgpct{43.07}{60.62} \\

\bottomrule
\end{tabularx}

\caption{\textbf{Performance Evaluation on Code.} We evaluate $\mathcal{X} \rightarrow \mathcal{C}$ ($\mathcal{X} \in \{\mathcal{V},\mathcal{T}\}$) performance for $M_\theta$ ($\mathcal{V} \rightarrow \mathcal{C}$) and $M_\psi$ ($\mathcal{T} \rightarrow \mathcal{C}$) using \textit{rule-as-judge} and \textit{LLM-as-judge}.
We calculate the mean score over four rollouts for both $M_\theta$ and $M_\psi$, denoted as \textit{m@4}. Compared to \textcolor{notunecolor}{\textit{baselines}}, \ours-enhanced models are optimized at train time and/or test time in \textit{student} mode.}
\label{tab:exp:exp_main_code_evaluation}

\end{table*}


\begin{table}[t!]

\vspace{-6pt}

\small
\centering
\renewcommand{\arraystretch}{1.1}
\setlength{\tabcolsep}{0pt}

\begin{tabularx}{\textwidth}{c*{11}{>{\centering\arraybackslash}X}}

\toprule

\multirow{2}{*}{\textbf{Model}} & \multicolumn{5}{c}{\textbf{Rule-as-Judge (\%)}} & \multicolumn{6}{c}{\textbf{MLLM-as-Judge (\%)}} \\
\cmidrule(lr){2-6} \cmidrule(lr){7-12}
& \textbf{str.} & \textbf{sem.} & \textbf{sim.} & \textbf{acc.} & \textbf{M@4} & \textbf{type} & \textbf{data} & \textbf{text} & \textbf{style} & \textbf{visual} & \textbf{M@4} \\

\midrule


\multicolumn{12}{c}{\cellcolor{BASELINE_BG}\textbf{\textcolor{notunecolor}{Baselines}}} \\
\addlinespace[3pt]

\textcolor{notunecolor}{\textbf{Qwen3-VL-2B}} & \textcolor{gray}{38.06} & \textcolor{gray}{55.59} & \textcolor{gray}{54.37} & \textcolor{gray}{36.47} & \textcolor{gray}{46.12} & \textcolor{gray}{36.39} & \textcolor{gray}{20.56} & \textcolor{gray}{27.92} & \textcolor{gray}{18.06} & \textcolor{gray}{16.67} & \textcolor{gray}{23.92} \\

\textcolor{notunecolor}{\textbf{Qwen3-VL-4B}} & \textcolor{gray}{51.53} & \textcolor{gray}{75.22} & \textcolor{gray}{74.48} & \textcolor{gray}{54.23} & \textcolor{gray}{63.87} & \textcolor{gray}{55.93} & \textcolor{gray}{33.84} & \textcolor{gray}{44.81} & \textcolor{gray}{30.42} & \textcolor{gray}{27.92} & \textcolor{gray}{38.58} \\

\textcolor{notunecolor}{\textbf{InternVL3.5-4B}} & \textcolor{gray}{39.10} & \textcolor{gray}{56.35} & \textcolor{gray}{54.37} & \textcolor{gray}{37.68} & \textcolor{gray}{46.88} & \textcolor{gray}{35.97} & \textcolor{gray}{23.94} & \textcolor{gray}{29.35} & \textcolor{gray}{15.42} & \textcolor{gray}{16.25} & \textcolor{gray}{24.19} \\

\arrayrulecolor{gray!50}\cmidrule(l){1-12}\arrayrulecolor{black}

\textcolor{notunecolor}{\textbf{Llama3.2-3B}} & \textcolor{gray}{14.71} & \textcolor{gray}{20.95} & \textcolor{gray}{20.31} & \textcolor{gray}{12.29} & \textcolor{gray}{17.07} & \textcolor{gray}{12.31} & \textcolor{gray}{10.15} & \textcolor{gray}{8.54} & \textcolor{gray}{6.32} & \textcolor{gray}{5.47} & \textcolor{gray}{8.56} \\

\textcolor{notunecolor}{\textbf{DeepSeek-1.3B}} & \textcolor{gray}{30.75} & \textcolor{gray}{44.62} & \textcolor{gray}{42.96} & \textcolor{gray}{26.83} & \textcolor{gray}{36.29} & \textcolor{gray}{25.03} & \textcolor{gray}{18.32} & \textcolor{gray}{17.97} & \textcolor{gray}{9.47} & \textcolor{gray}{10.46} & \textcolor{gray}{16.25} \\

\textcolor{notunecolor}{\textbf{DeepSeek-6.7B}} & \textcolor{gray}{44.53} & \textcolor{gray}{63.95} & \textcolor{gray}{62.29} & \textcolor{gray}{42.75} & \textcolor{gray}{53.38} & \textcolor{gray}{43.06} & \textcolor{gray}{31.56} & \textcolor{gray}{32.18} & \textcolor{gray}{19.04} & \textcolor{gray}{20.58} & \textcolor{gray}{29.28} \\

\textcolor{notunecolor}{\textbf{Qwen3-1.7B}} & \textcolor{gray}{36.69} & \textcolor{gray}{54.28} & \textcolor{gray}{53.14} & \textcolor{gray}{35.22} & \textcolor{gray}{44.83} & \textcolor{gray}{37.77} & \textcolor{gray}{23.04} & \textcolor{gray}{26.19} & \textcolor{gray}{16.88} & \textcolor{gray}{17.09} & \textcolor{gray}{24.19} \\

\textcolor{notunecolor}{\textbf{Qwen3-4B}} & \textcolor{gray}{51.36} & \textcolor{gray}{73.26} & \textcolor{gray}{72.27} & \textcolor{gray}{50.28} & \textcolor{gray}{61.79} & \textcolor{gray}{54.70} & \textcolor{gray}{34.66} & \textcolor{gray}{35.96} & \textcolor{gray}{23.61} & \textcolor{gray}{25.50} & \textcolor{gray}{34.89} \\

\textcolor{notunecolor}{\textbf{Qwen3-8B}} & \textcolor{gray}{54.60} & \textcolor{gray}{78.65} & \textcolor{gray}{77.66} & \textcolor{gray}{55.72} & \textcolor{gray}{66.66} & \textcolor{gray}{61.32} & \textcolor{gray}{38.38} & \textcolor{gray}{43.68} & \textcolor{gray}{27.84} & \textcolor{gray}{29.11} & \textcolor{gray}{40.07} \\[2pt]


\multicolumn{12}{c}{\cellcolor{STUDENT_BG}\textbf{\ours (\textcolor{oursttcolor}{@Test})}} \\
\addlinespace[3pt]

\textcolor{ourscolor}{\textbf{Qwen3-VL-2B}} & \chartdeltabgpct{38.06}{58.76} & \chartdeltabgpct{55.59}{83.36} & \chartdeltabgpct{54.37}{81.80} & \chartdeltabgpct{36.47}{58.58} & \chartdeltabgpct{46.12}{70.63} & \chartdeltabgpct{36.39}{66.85} & \chartdeltabgpct{20.56}{37.04} & \chartdeltabgpct{27.92}{52.96} & \chartdeltabgpct{18.06}{37.04} & \chartdeltabgpct{16.67}{32.59} & \chartdeltabgpct{23.92}{45.30} \\

\textcolor{ourscolor}{\textbf{Qwen3-VL-4B}} & \chartdeltabgpct{51.53}{63.11} & \chartdeltabgpct{75.22}{90.27} & \chartdeltabgpct{74.48}{89.64} & \chartdeltabgpct{54.23}{68.67} & \chartdeltabgpct{63.87}{77.92} & \chartdeltabgpct{55.93}{85.56} & \chartdeltabgpct{33.84}{56.11} & \chartdeltabgpct{44.81}{71.48} & \chartdeltabgpct{30.42}{54.44} & \chartdeltabgpct{27.92}{47.22} & \chartdeltabgpct{38.58}{62.96} \\

\textcolor{ourscolor}{\textbf{InternVL3.5-4B}} & \chartdeltabgpct{39.10}{62.84} & \chartdeltabgpct{56.35}{88.01} & \chartdeltabgpct{54.37}{85.51} & \chartdeltabgpct{37.68}{61.89} & \chartdeltabgpct{46.88}{74.56} & \chartdeltabgpct{35.97}{68.15} & \chartdeltabgpct{23.94}{47.59} & \chartdeltabgpct{29.35}{58.70} & \chartdeltabgpct{15.42}{33.52} & \chartdeltabgpct{16.25}{32.59} & \chartdeltabgpct{24.19}{48.11} \\

\arrayrulecolor{gray!50}\cmidrule(l){1-12}\arrayrulecolor{black}

\textcolor{ourscolor}{\textbf{Llama3.2-3B}} & \chartdeltabgpct{14.71}{58.79} & \chartdeltabgpct{20.95}{80.66} & \chartdeltabgpct{20.31}{78.98} & \chartdeltabgpct{12.29}{56.39} & \chartdeltabgpct{17.07}{68.71} & \chartdeltabgpct{12.31}{58.52} & \chartdeltabgpct{10.15}{45.74} & \chartdeltabgpct{8.54}{46.48} & \chartdeltabgpct{6.32}{34.63} & \chartdeltabgpct{5.47}{28.15} & \chartdeltabgpct{8.56}{42.70} \\

\textcolor{ourscolor}{\textbf{DeepSeek-1.3B}} & \chartdeltabgpct{30.75}{67.22} & \chartdeltabgpct{44.62}{91.95} & \chartdeltabgpct{42.96}{89.94} & \chartdeltabgpct{26.83}{63.97} & \chartdeltabgpct{36.29}{78.27} & \chartdeltabgpct{25.03}{76.48} & \chartdeltabgpct{18.32}{55.93} & \chartdeltabgpct{17.97}{57.78} & \chartdeltabgpct{9.47}{34.27} & \chartdeltabgpct{10.46}{34.44} & \chartdeltabgpct{16.25}{51.78} \\

\textcolor{ourscolor}{\textbf{DeepSeek-6.7B}} & \chartdeltabgpct{44.53}{70.26} & \chartdeltabgpct{63.95}{95.36} & \chartdeltabgpct{62.29}{93.71} & \chartdeltabgpct{42.75}{73.60} & \chartdeltabgpct{53.38}{83.23} & \chartdeltabgpct{43.06}{88.52} & \chartdeltabgpct{31.56}{68.15} & \chartdeltabgpct{32.18}{74.63} & \chartdeltabgpct{19.04}{49.81} & \chartdeltabgpct{20.58}{46.48} & \chartdeltabgpct{29.28}{65.52} \\

\textcolor{ourscolor}{\textbf{Qwen3-1.7B}} & \chartdeltabgpct{36.69}{65.60} & \chartdeltabgpct{54.28}{90.47} & \chartdeltabgpct{53.14}{92.31} & \chartdeltabgpct{35.22}{69.89} & \chartdeltabgpct{44.83}{79.57} & \chartdeltabgpct{37.77}{87.96} & \chartdeltabgpct{23.04}{64.44} & \chartdeltabgpct{26.19}{70.74} & \chartdeltabgpct{16.88}{48.89} & \chartdeltabgpct{17.09}{45.93} & \chartdeltabgpct{24.19}{63.59} \\

\textcolor{ourscolor}{\textbf{Qwen3-4B}} & \chartdeltabgpct{51.36}{68.94} & \chartdeltabgpct{73.26}{95.47} & \chartdeltabgpct{72.27}{94.89} & \chartdeltabgpct{50.28}{74.48} & \chartdeltabgpct{61.79}{83.45} & \chartdeltabgpct{54.70}{95.74} & \chartdeltabgpct{34.66}{75.74} & \chartdeltabgpct{35.96}{74.44} & \chartdeltabgpct{23.61}{55.74} & \chartdeltabgpct{25.50}{53.52} & \chartdeltabgpct{34.89}{71.04} \\

\textcolor{ourscolor}{\textbf{Qwen3-8B}} & \chartdeltabgpct{54.60}{71.22} & \chartdeltabgpct{78.65}{96.80} & \chartdeltabgpct{77.66}{96.51} & \chartdeltabgpct{55.72}{77.38} & \chartdeltabgpct{66.66}{85.48} & \chartdeltabgpct{61.32}{97.04} & \chartdeltabgpct{38.38}{77.41} & \chartdeltabgpct{43.68}{85.37} & \chartdeltabgpct{27.84}{58.52} & \chartdeltabgpct{29.11}{55.93} & \chartdeltabgpct{40.07}{74.85} \\[2pt]


\multicolumn{12}{c}{\cellcolor{STUDENT_BG}\textbf{\ours (\textcolor{oursttcolor}{@Train})}} \\
\addlinespace[3pt]

\textcolor{ourscolor}{\textbf{Qwen3-VL-2B}} & \charttrainvaldeltabgpct{38.06}{46.80} & \charttrainvaldeltabgpct{55.59}{60.89} & \charttrainvaldeltabgpct{54.37}{61.39} & \charttrainvaldeltabgpct{36.47}{42.83} & \charttrainvaldeltabgpct{46.12}{52.98} & \charttrainvaldeltabgpct{36.39}{40.03} & \charttrainvaldeltabgpct{20.56}{26.78} & \charttrainvaldeltabgpct{27.92}{29.18} & \charttrainvaldeltabgpct{18.06}{21.32} & \charttrainvaldeltabgpct{16.67}{20.16} & \charttrainvaldeltabgpct{23.92}{27.49} \\

\textcolor{ourscolor}{\textbf{Qwen3-VL-4B}} & \charttrainvaldeltabgpct{51.53}{59.05} & \charttrainvaldeltabgpct{75.22}{82.02} & \charttrainvaldeltabgpct{74.48}{81.24} & \charttrainvaldeltabgpct{54.23}{60.32} & \charttrainvaldeltabgpct{63.87}{70.66} & \charttrainvaldeltabgpct{55.93}{59.40} & \charttrainvaldeltabgpct{33.84}{39.81} & \charttrainvaldeltabgpct{44.81}{45.63} & \charttrainvaldeltabgpct{30.42}{33.33} & \charttrainvaldeltabgpct{27.92}{31.39} & \charttrainvaldeltabgpct{38.58}{41.91} \\

\arrayrulecolor{gray!50}\cmidrule(l){1-12}\arrayrulecolor{black}

\textcolor{ourscolor}{\textbf{Llama3.2-3B}} & \chartdeltabgpct{14.71}{56.04} & \chartdeltabgpct{20.95}{78.43} & \chartdeltabgpct{20.31}{74.12} & \chartdeltabgpct{12.29}{49.57} & \chartdeltabgpct{17.07}{64.54} & \chartdeltabgpct{12.31}{38.11} & \chartdeltabgpct{10.15}{27.86} & \chartdeltabgpct{8.54}{31.67} & \chartdeltabgpct{6.32}{13.01} & \chartdeltabgpct{5.47}{17.19} & \chartdeltabgpct{8.56}{25.57} \\

\textcolor{ourscolor}{\textbf{DeepSeek-1.3B}} & \chartdeltabgpct{30.75}{63.12} & \chartdeltabgpct{44.62}{88.03} & \chartdeltabgpct{42.96}{84.29} & \chartdeltabgpct{26.83}{58.92} & \chartdeltabgpct{36.29}{73.59} & \chartdeltabgpct{25.03}{50.68} & \chartdeltabgpct{18.32}{32.13} & \chartdeltabgpct{17.97}{43.00} & \chartdeltabgpct{9.47}{13.77} & \chartdeltabgpct{10.46}{22.80} & \chartdeltabgpct{16.25}{32.48} \\

\textcolor{ourscolor}{\textbf{Qwen3-1.7B}} & \charttrainvaldeltabgpct{36.69}{49.85} & \charttrainvaldeltabgpct{54.28}{72.59} & \charttrainvaldeltabgpct{53.14}{70.79} & \charttrainvaldeltabgpct{35.22}{48.05} & \charttrainvaldeltabgpct{44.83}{60.32} & \charttrainvaldeltabgpct{37.77}{49.55} & \charttrainvaldeltabgpct{23.04}{33.54} & \charttrainvaldeltabgpct{26.19}{37.25} & \charttrainvaldeltabgpct{16.88}{21.34} & \charttrainvaldeltabgpct{17.09}{23.28} & \charttrainvaldeltabgpct{24.19}{32.99} \\

\textcolor{ourscolor}{\textbf{Qwen3-4B}} & \charttrainvaldeltabgpct{51.36}{60.88} & \charttrainvaldeltabgpct{73.26}{84.82} & \charttrainvaldeltabgpct{72.27}{83.45} & \charttrainvaldeltabgpct{50.28}{60.36} & \charttrainvaldeltabgpct{61.79}{72.38} & \charttrainvaldeltabgpct{54.70}{64.66} & \charttrainvaldeltabgpct{34.66}{38.94} & \charttrainvaldeltabgpct{35.96}{46.86} & \charttrainvaldeltabgpct{23.61}{25.99} & \charttrainvaldeltabgpct{25.50}{30.89} & \charttrainvaldeltabgpct{34.89}{41.47} \\

\textcolor{ourscolor}{\textbf{Qwen3-8B}} & \charttrainvaldeltabgpct{54.60}{61.83} & \charttrainvaldeltabgpct{78.65}{86.21} & \charttrainvaldeltabgpct{77.66}{85.34} & \charttrainvaldeltabgpct{55.72}{62.89} & \charttrainvaldeltabgpct{66.66}{74.07} & \charttrainvaldeltabgpct{61.32}{68.93} & \charttrainvaldeltabgpct{38.38}{42.37} & \charttrainvaldeltabgpct{43.68}{48.97} & \charttrainvaldeltabgpct{27.84}{30.96} & \charttrainvaldeltabgpct{29.11}{32.78} & \charttrainvaldeltabgpct{40.07}{44.80} \\[2pt]


\multicolumn{12}{c}{\cellcolor{STUDENT_AUG_BG}\textbf{\ours (\textcolor{oursttcolor}{@Train + @Test})}} \\
\addlinespace[3pt]

\textcolor{ourscolor}{\textbf{Qwen3-VL-2B}} & \chartdeltabgpct{38.06}{61.32} & \chartdeltabgpct{55.59}{89.36} & \chartdeltabgpct{54.37}{88.38} & \chartdeltabgpct{36.47}{63.37} & \chartdeltabgpct{46.12}{75.61} & \chartdeltabgpct{36.39}{72.37} & \chartdeltabgpct{20.56}{52.89} & \chartdeltabgpct{27.92}{60.38} & \chartdeltabgpct{18.06}{45.83} & \chartdeltabgpct{16.67}{40.76} & \chartdeltabgpct{23.92}{54.45} \\

\textcolor{ourscolor}{\textbf{Qwen3-VL-4B}} & \chartdeltabgpct{51.53}{67.52} & \chartdeltabgpct{75.22}{93.86} & \chartdeltabgpct{74.48}{93.07} & \chartdeltabgpct{54.23}{70.39} & \chartdeltabgpct{63.87}{81.21} & \chartdeltabgpct{55.93}{88.15} & \chartdeltabgpct{33.84}{61.67} & \chartdeltabgpct{44.81}{71.30} & \chartdeltabgpct{30.42}{55.37} & \chartdeltabgpct{27.92}{49.44} & \chartdeltabgpct{38.58}{65.19} \\

\arrayrulecolor{gray!50}\cmidrule(l){1-12}\arrayrulecolor{black}

\textcolor{ourscolor}{\textbf{Llama3.2-3B}} & \chartdeltabgpct{14.71}{69.86} & \chartdeltabgpct{20.95}{95.55} & \chartdeltabgpct{20.31}{91.90} & \chartdeltabgpct{12.29}{69.07} & \chartdeltabgpct{17.07}{81.60} & \chartdeltabgpct{12.31}{74.26} & \chartdeltabgpct{10.15}{59.81} & \chartdeltabgpct{8.54}{60.00} & \chartdeltabgpct{6.32}{32.78} & \chartdeltabgpct{5.47}{35.19} & \chartdeltabgpct{8.56}{52.41} \\

\textcolor{ourscolor}{\textbf{DeepSeek-1.3B}} & \chartdeltabgpct{30.75}{69.93} & \chartdeltabgpct{44.62}{94.84} & \chartdeltabgpct{42.96}{92.59} & \chartdeltabgpct{26.83}{71.61} & \chartdeltabgpct{36.29}{82.24} & \chartdeltabgpct{25.03}{82.22} & \chartdeltabgpct{18.32}{61.30} & \chartdeltabgpct{17.97}{73.71} & \chartdeltabgpct{9.47}{35.26} & \chartdeltabgpct{10.46}{41.67} & \chartdeltabgpct{16.25}{58.83} \\

\textcolor{ourscolor}{\textbf{Qwen3-1.7B}} & \chartdeltabgpct{36.69}{70.18} & \chartdeltabgpct{54.28}{96.26} & \chartdeltabgpct{53.14}{95.35} & \chartdeltabgpct{35.22}{73.41} & \chartdeltabgpct{44.83}{83.80} & \chartdeltabgpct{37.77}{89.63} & \chartdeltabgpct{23.04}{72.96} & \chartdeltabgpct{26.19}{77.59} & \chartdeltabgpct{16.88}{55.74} & \chartdeltabgpct{17.09}{50.37} & \chartdeltabgpct{24.19}{69.26} \\

\textcolor{ourscolor}{\textbf{Qwen3-4B}} & \chartdeltabgpct{51.36}{71.10} & \chartdeltabgpct{73.26}{96.31} & \chartdeltabgpct{72.27}{95.48} & \chartdeltabgpct{50.28}{75.50} & \chartdeltabgpct{61.79}{84.60} & \chartdeltabgpct{54.70}{98.33} & \chartdeltabgpct{34.66}{76.81} & \chartdeltabgpct{35.96}{78.15} & \chartdeltabgpct{23.61}{56.85} & \chartdeltabgpct{25.50}{54.07} & \chartdeltabgpct{34.89}{72.84} \\

\textcolor{ourscolor}{\textbf{Qwen3-8B}} & \chartdeltabgpct{54.60}{73.75} & \chartdeltabgpct{78.65}{98.03} & \chartdeltabgpct{77.66}{97.21} & \chartdeltabgpct{55.72}{78.56} & \chartdeltabgpct{66.66}{86.89} & \chartdeltabgpct{61.32}{98.98} & \chartdeltabgpct{38.38}{78.74} & \chartdeltabgpct{43.68}{88.62} & \chartdeltabgpct{27.84}{59.26} & \chartdeltabgpct{29.11}{56.69} & \chartdeltabgpct{40.07}{76.46} \\

\bottomrule
\end{tabularx}

\caption{\textbf{Performance Evaluation on Chart.} We evaluate \textit{$\mathcal{C} \rightarrow \mathcal{V}$} performance for $M_\theta$ ($\mathcal{V} \rightarrow \mathcal{C}$) and $M_\psi$ ($\mathcal{T} \rightarrow \mathcal{C}$) using \textit{rule-as-judge} and \textit{MLLM-as-judge}. Some evaluation dimensions are shown as abbreviations for clarity, where \textit{str.}, \textit{sem.}, \textit{sim.}, and \textit{acc.}, represent \textit{visual structure}, \textit{visual semantics}, \textit{visual similarity}, and \textit{visual accuracy}, respectively. Compared to \textcolor{notunecolor}{\textit{baselines}}, \ours-enhanced models are optimized at train time and/or test time in \textit{student} mode.}
\label{tab:exp:exp_main_chart_evaluation}

\vspace{12pt}

\end{table}

\section{Experiments}
\label{sec:experiments}

\subsection{Setup}
\label{subsec:exps:setup}

\textbf{Data.}
We randomly sample non-overlapping training and test subsets from \textit{ChartCoder}~\citep{chartcoder2025benchmark} through outcome-grounded filtering (\S\ref{appendix:sec:dataset_construction}).
For main evaluation, we measure model performance on six cyclic tasks of the \textit{chart-table-code} learning cycle (\S\ref{subsec:problem_formulation}), including $\mathcal{V}\Leftrightarrow\mathcal{T}$, $\mathcal{V}\Leftrightarrow\mathcal{C}$, and $\mathcal{C}\Leftrightarrow\mathcal{T}$.
To evaluate model generalizability and robustness, we extend our evaluation to (1) \textit{out-of-domain data}: multi-domain test sets adapted from \textit{ChartMimic}~\citep{chartmimic2025benchmark} (\textit{e.g., physics, mathematics, economics, biology, etc.}) and \textit{ChartNet}~\citep{2026chartnet} (\textit{e.g., health, finance, etc.}); and (2) \textit{out-of-domain tasks}: multi-level complexity tasks on \textit{Chart2Code}~\citep{chart2code2025benchmark} and \textit{ChartMimic}~\citep{chartmimic2025benchmark} (\S\ref{appendix:subsec:dataset_construction:evaluation}).

\noindent\textbf{Model.}
We employ three MLLMs and six LLMs as $M_\theta$ and $M_\psi$, including \texttt{Qwen3-VL} (2B \& 4B) and \texttt{InternVL3.5} (4B), \texttt{Qwen3} (1.7B, 4B, 8B), \texttt{DeepSeek-Coder} (1.3B \& 6.7B), and \texttt{Llama-3.2} (3B)~\citep{qwen3technicalreport2025,wang2025internvl35,guo2024deepseekcoder,grattafiori2024llama3herdmodels}. For evaluation, we leverage \texttt{GPT-5-mini} and \texttt{Gemini-3-Pro} as LLM and MLLM judges (\S\ref{subsec:method:evaluation_metrics} \& Fig.~\ref{fig:interrater_agreement}).

\noindent\textbf{Implementation Details.}
\ours trains models for $N_{\textit{epoch}}=2$ epochs across $N_{\textit{train}}=200$ steps, with an initial learning rate ${\textit{lr}}=1\times10^{-6}$ and weighted decay $1\times10^{-2}$ via \texttt{AdamW}. We configure $K_\theta=K_\psi=4$, with global batch size $B=8$.
%
%
Our configuration details are in \S\ref{appendix:sec:implementation_details}.


\subsection{Cross-Representation Learning}
\label{subsec:exp:main_results}

\begin{figure}[!t]
    \vspace{0pt}

    \small
    \centering
    \includegraphics[width=1.0\textwidth]{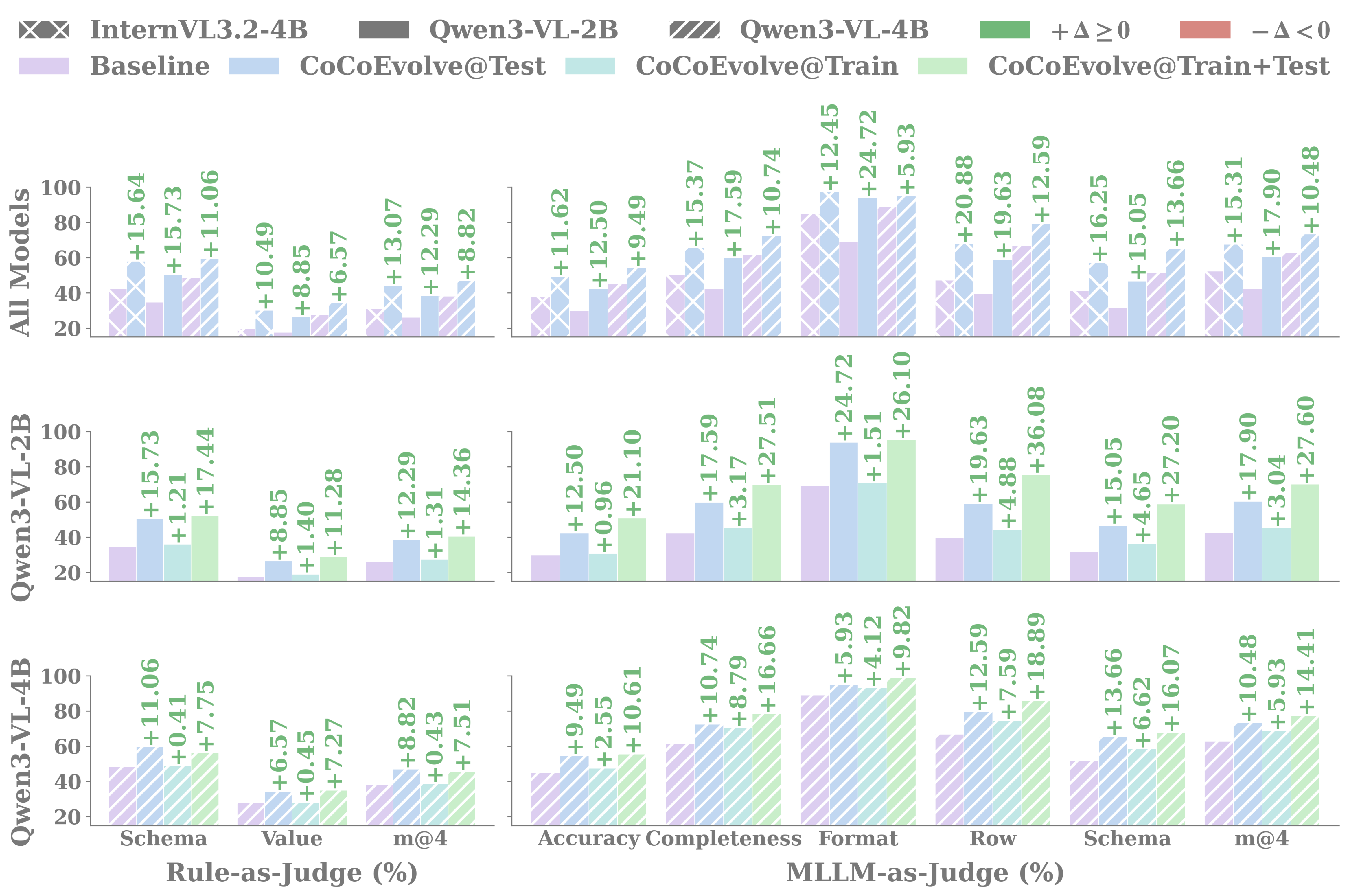}

    \vspace{0pt}
    
    \caption{\textbf{Performance Evaluation on Table.} We evaluate table performance for $M_\theta$ ($\mathcal{V} \rightarrow \mathcal{T}$ and $\mathcal{C} \rightarrow \mathcal{T}$) using \textit{rule-as-judge} and \textit{MLLM-as-judge}. The \textit{data accuracy}, \textit{coverage completeness}, and \textit{row alignment} evaluation dimensions are represented respectively as \textit{acc.}, \textit{coverage}, and \textit{row} for clarity. Compared to \textcolor{notunecolor}{\textit{baselines}}, \ours-enhanced models are optimized at train time and/or test time in \textit{student} mode.}
    \label{fig:exp:exp_main_table_evaluation}

    \vspace{6pt}

\end{figure}

\noindent\textbf{\ours Improves Cross-Representation Understanding Across $\bm{\mathcal{V}\Leftrightarrow\mathcal{C}\Leftrightarrow\mathcal{T}}$ Representation Cycle.} As shown in Tab.~\ref{tab:exp:exp_main_code_evaluation}, \ours consistently improves code quality across $M_\theta$ and $M_\psi$, achieving up to 100.00\% sandbox execution succuss with \ourstrain{}+\textbf{\textcolor{oursttcolor}{@test}}, together with gains of up to $\Delta=\uparrow17.48\%$ and $\Delta=26.57\%$ in rule-as-judge and LLM-as-judge evaluations, respectively. Chart evaluation further showcases robust improvements, yielding gains of $\Delta\geq6.79\%$ and $\Delta\geq3.33\%$ across rule-as-judge and MLLM-as-judge metrics, respectively. Moreover, table evaluation (Fig.~\ref{fig:exp:exp_main_table_evaluation}) also reveals stable gains through \ours, showcasing enhanced performance with up to $\Delta=15.64\%$ and $\Delta=24.72\%$ in rule-as-judge and MLLM-as-judge, respectively.
Collectively, these evaluations demonstrate that \ours effectively strengthens multifaceted cross-representation understanding capabilities of various models across chart, table, and code.

\subsection{Generalizability \& Robustness}

\begin{figure*}[!t]
    \vspace{0pt}

    \small
    \centering
    \includegraphics[width=1.0\textwidth]{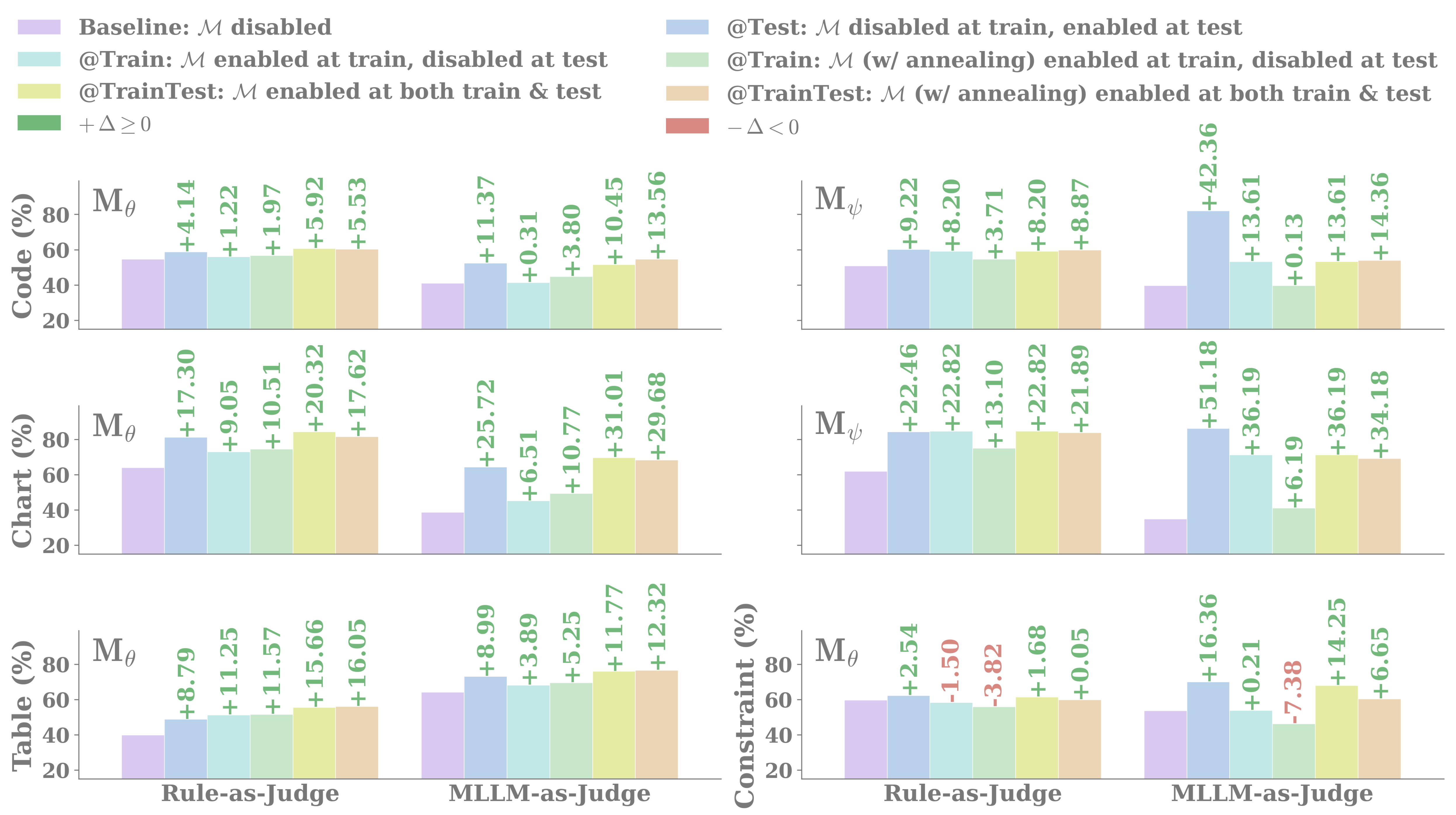}

    \vspace{0pt}
    
    \caption{\textbf{Ablation Study on Teacher Module.} Capable of seamlessly generalizing to supervised learning, we study the effects of $\mathcal{M}$ (\S\ref{subsec:teacher_module}) in guiding cross-representation understanding.}
    \label{fig:exp:ablation_teacher_module}

    \vspace{6pt}

\end{figure*}

\noindent\textbf{Teacher Module with High-Quality Supervision Facilitates Cross-Representation Learning.}
Equipped with $\mathcal{M}$, \ours is naturally generalizable to supervised learning. Using outcome-verified supervision to guide optimization (\S\ref{appendix:sec:dataset_construction}), we study the effect of $\mathcal{M}$ during train and test time. As shown in Fig.~\ref{fig:exp:ablation_teacher_module}, enabling $\mathcal{M}$ only at test time consistently improves performance across all six cross-representation tasks, yielding gains of $\Delta\geq2.54\%$ across representations. This finding suggests that $\mathcal{M}$ with high-quality supervision can effectively promote transferable cross-representation understanding. 
Nevertheless, performance becomes less stable when $\mathcal{M}$ is enabled only during training. In contrast, enabling $\mathcal{M}$ at both training and test time yields more balanced overall performance among all cross-representation tasks, indicating that supervision learned during training and teacher-assisted inference complement each other in improving cross-representation reasoning.


\noindent\textbf{\ours Improves Generalizability to Out-of-Domain Tasks.}
Leveraging Chart2Code and Chart2Mimic with varying tasks and complexity levels (\S\ref{subsec:exps:setup}), Fig.~\ref{fig:exp_ood} demonstrates that \ours generalizes effectively to out-of-domain settings, extending beyond direct chart$\rightarrow$code reproduction to more challenging scenarios involving multi-input chart reproduction and modification. On ChartMimic, \ours improves the performance of both $M_\theta$ and $M_\psi$ by up to $\uparrow37.97\%$ on chart$\rightarrow$code reproduction, and up to $\uparrow34.65\%$ on chart+data$\rightarrow$code reproduction. Likewise, on Chart2Code, \ours yields substantial gains across all complexity levels and out-of-domain tasks, achieving improvements of up to $\uparrow37.91\%$ on chart$\rightarrow$code reproduction, $\uparrow46.88\%$ on chart+figure$\rightarrow$code modification, $\uparrow27.85\%$ on chart+table$\rightarrow$code modification, and $\uparrow45.44\%$ on chart+instruction$\rightarrow$code modification. Collectively, these results reveal the robust generalizability of \ours to out-of-domain tasks with varying complexity and distribution shifts.

\noindent\textbf{\ours Improves Generalizability to Multi-Domain Cross-Representation Understanding.}
Finetuned on $10,298$ instances adapted exclusively from ChartCoder (\S\ref{appendix:sec:dataset_construction}), \ours exhibits strong generalizability to multi-domain cross-representation understanding. As shown in Fig.~\ref{fig:exp_ood}, in addition to achieving $\Delta\geq2.41\%$ on the non-overlapping test set of ChartCoder (Tab.~\ref{tab:data_statistics}), \ours further transfers effectively to data across various domains, yielding improvements of up to $\uparrow35.68\%$ on ChartNet and $\uparrow37.97\%$ on ChartMimic. Together, these results demonstrate the robust cross-domain generalization of \ours across diverse tasks and representations.

\clearpage
\begin{wrapfigure}{r}{0.6\textwidth}

    \vspace{-6pt}

    \small
    \centering
    \includegraphics[width=\linewidth]{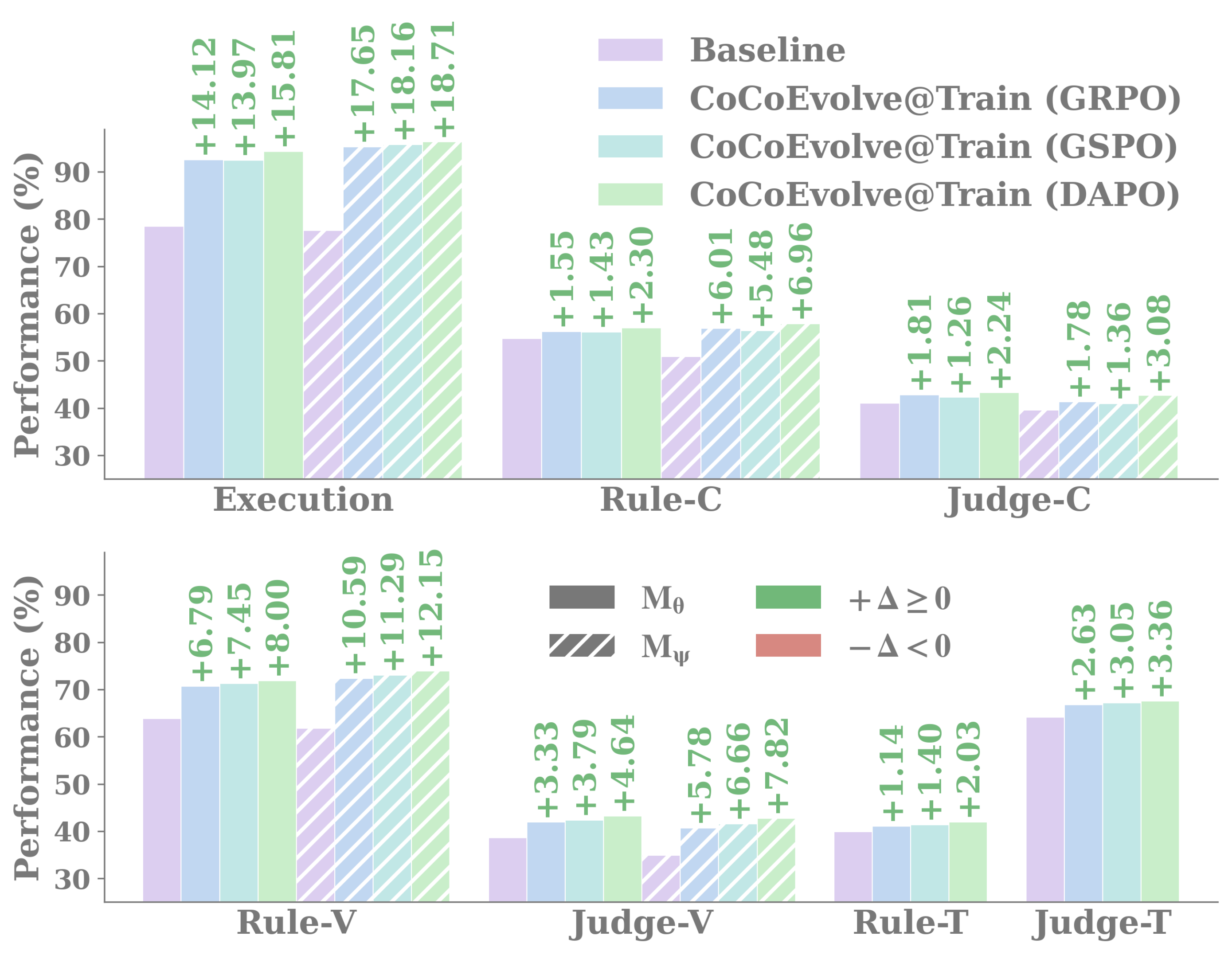}

    \vspace{-10pt}
    
    \caption{\textbf{Generalizability to Different RL Algorithms.} Rule- and Judge- indicate if they are rule-based or LLM- / MLLM- based judges.}
    \label{fig:exp_ablation_rl_algorithm}

    \vspace{-16pt}
\end{wrapfigure}

\noindent\textbf{\ours Integrates Seamlessly with Different RL Algorithms.}
Through consistency-driven optimization (\S\ref{sec:method}), we integrate \ours into GRPO, DAPO, and GSPO and evaluate its effectiveness across $M_\theta$=\texttt{Qwen3-VL-4B} and $M_\psi$=\texttt{Qwen3-4B} (\S\ref{subsec:ours_at_train_time}).
As shown in Fig.~\ref{fig:exp_ablation_rl_algorithm}, \ours consistently improves the cross-understanding performance of both $M_\theta$ and $M_\psi$ across all six tasks in the representation cycle, irrespective of the underlying RL algorithm ($\Delta$ up to $\uparrow18.71\%$ in execution, $\uparrow6.96\%$ in code, $\uparrow12.15\%$ in chart, $\uparrow3.36\%$ in table). Among evaluated algorithms, DAPO yields the most systematic improvements in cross-representation understanding.

\section{Conclusions}

In this work, we introduce \ours, a \textbf{cross-representation consistency-driven co-evolve framework} for structured multimodal reasoning with \textit{self-evolving training dynamic}: as one model improves, it provides stronger supervision to the other, gradually refining both. More broadly, our approach demonstrates that cross-modal consistency can serve as a scalable and reliable supervision signal, offering an alternative to annotation-heavy pipelines for structured reasoning tasks.


\bibliography{iclr2027_conference}
\bibliographystyle{iclr2027_conference}

\clearpage
\appendix

\section{Limitations}
\label{sec:limitations}

In this work, we propose \ours, a consistency-driven co-evolution framework for optimizing cross-presentation understanding (\S\ref{sec:method}). While \ours demonstrates promising results in annotation-free cross-representation learning, several limitations remain. First, our current framework instantiates $M_{\theta}$ and $M_{\psi}$ with small-size model architectures, exploring broader families of MLLMs and LLMs with larger sizes (\textit{e.g.}, 480B) may reveal how architectural choices interact with the co-evolution dynamic and further improve performance. Second, \ours currently operates on static chart images, while its applicability to dynamic or animated visualizations, where temporal relationships across frames introduce additional representational complexity beyond the \textit{chart-table-code} cycle (\S\ref{subsec:problem_formulation}), is underexplored. Extending \ours to animated cross-representation understanding, where visual, tabular, and code representations must jointly account for temporal semantics, remains an exciting direction for future work.

\section{Preliminaries}
\label{appendix:sec:preliminaries}

\subsection{Can Assumed One-to-One Annotations Reliably Supervise Model Learning?}
\label{appendix:subsec:preliminary:Q1:one_to_one}

\begin{hlbox}{Q1: Can Assumed One-to-One Annotations Reliably Supervise Model Learning?}
Existing cross-representation benchmarks assume \textit{one-to-one} chart-table-code mappings despite their inherently \textit{one-to-many} nature. Can an MLLM trained on such ill-defined \textit{unconstrained one-to-one} annotations achieve optimized performance?
\end{hlbox}

To answer Q1, we finetune \texttt{Qwen3-VL-4B} ($M_\theta$) and \texttt{Qwen3-4B} ($M_\psi$) via SFT, finetuned for 1 epoch on 30,000 instances randomly sampled from \textit{ChartNet}~\citep{2026chartnet}, a benchmark that assumes \textit{one-to-one} chart-table-code correspondences. We then evaluate both the base models and the SFT-finetuned $M_\theta$ and $M_\psi$ on the same held-out test set of 542 non-overlapping instances randomly sampled from the same benchmark (Tab.~\ref{tab:data_statistics}).

\begin{figure*}[!t]
    \vspace{-16pt}

    \small
    \centering
    \includegraphics[width=1.0\textwidth]{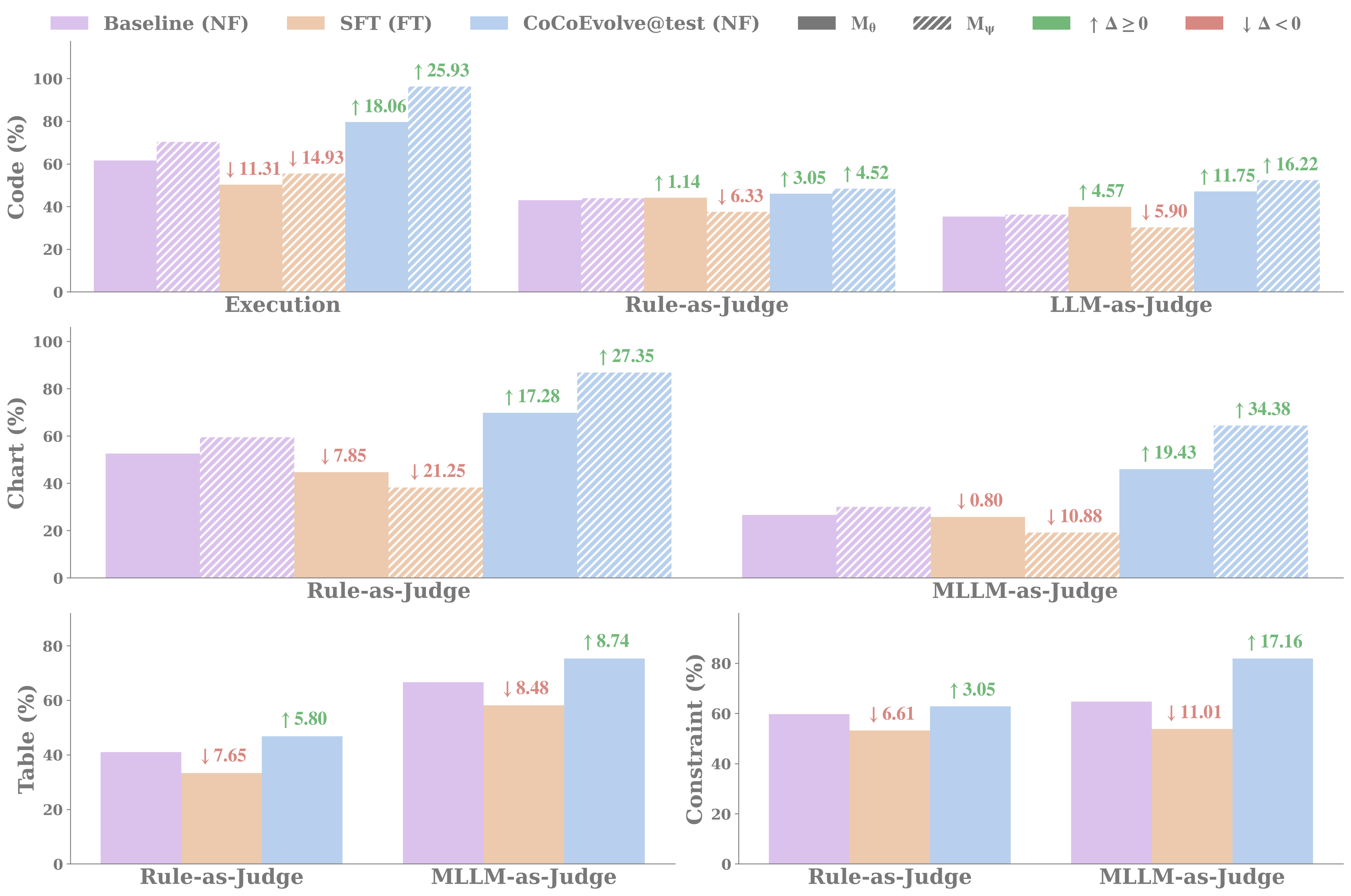}
    
    \vspace{-6pt}
    
    \caption{\textbf{Significance of Constraints in Cross-Representation Understanding.} Comparing $M_\theta$ and $M_\psi$ variants among baseline, non-finetuned (NF) \ourstest (\S\ref{subsec:ours_at_test_time}), and SFT (\S\ref{appendix:subsec:preliminary:Q1:one_to_one}), SFT-trained $M_\theta$ and $M_\psi$ exhibit notably degraded performance, highlighting the significance of constrained one-to-one mapping in cross-representation understanding.}
    \label{fig:preliminary1_result}

    \vspace{12pt}

\end{figure*}

As shown in Fig.~\ref{fig:preliminary1_result}, we evaluate $M_\theta$ and $M_\psi$ under three settings: non-finetuned baseline, SFT, and non-finetuned \ourstest, using the same non-overlapping test set adapted from ChartNet. Although the SFT-trained $M_\theta$, finetuned on additional training instances, achieves slightly improved performance on chart-to-code generation ($\Delta=\uparrow1.14\%$ in rule-as-judge code evaluation, $\Delta=\uparrow4.57\%$ in LLM-as-judge code evaluation), it exhibits noticeably degraded performance on code, chart, and table evaluation across sandbox execution ($\Delta=\downarrow11.31\%$), rule-as-judge (up to $\Delta=\downarrow7.85\%$), and MLLM-as-judge (up tp $\Delta=\downarrow8.48\%$). Moreover, not only does $M_\theta$ perform worse on chart and table evaluation, but $M_\psi$ also shows significant degradation across chart, code, and table evaluation (up to $\Delta=\downarrow21.25\%$). In addition, both $M_\theta$ and $M_\psi$ exhibit significantly reduced code execution success rates (up to $\Delta=\downarrow14.93\%$). These results reveal that the assumed one-to-one annotations, \textit{without} properly defined constraints, are inherently one-to-many and thus ill-defined and unreliable for guiding and supervising model learning (\S\ref{sec:intro}). This motivates the critical need for a principled constraint definition to properly ground \textit{one-to-many} mappings to precise \textit{one-to-one} correspondences (\S\ref{subsec:problem_formulation}).

\subsection{Can Fixed Supervision Reliably Guides Cross-Representation Learning?}
\label{appendix:subsec:preliminary:Q2:supervision_signal}

\begin{hlbox}{Q2: Can Fixed Supervision Reliably Guides Cross-Representation Learning?}
Even with principled constraints (\S\ref{subsec:problem_formulation}), can an MLLM trained on fixed \textit{chart-to-table} and \textit{chart-to-code} pairs not only achieve robust performance on unseen instances of trained tasks (\textit{chart-to-table} and \textit{chart-to-code}), but also reliably generalize to held-out tasks, directions, and representation combinations?
\end{hlbox}

To answer \textit{Q2}, we further finetune \texttt{Qwen3-VL-4B} and \texttt{Qwen3-4B} via SFT (1 epoch) on exactly the same number of \textit{fixed} yet \textit{constrained} (\S\ref{subsec:problem_formulation}) \textit{chart-to-table} and \textit{chart-to-code} pairs as in our main experiments (\S\ref{appendix:sec:dataset_construction} \& Tab.~\ref{tab:data_statistics}). Note that while our main experiments use only \textit{chart-to-code} instances adapted from the randomly sampled training subset of the original benchmark, this preliminary study additionally includes fixed \textit{chart-to-table} annotations. Since ground-truth tables are unavailable, we use \texttt{Gemini-3-Pro} to generate the reference table and constraint through outcome-grounded filtering (\S\ref{appendix:sec:dataset_construction}). These MLLM-generated tables are used for both training and evaluation in this study (Tab.~\ref{tab:data_statistics}).

\begin{figure*}[!t]
    \vspace{0pt}

    \small
    \centering
    \includegraphics[width=1.0\textwidth]{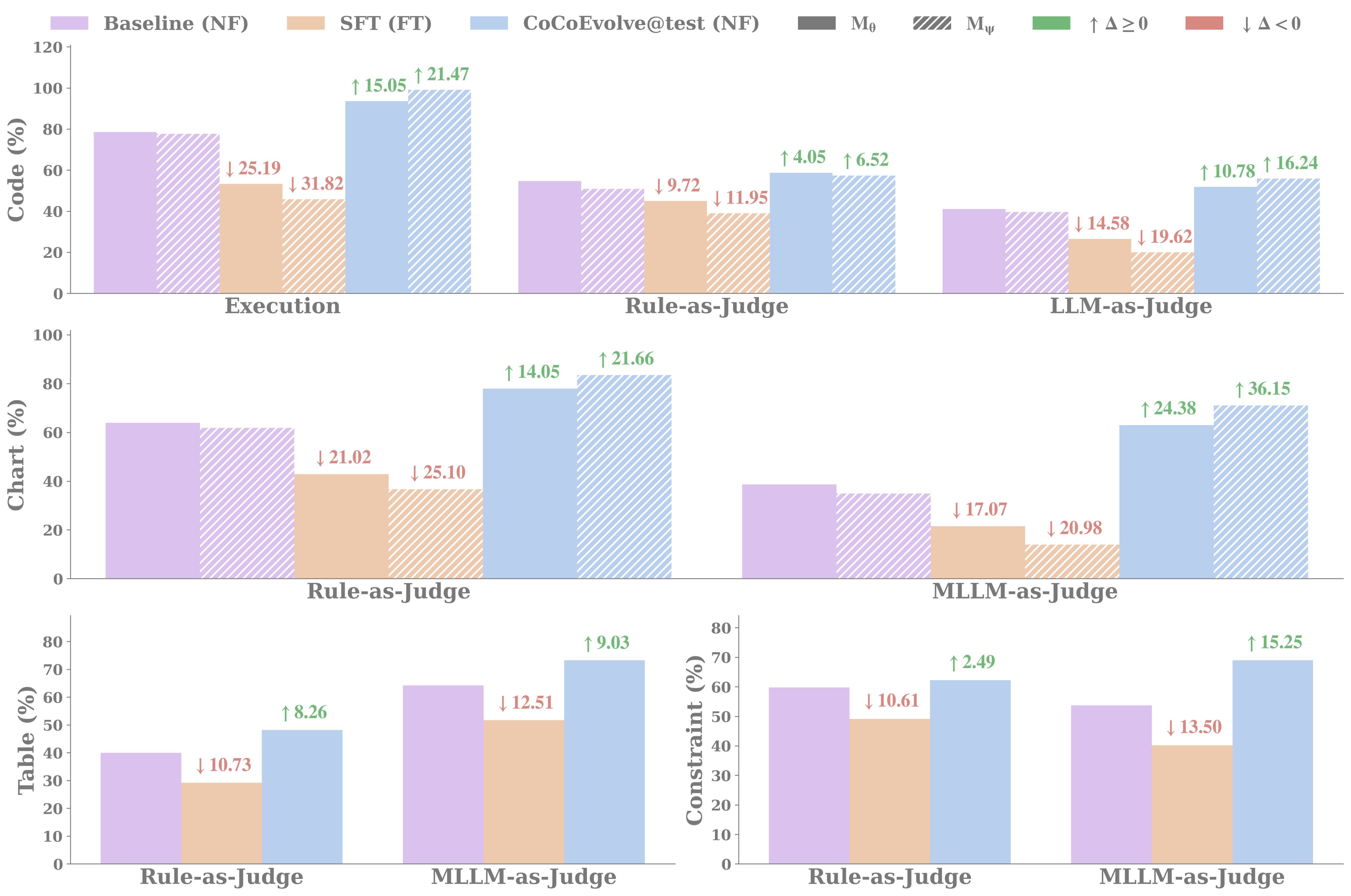}
    
    \vspace{-6pt}
    
    \caption{\textbf{Limitations of Fixed Supervision Signals in Cross-Representation Understanding.} Comparing $M_\theta$ and $M_\psi$ variants among baseline, non-finetuned (NF) \ourstest (\S\ref{subsec:ours_at_test_time}), and SFT (\S\ref{appendix:subsec:preliminary:Q1:one_to_one}), SFT-trained (FT) $M_\theta$ and $M_\psi$ show noticeably degraded performance, revealing the critical limitations of fixed annotations in supervising cross-representation learning.}
    \label{fig:preliminary2_result}

    \vspace{0pt}

\end{figure*}

Next, we evaluate $M_\theta$ and $M_\psi$ under three settings: non-finetuned baseline, SFT, and non-finetuned \ourstest, using the same non-overlapping test set across 
\textit{six} cross-representation tasks (\S\ref{subsec:problem_formulation} \& 
\S\ref{subsec:exps:setup}), covering trained tasks ($\mathcal{V}\Leftrightarrow\mathcal{T}$, $\mathcal{V}\Leftrightarrow\mathcal{C}$), reversed directions ($\mathcal{T}\Leftrightarrow\mathcal{V}$, $\mathcal{C}\Leftrightarrow\mathcal{V}$), and unseen representation combinations ($\mathcal{T}\Leftrightarrow\mathcal{C}$).
Results in Fig.~\ref{fig:preliminary2_result} demonstrate that fixed supervision signals not only fail to reliably optimize $M_\theta$ and $M_\psi$ on their trained tasks (with $\Delta$ up to $\downarrow25.19\%$), but are also unable to generalize to reversed directions (with $\Delta$ up to $\downarrow25.10\%$) and unseen representation combinations (with $\Delta$ up to $\downarrow31.82\%$), motivating the need for a principled, task-agnostic, and direction-agnostic supervision paradigm as proposed in \ours (\S\ref{sec:method}).

\subsection{How Do Models Fail in Cross-Representation Understanding?}
\label{appendix:subsec:preliminary:Q3:failure_analysis}


Motivated by our findings on the substantial unreliability of unconstrained cross-representation understanding (\S\ref{appendix:subsec:preliminary:Q1:one_to_one}) and the limitations of fixed supervision signals (\S\ref{appendix:subsec:preliminary:Q2:supervision_signal}), we conduct a third preliminary study to qualitatively analyze \emph{how} models fail during cross-representation understanding.

\begin{figure*}[!t]
    \vspace{36pt}

    \small
    \centering
    \includegraphics[width=1.0\textwidth]{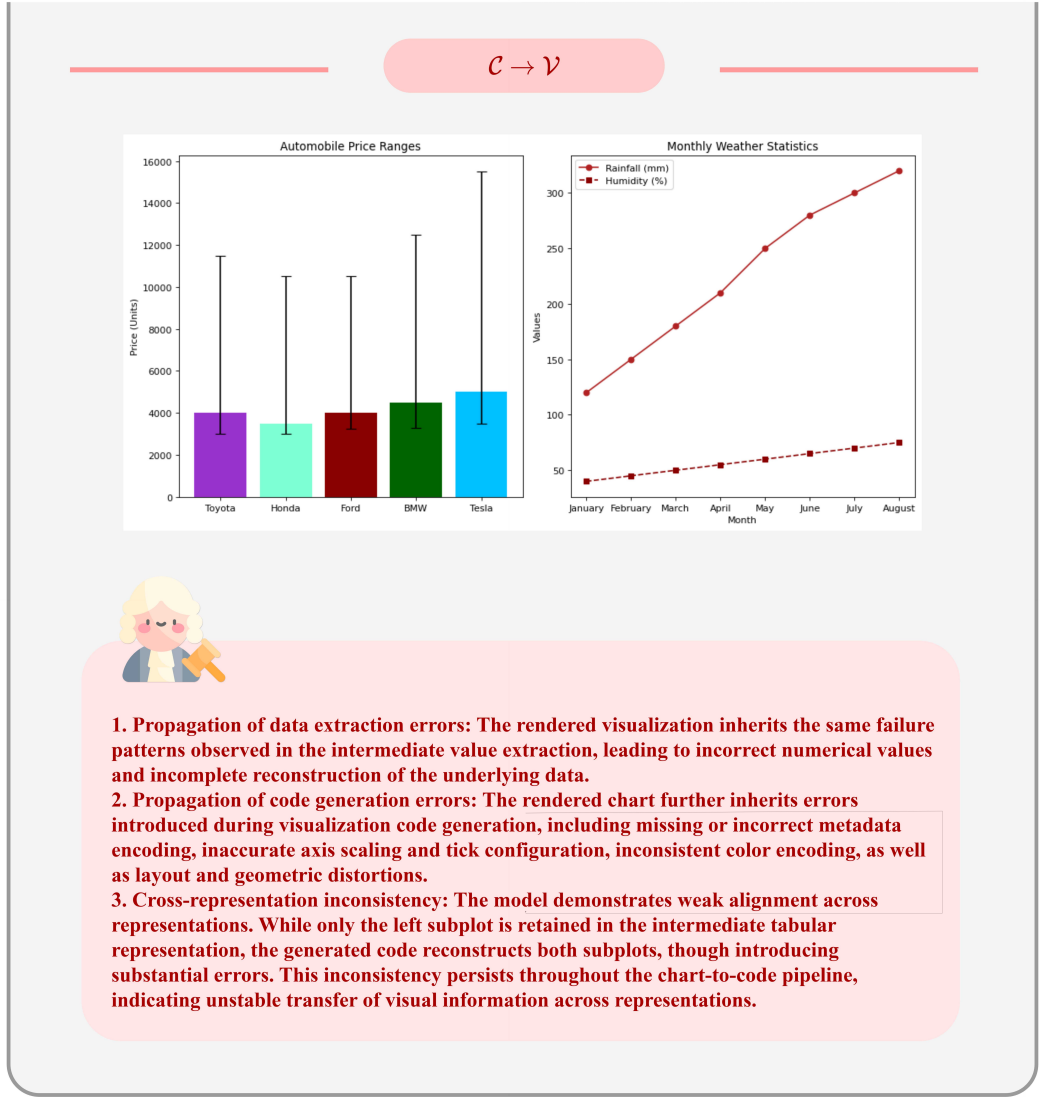}
    
    \caption{\textbf{Cross-Representation Understanding Failures.} A case study from our preliminary analysis illustrating how model errors accumulate and propagate across representations, resulting in inaccurate data extraction, incomplete information transfer, and significant cross-representation inconsistencies.}
    \label{fig:preliminary3}

    \vspace{60pt}

\end{figure*}

Fig.~\ref{fig:preliminary3} shows a representative failure example from our case studies. Given a chart image, we analyze the model outputs across multiple representations, including structured tables, visualization code, and rendered charts. We observe two key failure patterns:

\begin{oursenumerate}
    \item \textbf{Model failures are highly correlated across representations.} When the model fails to understand a chart, similar failures consistently appear across its generated representations. For example, incorrect numerical values, missing titles and labels, and incomplete information extraction observed in the table are also reflected in the generated code and rendered visualization. Different representations tend to exhibit similar understanding errors despite expressing information in different forms, suggesting that model failures originate from a shared deficiency in underlying chart understanding rather than isolated representation-specific errors. When situated in the representation cycle (\S\ref{subsec:problem_formulation}), such failures propagate rapidly across different representations, resulting in accumulated errors and increasingly amplified inaccuracies.

    \item \textbf{Representations remain highly inconsistent despite describing the same semantics.} Although different representations are expected to encode the same chart semantics, the generated outputs frequently disagree. In Fig.~\ref{fig:preliminary3}, the extracted table only contains information from the left subplot, while the generated visualization code attempts to reconstruct both subplots with substantial inaccuracies. Similar inconsistencies appear in rendered outputs through mismatched metadata, incorrect scales, distorted layouts, and inconsistent visual encodings. These observations suggest that existing models struggle to maintain semantic agreement across representations.
\end{oursenumerate}

\noindent Collectively, our analysis reveals that cross-representation failures are characterized not only by \textit{low fidelity within individual representations}, but also by \textit{poor agreement among representations}. This indicates that optimizing representations independently or relying on fixed supervision signals are insufficient for robust cross-representation understanding.

These findings motivate our \textbf{consistency-driven co-evolution framework} (\S\ref{sec:method}). Instead of leveraging unconstrained or fixed supervision signals (\S\ref{appendix:subsec:preliminary:Q1:one_to_one}-\ref{appendix:subsec:preliminary:Q2:supervision_signal}), \ours explicitly encourage agreement across representations and reward mutually consistent generations. By enforcing cross-representation consistency, models co-evolve symbiotically, improving understanding fidelity within each representation and enhancing alignment across representations, thereby optimizing overall cross-representation understanding.

\clearpage

\begin{table*}[h]
\small
\centering

\renewcommand{\arraystretch}{1.1}
\setlength{\tabcolsep}{5.5pt}

\begin{tabular*}{\textwidth}{@{\extracolsep{\fill}}lcccccc}
\toprule
\multirow{3}{*}{\textbf{Source}}
    & \multicolumn{2}{c}{$\bm{\mathbf{\mathcal{V} \Leftrightarrow \mathcal{C}}}$}
    & \multicolumn{2}{c}{$\bm{\mathbf{\mathcal{V} \Leftrightarrow \mathcal{T}}}$}
    & \multicolumn{2}{c}{$\bm{\mathbf{\mathcal{T} \Leftrightarrow \mathcal{C}}}$} \\

\cmidrule(lr){2-3} \cmidrule(lr){4-5} \cmidrule(lr){6-7}

    & $\bm{\mathbf{\mathcal{V} \rightarrow \mathcal{C}}}$
    & $\bm{\mathbf{\mathcal{C} \rightarrow \mathcal{V}}}$
    & $\bm{\mathbf{\mathcal{V} \rightarrow \mathcal{T}}}$
    & $\bm{\mathbf{\mathcal{T} \rightarrow \mathcal{V}}}$
    & $\bm{\mathbf{\mathcal{T} \rightarrow \mathcal{C}}}$
    & $\bm{\mathbf{\mathcal{C} \rightarrow \mathcal{T}}}$ \\

\midrule

\multicolumn{7}{c}{\cellcolor{STUDENT_BG}\textbf{Preliminary Study - Q1 ( \textit{train}: $\bm{30,000}$ $\mid$ \textit{test}: $\bm{542}$ )}} \\
\addlinespace[3pt]

\textbf{ChartNet}~\citep{2026chartnet} & \tcmarkn{1} & \tcmark & \tcmark & \tcmark & \tcmark & \tcmark \\

\textbf{$\quad\quad\quad\quad$Data} & \textit{adapted} & \textcolor{sandboxcolor}{\textit{sandbox}} & \textit{adapted} & \textit{adapted} & \textit{adapted} & \textit{adapted} \\

\textbf{$\quad\quad\quad\quad N_{\textit{train}}$} & $30,000$ & - & - & - & $30,000$ & - \\

\textbf{$\quad\quad\quad\quad N_{\textit{test}}$} & $542$ & - & $542$ & $542$ & $542$ & $542$ \\

\multicolumn{7}{c}{\cellcolor{STUDENT_BG}\textbf{Preliminary Study - Q2 ( \textit{train}: $\bm{10,298}$ $\mid$ \textit{test}: $\bm{542}$ )}} \\
\addlinespace[3pt]

\textbf{ChartCoder}~\citep{chartcoder2025benchmark} & \tcmarkn{1} & \tcmark & \tcmark & \tcmark & \tcmark & \tcmark \\

\textbf{$\quad\quad\quad\quad$Data} & \textit{adapted} & \textcolor{sandboxcolor}{\textit{sandbox}} & \textcolor{teachergreen}{\textit{generated}} & \textcolor{teachergreen}{\textit{generated}} & \textcolor{teachergreen}{\textit{generated}} & \textcolor{teachergreen}{\textit{generated}} \\

\textbf{$\quad\quad\quad\quad N_{\textit{train}}$} & $10,298$ & - & - & - & - & - \\

\textbf{$\quad\quad\quad\quad N_{\textit{test}}$} & $3,411$ & - & $1082$ & $1082$ & $1082$ & $1082$ \\

\multicolumn{7}{c}{\cellcolor{STUDENT_BG}\textbf{Train ( \textit{train}: $\bm{10,298}$ )}} \\
\addlinespace[3pt]

\textbf{ChartCoder}~\citep{chartcoder2025benchmark} & \tcmarkn{1} & \tcmark & \tcmark & \tcmark & \tcmark & \tcmark \\


\textbf{$\quad\quad\quad\quad$Data} & \textit{adapted} & \textcolor{sandboxcolor}{\textit{sandbox}} & - & - & - & - \\

\textbf{$\quad\quad\quad\quad N_{\textit{train}}$} & $10,298$ & - & - & - & - & - \\


\multicolumn{7}{c}{\cellcolor{STUDENT_BG}\textbf{Test ( \textit{test}: $\bm{3,411}$ )}} \\
\addlinespace[3pt]

\textbf{ChartCoder}~\citep{chartcoder2025benchmark} & \tcmarkn{1} & \tcmark & \tcmark & \tcmark & \tcmark & \tcmark \\

\textbf{$\quad\quad\quad\quad$Data} & \textit{adapted} & \textcolor{sandboxcolor}{\textit{sandbox}} & \textcolor{teachergreen}{\textit{generated}} & \textcolor{teachergreen}{\textit{generated}} & \textcolor{teachergreen}{\textit{generated}} & \textcolor{teachergreen}{\textit{generated}} \\

\textbf{$\quad\quad\quad\quad N_{\textit{test}}$} & $542$ & - & $542$ & $542$ & $542$ & $542$ \\

\arrayrulecolor{gray!50}\cmidrule(l){1-7}\arrayrulecolor{black}

\textbf{ChartMimic}~\citep{chartmimic2025benchmark} & \tcmarkn{2} & \tcmark & \txmark & \txmark & \txmark & \txmark \\

\textbf{$\quad\quad\quad\quad$Data} & \textit{adapted} & \textcolor{sandboxcolor}{\textit{sandbox}} & - & - & - & - \\

\textbf{$\quad\quad\quad\quad N_{\textit{test}}$} & $1084$ & - & - & - & - & - \\

\arrayrulecolor{gray!50}\cmidrule(l){1-7}\arrayrulecolor{black}

\textbf{Chart2Code}~\citep{chart2code2025benchmark} & \tcmarkn{4} & \tcmark & \txmark & \txmark & \txmark & \txmark \\

\textbf{$\quad\quad\quad\quad$Data} & \textit{adapted} & \textcolor{sandboxcolor}{\textit{sandbox}} & - & - & - & - \\

\textbf{$\quad\quad\quad\quad N_{\textit{test}}$} & $1243$ & - & - & - & - & - \\

\arrayrulecolor{gray!50}\cmidrule(l){1-7}\arrayrulecolor{black}

\textbf{ChartNet}~\citep{2026chartnet} & \tcmarkn{1} & \tcmark & \tcmark & \tcmark & \tcmark & \tcmark \\

\textbf{$\quad\quad\quad\quad$Data} & \textit{adapted} & \textcolor{sandboxcolor}{\textit{sandbox}} & \textit{adapted} & \textit{adapted} & \textit{adapted} & \textit{adapted} \\

\textbf{$\quad\quad\quad\quad N_{\textit{test}}$} & $542$ & - & $542$ & $542$ & $542$ & $542$ \\

\bottomrule
\end{tabular*}

\caption{\textbf{Data Statistics.} We summarize the data statistics of our datasets used in preliminary studies, training, and evaluation. All \textit{adapted} data are randomly sampled from the original benchmarks and adapted to our cross-representation tasks, \textcolor{sandboxcolor}{\textit{sandbox}} denotes sandbox execution, and \textcolor{teachergreen}{\textit{generated}} represents MLLM-generated samples where we use \texttt{Gemini-3.1-Pro} to generate the ground-truth table and constraint for each chart instance in the training subset randomly sampled from the original benchmark. \textcolor{checkgreen}{\textbf{\cmark ($n$)}} denotes the number of fine-grained chart-to-code generation tasks: (a) \textit{ChartCoder}: (1) chart$\rightarrow$code (reproduction); (b) \textit{ChartMimic}: (1) chart$\rightarrow$code (reproduction), (2) chart+data$\rightarrow$code (reproduction); (c) \textit{Chart2Code}: (1) chart$\rightarrow$code (reproduction), (2) chart+ instruction$\rightarrow$code (modification), (3) chart+table$\rightarrow$code (modification), (4) chart+figure$\rightarrow$code (modification); (d) \textit{ChartNet}: (1) chart$\rightarrow$code (reproduction).}
\label{tab:data_statistics}

\end{table*}

\begin{figure*}[!t]
    \vspace{12pt}

    \small
    \centering
    \includegraphics[width=1.0\textwidth]{assets/preliminary3_1.pdf}

    \vspace{24pt}

\end{figure*}

\begin{figure*}[!t]
    \vspace{12pt}

    \small
    \centering
    \includegraphics[width=1.0\textwidth]{assets/preliminary3_2.pdf}

    \vspace{12pt}

\end{figure*}

\begin{figure*}[h]
    \vspace{12pt}

    \small
    \centering
    \includegraphics[width=1.0\textwidth]{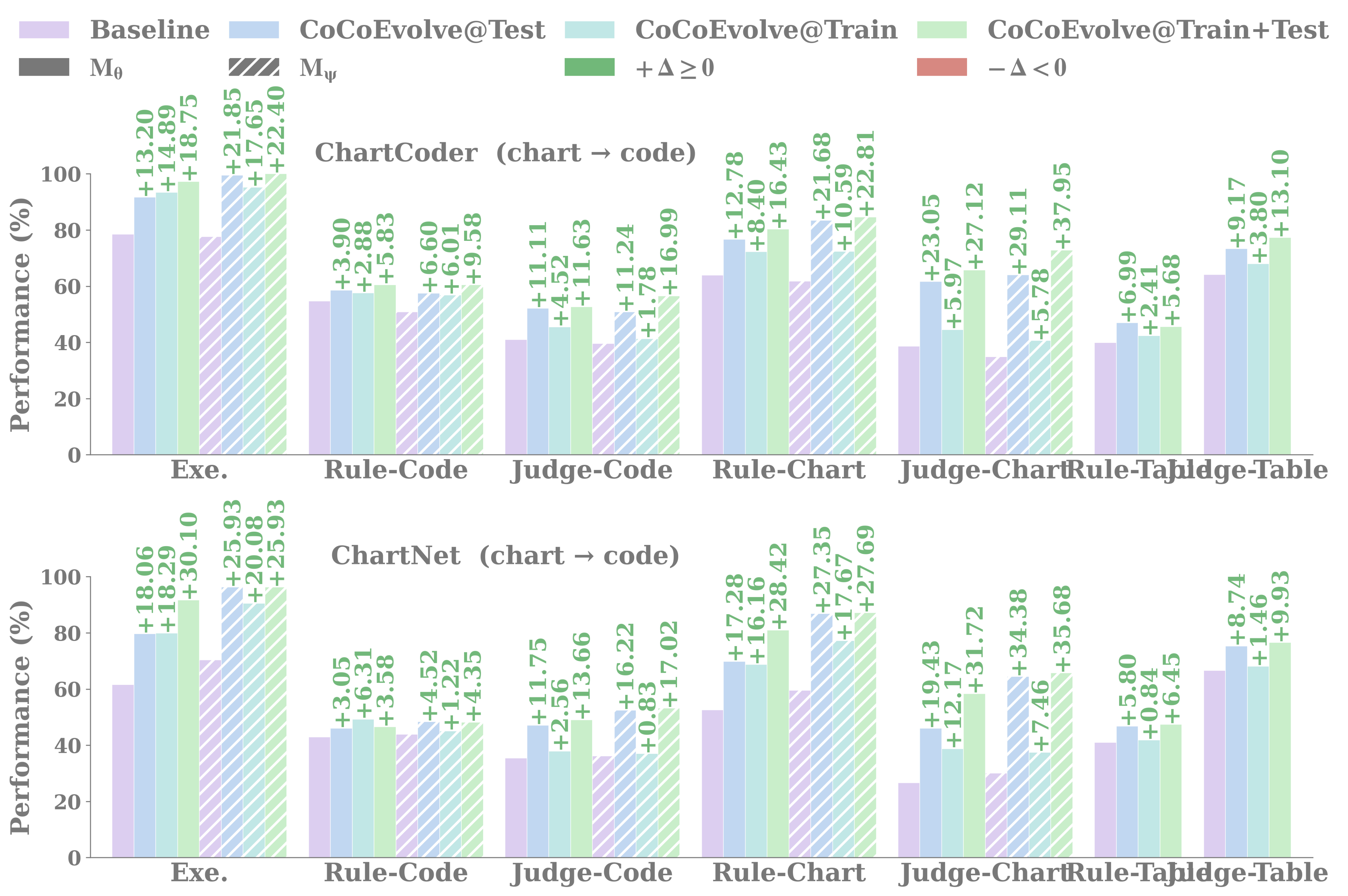}

    \vspace{36pt}
    
\end{figure*}

\begin{figure*}[h]
    \vspace{0pt}

    \small
    \centering
    \includegraphics[width=1.0\textwidth]{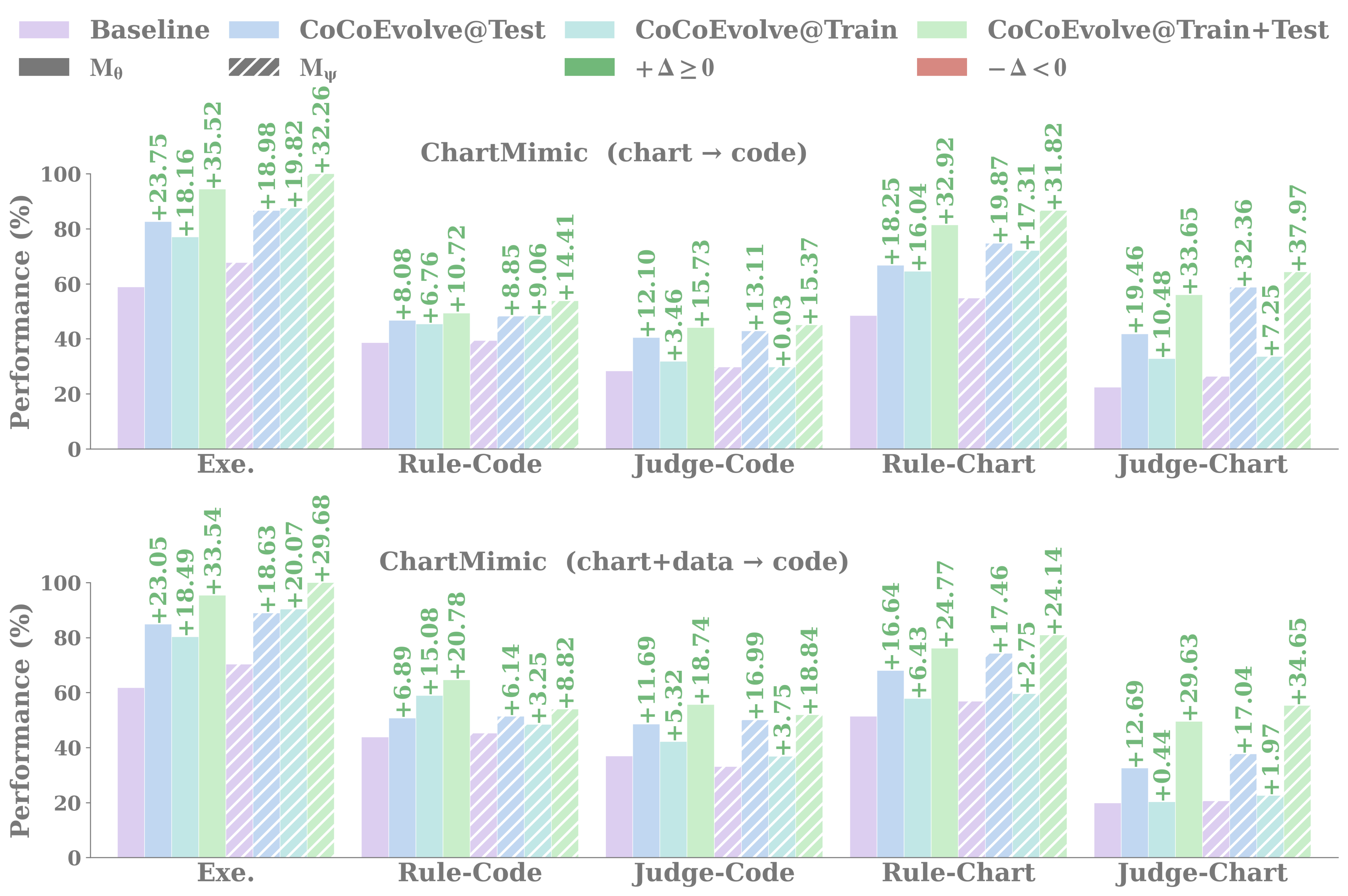}

    \vspace{0pt}
    
\end{figure*}

\begin{figure*}[h]
    \vspace{0pt}

    \small
    \centering
    \includegraphics[width=1.0\textwidth]{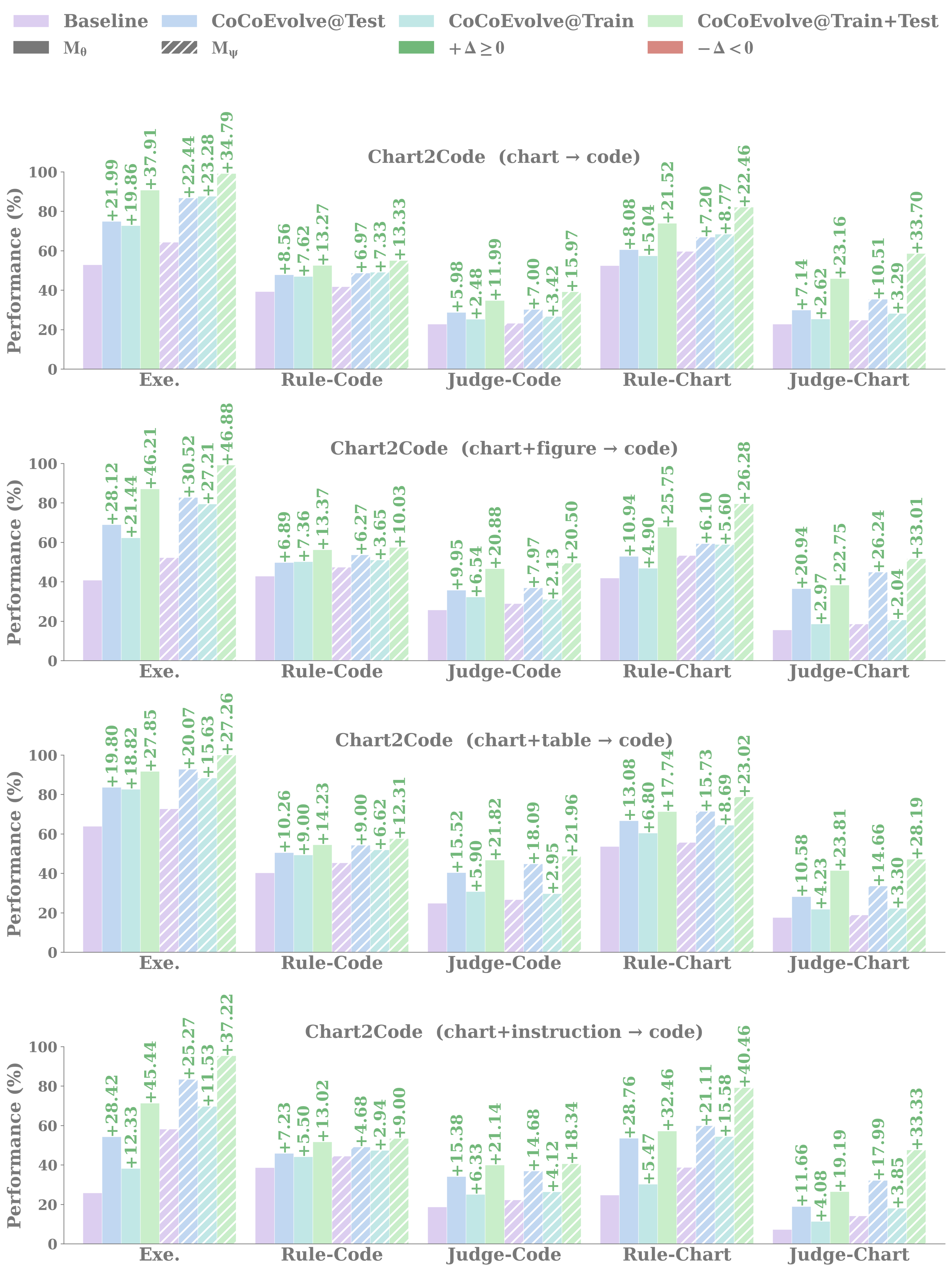}
    
    \caption{\textbf{Out-of-Domain Generalization.} We extend our evaluation to out-of-domain datasets and tasks (\S\ref{subsec:exps:setup}), leveraging \texttt{Qwen3-VL-4B} and \texttt{Qwen3-4B} as $M_
    \theta$ and $M_\psi$, respectively.}
    \label{fig:exp_ood}

    \vspace{0pt}

\end{figure*}
\begin{figure*}[!t]
    \vspace{0pt}

    \small
    \centering
    \includegraphics[width=1.0\textwidth]{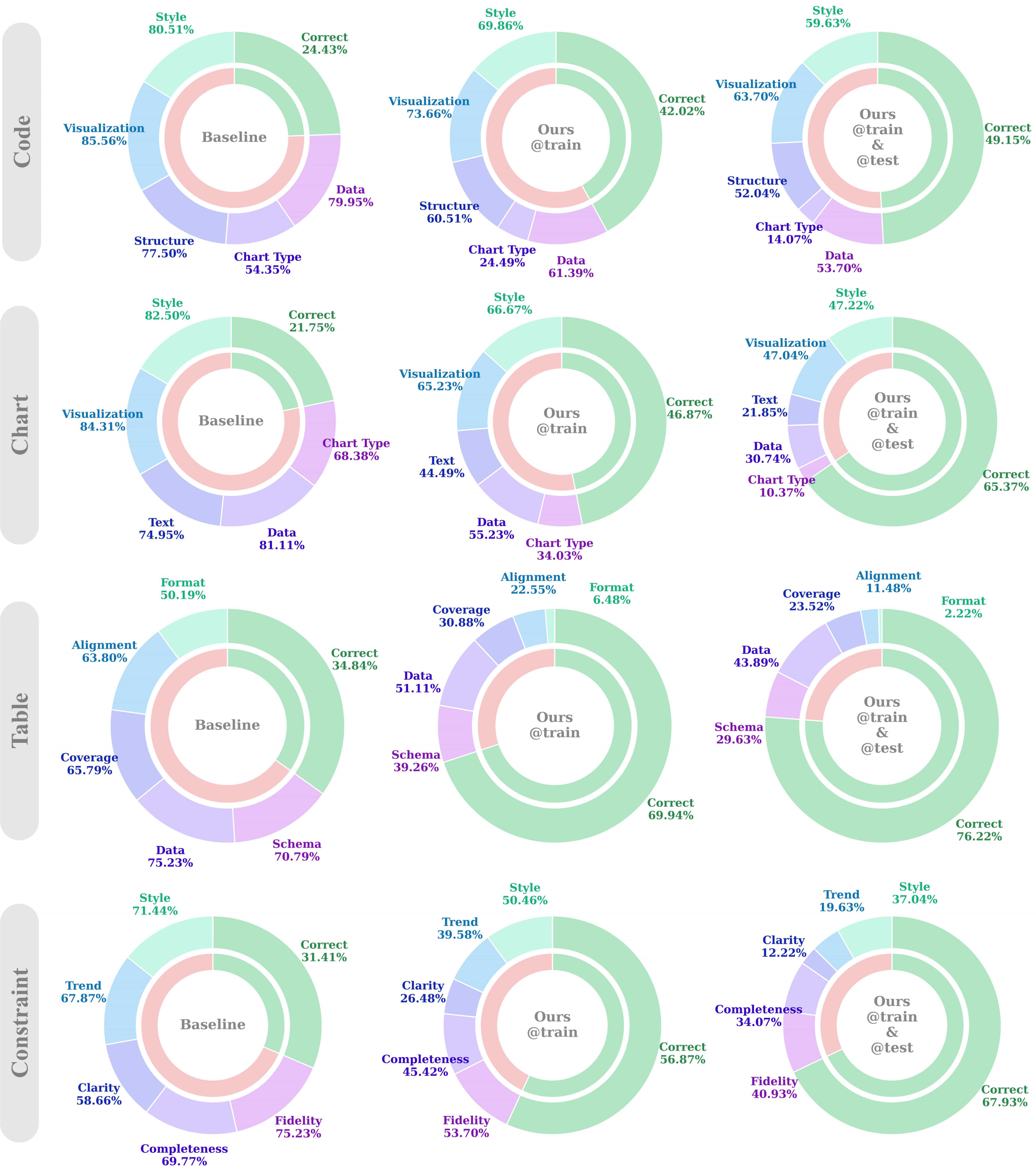}

    \vspace{6pt}
    
    \caption{\textbf{Failure Analysis.} We visualize the failure distribution of \texttt{Qwen3-4B-VL} in cross-representation understanding, where baseline is the non-finetuned base model, ours (@train) is \texttt{Qwen3-4B-VL} finetuned through \ourstrain, and ours (@train \& @test) is ours (@train) further optimized by \ourstest.}
    \label{fig:failure_analysis}

    \vspace{0pt}

\end{figure*}

\clearpage

\section{Dataset Construction}
\label{appendix:sec:dataset_construction}


We construct our training and evaluation datasets by adapting existing chart benchmarks to our cross-representation learning framework (\S\ref{sec:method}). All adapted instances are reformatted to our representation cycle instructions (Figs.~\ref{fig:eval_prompt:code_llm_as_judge}-\ref{fig:eval_prompt:constraint_mllm_as_judge}) and verified through sandbox execution. Tab.~\ref{tab:data_statistics} summarizes the full data statistics across our training and evaluation datasets.

\subsection{Sandbox Environment}
\label{appendix:subsec:dataset_construction:sandbox}

To verify the executability of visualization code across dataset construction, training, and evaluation, we create our sandbox execution environment via Docker~\citep{docker4sandbox} pre-installed with a set of commonly used data visualization and scientific computing packages, including \texttt{matplotlib}, \texttt{seaborn}, \texttt{plotly}, \texttt{pandas}, \texttt{numpy}, \texttt{scipy}, \texttt{scikit-learn}, \texttt{Pillow}, \texttt{squarify}, etc. Each code execution is run in an isolated container with a fixed timeout of $l_{\textit{timeout}}$ seconds, ensuring reproducibility and safety across all dataset construction, training, and evaluation stages. As such, our sandbox reliably powers our deterministic code executor $h$ (Eq.~\ref{eq:cycle}), where $h(\hat{c})$ returns the rendered chart image on success and fails otherwise.

\subsection{Training Dataset}
\label{appendix:subsec:dataset_construction:training}

We construct our training dataset exclusively from ChartCoder~\citep{chartcoder2025benchmark} through three filtering stages followed by a final sampling step:

\paragraph{Stage 1 on Filtering: Sandbox Executability.}
We filter instances whose reference code renders successfully via our deterministic executor $h$ (\S\ref{subsec:problem_formulation}), ensuring all training instances are verifiably executable in our sandbox environment.

\paragraph{Stage 2 on Filtering: Token Length.}
We filter instances satisfying $\texttt{len}\text{({\textit{input}})} \leq l_{\textit{input}}$ and $\texttt{len}\text{({\textit{c}})} \leq l_{\textit{code}}$, where $l_{\textit{input}}$ and $l_{\textit{code}}$ are maximum token length thresholds for the chart input and reference code, respectively.

\paragraph{Stage 3 on Filtering: Outcome-Grounded Verification.}
For each remaining instance, we generate candidate reference tables $t_{gt}$ and constraints $s_{gt}$ using \texttt{GPT-5-mini}~\citep{gpt5}, and retain only instances where the generated $t_{gt}$ and $s_{gt}$ verifiably improve downstream $M_\psi$ (powered by \texttt{Qwen3-4B}) code generation performance, ensuring their quality as reliable evaluation references. We further filter instances satisfying $\texttt{len}(t_{gt}) \leq l_{\textit{table}}$ and $\texttt{len}(s_{gt}) \leq l_{\textit{constraint}}$, where $l_{\textit{table}}$ and $l_{\textit{constraint}}$ are maximum token length thresholds for the reference table and constraint, respectively.

\noindent\textbf{Stage 4 on Sampling.} From the filtered dataset, we randomly downsample $10{,}298$ instances as our final training set. From the remaining instances, we further randomly sample $542$ non-overlapping instances as the in-domain evaluation set for the ChartCoder benchmark~\citep{chartcoder2025benchmark} adapted to our cross-representation understanding tasks.

\subsection{Evaluation Datasets}
\label{appendix:subsec:dataset_construction:evaluation}

We construct evaluation datasets from four benchmarks through the same filtering pipeline as the training dataset, adapting each to cross-representation learning instructions. All evaluation instances are non-overlapping with the training set.

\paragraph{Stages 1-3 on Filtering.}
All evaluation instances undergo the same three filtering stages as the training dataset (\S\ref{appendix:subsec:dataset_construction:training}): (1) sandbox executability verification via $h$; (2) token length filtering with thresholds $l_{\textit{input}}$, $l_{\textit{code}}$; and (3) reference table $t_{gt}$ and constraint $s_{gt}$ generation and outcome-grounded verification, followed by token length filtering with thresholds $l_{\textit{table}}$, $l_{\textit{constraint}}$.

\paragraph{Stage 4 on Sampling.}
From the filtered instances, we construct the following evaluation sets, all randomly sampled and adapted to our cross-representation learning instructions:

\begin{oursenumerate}
    \item \textbf{ChartCoder}~\citep{chartcoder2025benchmark} ($542$ instances): in-domain evaluation set adapted to our cross-representation understanding tasks, non-overlapping with the training set.

    \item \textbf{ChartNet}~\citep{2026chartnet} ($542$ instances): out-of-domain evaluation set adapted to our cross-representation understanding tasks, covering different domains (e.g., health, finance, etc.) and visualization packages (e.g., \texttt{matplotlib}, \texttt{plotly}, etc.)

    \item \textbf{ChartMimic}~\citep{chartmimic2025benchmark} ($1084$ instances): evaluation set adapted to our cross-representation understanding tasks, comprising $542$ in-domain cross-representation understanding tasks and $542$ out-of-domain chart modification instances covering diverse scientific domains, such as physics, mathematics, economics, biology, etc.

    \item \textbf{Chart2Code}~\citep{chart2code2025benchmark} ($1{,}243$ instances): evaluation set adapted to four tasks covering multi-level chart complexity, including cross-representation understanding (\textit{in-domain}), text-based chart modification (\textit{out-of-domain}), table-based chart modification (\textit{out-of-domain}), and figure-based chart modification (\textit{out-of-domain}).
\end{oursenumerate}

\noindent All token length thresholds $l_{\textit{input}}$, $l_{\textit{code}}$, $l_{\textit{table}}$, and $l_{\textit{constraint}}$, together with sampling configurations, are summarized in \S\ref{appendix:subsec:configuration} and Tab.~\ref{tab:configuration}.

\section{\ourstrain: Consistency-Driven Co-Evolution}
\label{appendix:sec:cocoevolve_at_train_details}

In the co-evolution objective (Eq.~\ref{eq:cocoevolve_objective}), each surrogate is a token-level clipped-ratio loss with KL penalty:
\begin{align}
\mathcal{L}_{\theta}^{(i)} &=
    \frac{1}{|o_{\theta}^{(i)}|}\sum_{\tau}\min\!\Big(
        \rho_{\theta,\tau}^{(i)} A_{\theta}^{(i)},\;
        \mathrm{clip}( \rho_{\theta,\tau}^{(i)},
        1{-}\epsilon, 1{+}\epsilon)
        A_{\theta}^{(i)}\Big) - \beta\,\mathbb{D}_{\mathrm{KL}}\!\big[
        \pi_{\theta}\,\|\,\pi_{\theta}^{\mathrm{ref}}\big] \\
\mathcal{L}_{\psi}^{(i,k)} &=
    \frac{1}{|o_{\psi}^{(i,k)}|}\sum_{\tau}\min\!\Big(
        \rho_{\psi,\tau}^{(i,k)} A_{\psi}^{(i,k)},\;
        \mathrm{clip}( \rho_{\psi,\tau}^{(i,k)},
        1{-}\epsilon, 1{+}\epsilon)
        A_{\psi}^{(i,k)}\Big) - \beta\,\mathbb{D}_{\mathrm{KL}}\!\big[
        \pi_{\psi}\,\|\,\pi_{\psi}^{\mathrm{ref}}\big]
\end{align}
The coupling is carried by the advantages. Let $r_{\theta}(o_{\theta}^{(i)}, o_{\psi}^{(i,k)}; v)$ and $r_{\psi}(o_{\theta}^{(i)}, o_{\psi}^{(i,k)}; v)$ denote the paired reward functionals defined in Eq.\ref{eq:hierarchical_reward}, where $R_{\theta}^{(i,k)} = r_{\theta}(o_{\theta}^{(i)}, o_{\psi}^{(i,k)}; v)$ and $R_{\psi}^{(i,k)} = r_{\psi}(o_{\theta}^{(i)}, o_{\psi}^{(i,k)}; v)$ denote the corresponding scalar rewards. The $\pi_{\theta}$ advantage marginalizes over $\pi_{\psi}$ noise by averaging across its $K_{\psi}$ children of each $\pi_\theta$ rollout:
\begin{align}
\bar{r}_{\theta}^{(i)} &= \frac{1}{K_{\psi}}\sum_{k=1}^{K_{\psi}} 
    r_{\theta}\!\big(o_{\theta}^{(i)}, o_{\psi}^{(i,k)}; v\big)
\label{eq:reward_theta} \\
A_{\theta}^{(i)} &= \frac{\bar{r}_{\theta}^{(i)} - \mu_{\theta}(v)}
    {\sigma_{\theta}(v) + \delta}
\label{eq:advantage_theta}
\end{align}
so that credit assignment to $\pi_{\theta}$ reflects expected cross-model quality rather than a single $\pi_{\psi}$ draw. The $\pi_{\psi}$ advantage is computed within the $K_{\psi}$-sibling group conditioned on the same $\pi_{\theta}$ parent:
\begin{align}
A_{\psi}^{(i,k)} = \frac{r_{\psi}\!\big(o_{\theta}^{(i)}, 
    o_{\psi}^{(i,k)}; v\big) - \mu_{\psi}^{(i)}}
    {\sigma_{\psi}^{(i)} + \delta}
\label{eq:advantage_psi}
\end{align}
where $(\mu_{\theta}(v), \sigma_{\theta}(v))$ are the mean and standard deviation of $\{\bar{r}_{\theta}^{(i)}\}_{i=1}^{K_{\theta}}$, $(\mu_{\psi}^{(i)}, \sigma_{\psi}^{(i)})$ those of $\{r_{\psi}(o_{\theta}^{(i)}, o_{\psi}^{(i,k)}; v)\}_{k=1}^{K_{\psi}}$, $\rho_{\theta,\tau}^{(i)} = \pi_{\theta}(o_{\theta,\tau}^{(i)} \mid v, o_{\theta,<\tau}^{(i)}) / \pi_{\theta,\text{old}}(\cdot)$ and $\rho_{\psi,\tau}^{(i,k)} = \pi_{\psi}(o_{\psi,\tau}^{(i,k)} \mid s^{(i)}, t^{(i)}, o_{\psi,<\tau}^{(i,k)}) / \pi_{\psi,\text{old}}(\cdot)$ are per-token importance ratios, and $\delta$ is a numerical stabilizer.

\section{Implementation Details}
\label{appendix:sec:implementation_details}

In Tabs.~\ref{tab:notation}-\ref{tab:configuration}, we summarize the core symbols, notations, and hyperparameter settings used in this paper. Below, we further explain the key design choices to facilitate reproducibility and adaptation to customized use cases.

\subsection{Experiment Configuration}
\label{appendix:subsec:configuration}

\paragraph{Three-Tier Notation.}
Our method involves two models interacting through a shared representation cycle, which requires careful notational separation between their functional roles, model identities, and training policies (Tab.~\ref{tab:notation}):
\begin{oursenumerate}
    \item $f_\theta$ and $g_\psi$ denote \textit{task mappings}: the core representational objectives each model is designed to perform. Note that $f_\theta$ formally describes the primary cross-representation objective without enumerating all outputs of $M_\theta$, whose full rollout $o_\theta^{(i)} = (\hat{s}^{(i)}, \hat{t}^{(i)}, \hat{c}_\theta^{(i)})$ additionally includes a constraint and a reference code for co-evolution.

    \item $M_\theta$ and $M_\psi$ denote \textit{model identities} instantiating these mappings: $M_\theta$ is a multimodal model and $M_\psi$ is a language model. They can be instantiated by either the same or different base models, depending on experimental settings (\S\ref{sec:experiments}).

    \item $\pi_\theta$ and $\pi_\psi$ denote the \textit{training policies} parameterizing $M_\theta$ and $M_\psi$ during RL: used specifically in the context of rollout sampling, importance ratio computation, advantage estimation, and policy updates.
\end{oursenumerate}
This three-tier design allows us to cleanly distinguish what each model \textit{does} ($f_\theta, g_\psi$), \textit{what} it \textit{is} ($M_\theta, M_\psi$), and \textit{how} it is \textit{optimized} ($\pi_\theta, \pi_\psi$).

\begin{wrapfigure}{r}{0.6\textwidth}
    \vspace{-20pt}

    \small
    \centering
    \includegraphics[width=0.6\textwidth]{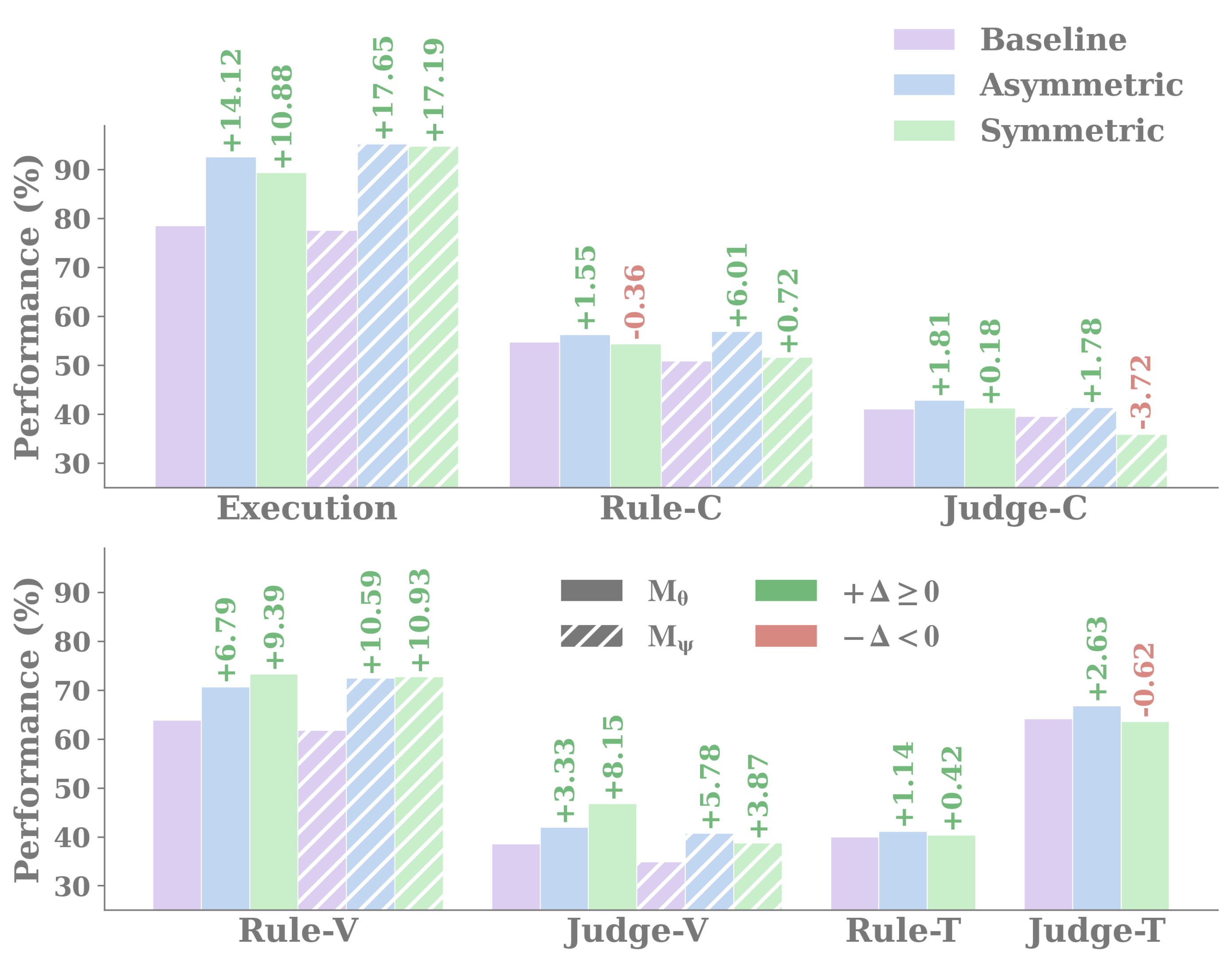}

    \vspace{-8pt}
    
    \caption{\textbf{Ablation Study on Reward Weighting.}}
    \label{fig:exp_ablation_reward_weighting}

    \vspace{-26pt}

\end{wrapfigure}

\paragraph{Asymmetric Reward Weighting.}
All reward weighting coefficients $\lambda_*^{(\pi)}$ and sub-weights $\omega_*^{(\pi)}$ are defined per-model, and their values in our main experiments are intentionally asymmetric between $M_\theta$ and $M_\psi$ (Tab.~\ref{tab:configuration}). This asymmetry reflects the fundamentally different roles the two models play in the representation cycle. Generating code, table, and constraint from a given chart, the primary challenges of $M_\theta$ lie in perceptual grounding and semantic understanding. We therefore assign $M_\theta$ a higher individual grounding weight $\lambda_\theta = 0.7$ while keeping its cross-model consistency weights lower ($\lambda_c^{(\theta)} = 0.05$, $\lambda_v^{(\theta)} = 0.05$), ensuring that $M_\theta$'s training signal is dominated by how well its understanding matches the original chart $v$, rather than how well it agrees with $M_\psi$. By contrast, $M_\psi$ performs single-modal understanding, with its primary challenge being the faithful translation of constrained table data into the chart it originally corresponds to. As such, cross-model consistency is the primary grounding signal of$M_\psi$. We therefore assign $M_\psi$ higher cross-model consistency weights ($\lambda_c^{(\psi)} = 0.2$, $\lambda_v^{(\psi)} = 0.2$) to reflect this. The same asymmetry applies to the code consistency sub-weights $\omega_e^{(\pi)}$ and $\omega_s^{(\pi)}$: $M_\theta$ is assigned higher execution weight ($\omega_e^{(\theta)} = 0.9$) since binary execution success is a stronger and more direct signal for its cross-representation learning objectives, while $M_\psi$ receives a more balanced split ($\omega_e^{(\psi)} = 0.7$, $\omega_s^{(\psi)} = 0.3$) that additionally rewards semantic code similarity. Importantly, these asymmetric weights also serve as an anti-collusion mechanism to avoid reward hacking and training crash: by assigning $M_\theta$ stronger individual grounding rewards relative to cross-model consistency rewards, we prevent the degenerate solution where both models collude to produce mutually consistent but semantically incorrect outputs (Fig.~\ref{fig:exp_ablation_reward_weighting}).

\begin{figure*}[!t]
    \vspace{0pt}

    \small
    \centering
    \includegraphics[width=1.0\textwidth]{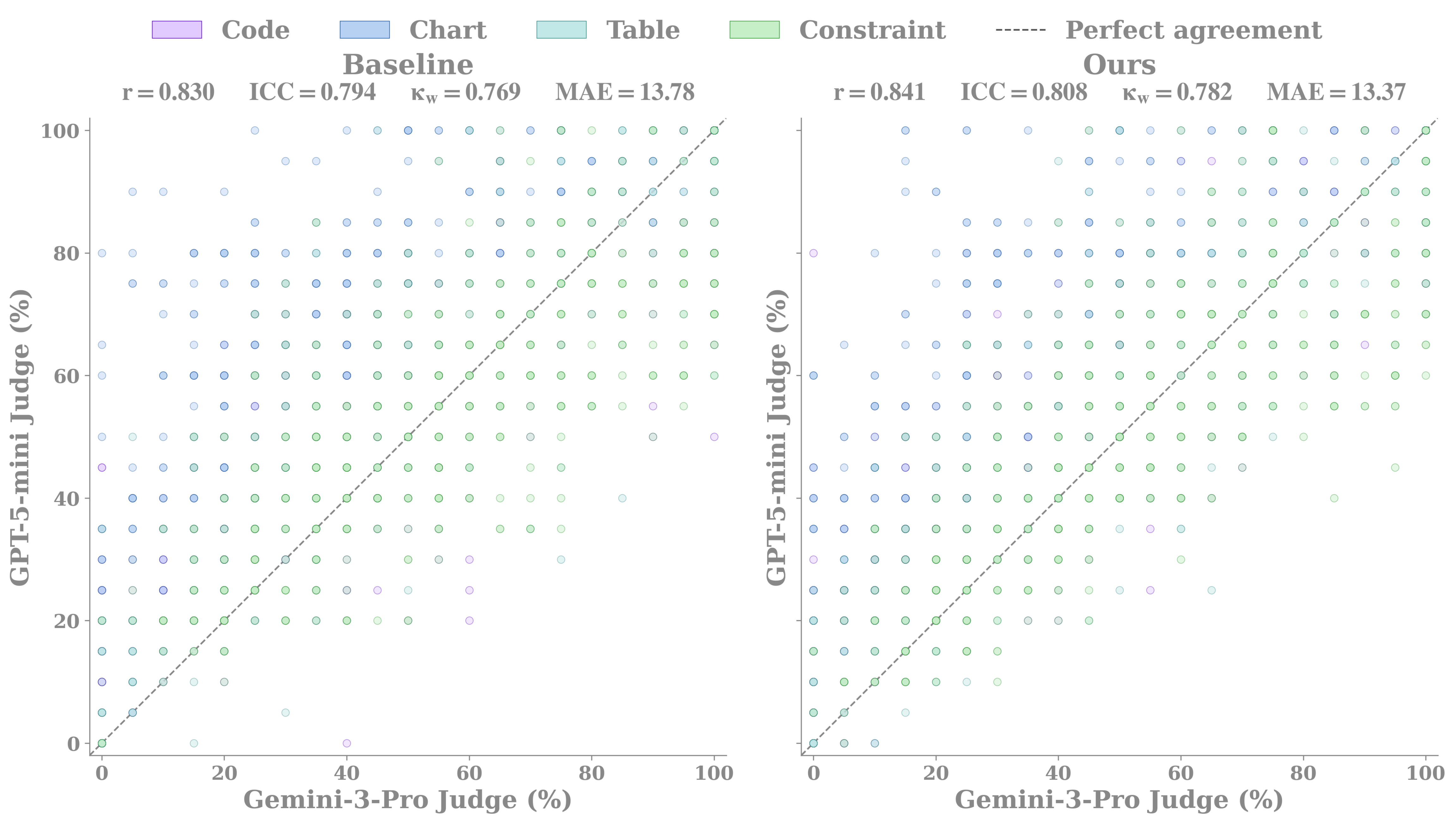}

    \vspace{6pt}
    
    \caption{\textbf{Interrater Agreement.} We calculate interrater agreement between two LLM- and MLLM- judges. Results suggest close alignment between two judges, revealing the effectiveness of \ourseval for cross-representation evaluation.}
    \label{fig:interrater_agreement}

    \vspace{0pt}

\end{figure*}

\paragraph{Weighting Coefficient Families.}
We use three distinct families of weighting coefficients for different stages of \ours, each serving a specialized purpose:
\begin{oursenumerate}
    \item $\omega_*^{(\pi)}$ denotes sub-weights \textit{within} a reward component, balancing the contribution of each constituent metric.

    \item $\lambda_*^{(\pi)}$ denotes per-model weights combining heterogeneous reward components into the final reward $R_\pi$ during training, with asymmetric configurations across $M_\theta$ and $M_\psi$ reflecting their distinct roles in tackling the representation cycle (\S\ref{subsec:ours_at_train_time}).

    \item $\gamma_*$ denotes weights combining individual evaluation metrics into final composite scores during evaluation, enabling comprehensive multidimensional assessment across all six cross-representation tasks (\S\ref{subsec:method:evaluation_metrics}).
\end{oursenumerate}
The three families are intentionally distinguished by symbol to make their scope, \textit{i.e.}, intra-component, inter-component training, and evaluation, immediately clear at each point of use. Their detailed configurations are summarized in Tab.~\ref{tab:configuration}.

\subsection{Computation Overhead}
\label{appendix:subsec:computation_overhead}

We summarize the computation overhead of \ours across main training configurations and evaluation in Tab.~\ref{tab:computation_overhead}, covering GPU resources, API cost, and wall-clock time.

\paragraph{Training.}
All training configurations run on H100 (96GB) GPUs with no API cost, as \ours relies exclusively on consistency-driven rewards computed via deterministic sandbox execution and lightweight embedding models during training. Compared to the Baseline, \ourstrain ($\mathcal{M}$ disabled) reduces wall-clock time, owing to the more efficient co-evolution objective as compared to sequential cross-representation learning. Enabling $\mathcal{M}$ further reduces training time to 71h (no annealing) and 72h (with annealing), as teacher-guided grounding accelerates early-stage policy convergence and reduces the number of low-quality rollouts that would otherwise contribute noisy reward signals.

\paragraph{Evaluation.}
All evaluation configurations similarly run on H100 (96GB) GPUs. The API cost decreases substantially across methods: baseline evaluation incurs \$0.8637 per sample due to sequential generation and rollouts, while \ours reduces the cost through test-time consistency-driven co-optimization (\S\ref{subsec:ours_at_test_time}) by selecting the best candidates via annotation-free consistency signals. Wall-clock time follows the same trend, demonstrating that \ours serves as an effective annotation-free optimization approach with reduced API cost and wall-clock time at test time.


\section{\ourseval: Evaluation Suite for Multidimensional Assessment}
\label{sec:appendix:ours_eval_suite}

Given the limitations of existing LLM-as-Judge approaches (\S\ref{subsec:method:evaluation_metrics}), we propose \ourseval, a systematic evaluation suite consisting of three complementary evaluation approaches across 27 metrics, covering all six cross-representation tasks in the representation cycle (\S\ref{subsec:problem_formulation}): \textit{rule-based judge} (\S\ref{subsec:appendix:ours_eval_suite:rule_judge}), \textit{LLM-as-Judge} (\S\ref{subsec:appendix:ours_eval_suite:llm_judge}), and \textit{MLLM-as-Judge} (\S\ref{subsec:appendix:ours_eval_suite:mllm_judge}).

\subsection{Rule-as-Judge Evaluation}
\label{subsec:appendix:ours_eval_suite:rule_judge}

\subsubsection{Rule-as-Judge Chart Evaluation}
\label{subsubsec:appendix:ours_eval_suite:rule_judge:chart}

To achieve comprehensive chart evaluation, our rule-as-judge chart evaluation assesses four complementary dimensions: structural accuracy $F_{\textit{ssim}}$ (Eq.~\ref{eq:rule_as_judge:chart:ssim}), semantic accuracy $F_{\textit{clip}}$ (Eq.~\ref{eq:rule_as_judge:chart:clip}), perceptual accuracy $F_{\textit{dino}}$ (Eq.~\ref{eq:rule_as_judge:chart:dino}), and textual accuracy $F_{\textit{ocr}}$ (Eq.~\ref{eq:rule_as_judge:chart:ocr}). Concretely, given a model-predicted chart image $\hat{v}$ and a reference chart image $v$, each metric targets a distinct dimension of chart quality:

\paragraph{Structural Accuracy.}
$F_{\textit{ssim}}$ (Eq.~\ref{eq:rule_as_judge:chart:ssim}) measures pixel-level structural similarity between $\hat{v}$ and $v$ via the structural similarity index~\citep{wang2004ssim}:
\begin{equation}
    F_{\textit{ssim}}(\hat{v}, v) = \texttt{SSIM}(\hat{v}, v) \in [0, 1]
    \label{eq:rule_as_judge:chart:ssim}
\end{equation}
where $F_{\textit{ssim}}$ captures local luminance, contrast, and structural patterns that reflect the layout and compositional fidelity of predicted charts.

\paragraph{Semantic Accuracy.}
$F_{\textit{clip}}$ (Eq.~\ref{eq:rule_as_judge:chart:clip}) measures semantic visual similarity between $\hat{v}$ and $v$ via normalized CLIP~\citep{radford2021clip} embeddings:
\begin{equation}
    F_{\textit{clip}}(\hat{v}, v) = 
    \frac{\cos(e_{\textit{clip}}(\hat{v}),\, e_{\textit{clip}}(v)) + 1}{2} 
    \in [0, 1]
    \label{eq:rule_as_judge:chart:clip}
\end{equation}
where $e_{\textit{clip}}(\cdot)$ denotes the visual embedding functions of CLIP~\citep{radford2021clip}, and thus $F_{\textit{clip}}$ captures high-level semantic alignment between the predicted and reference charts.

\paragraph{Perceptual Accuracy.}
$F_{\textit{dino}}$ (Eq.~\ref{eq:rule_as_judge:chart:dino}) measures fine-grained perceptual similarity between $\hat{v}$ and $v$ via normalized DINOv2~\citep{oquab2024dinov2} embeddings:
\begin{equation}
    F_{\textit{dino}}(\hat{v}, v) = 
    \frac{\cos(e_{\textit{dino}}(\hat{v}),\, e_{\textit{dino}}(v)) + 1}{2} 
    \in [0, 1]
    \label{eq:rule_as_judge:chart:dino}
\end{equation}
where $e_{\textit{dino}}(\cdot)$ denotes the visual embedding functions of DINOv2~\citep{oquab2024dinov2}, and thus $F_{\textit{dino}}$ captures spatial and feature-level visual fidelity beyond semantic-level alignment.

\paragraph{Textual Accuracy.}
$F_{\textit{ocr}}$ (Eq.~\ref{eq:rule_as_judge:chart:ocr}) measures text-level fidelity between $\hat{v}$ and $v$ by calculating token-level $F_1$ over OCR-extracted text via EasyOCR~\citep{easyocr}:
\begin{equation}
    F_{\textit{ocr}}(\hat{v}, v) = 
    F_1(\texttt{OCR}(\hat{v}),\, \texttt{OCR}(v)) \in [0, 1]
    \label{eq:rule_as_judge:chart:ocr}
\end{equation}
where $F_{\textit{ocr}}$ captures the textual accuracy of visual elements, such as axis labels, legends, titles, data annotations, etc.

\paragraph{Final Score.}
Based on these four complementary dimensions, the final rule-as-judge chart evaluation score $F_{\textit{chart}}$ is the weighted combination of four dimensions:
\begin{equation}
    F_{\textit{chart}}(\hat{v}, v) = \sum_{m} \gamma_{m} \cdot F_{m}(\hat{v}, v)
    \label{eq:rule_as_judge:chart_avg}
\end{equation}
where $m \in \{\textit{clip}, \textit{ssim}, \textit{ocr}, \textit{dino}\}$, with $\gamma_m \in \{\gamma_{\textit{clip}}, \gamma_{\textit{ssim}}, \gamma_{\textit{ocr}}, \gamma_{\textit{dino}}\}$ as evaluation weighting coefficients. Our detailed configurations are summarized in \S\ref{appendix:subsec:configuration} and Tab.~\ref{tab:configuration}.

\subsubsection{Rule-as-Judge Code Evaluation}
\label{subsubsec:appendix:ours_eval_suite:rule_judge:code}

For more comprehensive code evaluation, our rule-as-judge code evaluation assesses six complementary dimensions: code executability $F_{\textit{exec}}$ (Eq.~\ref{eq:rule_as_judge:code:exec}), code quality $F_{\textit{codebleu}}$ (Eq.~\ref{eq:rule_as_judge:code:codebleu}), structural accuracy $F_{\textit{ast}}$ (Eq.~\ref{eq:rule_as_judge:code:ast}), lexical accuracy $F_{\textit{cosine}}$ (Eq.~\ref{eq:rule_as_judge:code:tfidf}), contextual accuracy $F_{\textit{codebert}}$, and semantic accuracy $F_{\textit{unixcoder}}$. Concretely, given a model-predicted code $\hat{c}$ and a reference code $c$, each metric targets a distinct dimension of code quality:

\paragraph{Code Executability.}
$F_{\textit{exec}}$ measures whether the predicted code executes successfully via the deterministic sandbox executor $h$ (Eq.~\ref{eq:cycle}):
\begin{equation}
    F_{\textit{exec}}(\hat{c}) = \mathds{1}[h(\hat{c})] \in \{0, 1\}
    \label{eq:rule_as_judge:code:exec}
\end{equation}
We report $F_{\textit{exec}}$ as a standalone execution success rate metric to complement the final code evaluation score $F_{\textit{code}}$ (Eq.~\ref{eq:rule_as_judge:code_avg}).

\paragraph{Code Quality.}
$F_{\textit{codebleu}}$ measures overall code similarity via CodeBLEU~\citep{ren2020codebleu}, computed as a weighted combination of four complementary dimensions:
\begin{equation}
    F_{\textit{codebleu}}(\hat{c}, c) = \sum_{m} \gamma_{m} \cdot F_{m}(\hat{c}, c)
    \label{eq:rule_as_judge:code:codebleu}
\end{equation}
where $m \in \{\textit{ngram}, \textit{wngram}, \textit{syntax}, \textit{dataflow}\}$, with $F_{\textit{ngram}}$ as $n$-gram match, $F_{\textit{wngram}}$ as weighted $n$-gram match, $F_{\textit{syntax}}$ as syntax tree match, and $F_{\textit{dataflow}}$ as data-flow graph match.

\paragraph{Structural Accuracy.}
$F_{\textit{ast}}$ measures structural similarity between $\hat{c}$ and $c$ via Abstract Syntax Tree (AST) sequence matching:
\begin{align}
    F_{\textit{ast}}(\hat{c}, c) &= 
    \texttt{SequenceMatcher}(\texttt{AST}(\hat{c}),\, \texttt{AST}(c)) \nonumber \\
    &\hspace{1em}\in [0, 1]
    \label{eq:rule_as_judge:code:ast}
\end{align}
where $\texttt{AST}(\cdot)$ denotes the depth-first node label sequence of the parsed syntax tree, and $\texttt{SequenceMatcher}$ computes the longest common subsequence (LCS) ratio.

\paragraph{Lexical Accuracy.}
$F_{\textit{cosine}}$ measures token-level lexical similarity between $\hat{c}$ and $c$ via weighted cosine similarity over code tokens:
\begin{equation}
    F_{\textit{cosine}}(\hat{c}, c) = 
    \frac{\mathbf{w}(\hat{c}) \cdot \mathbf{w}(c)}
    {\|\mathbf{w}(\hat{c})\| \cdot \|\mathbf{w}(c)\|} \in [0, 1]
    \label{eq:rule_as_judge:code:tfidf}
\end{equation}
where $\mathbf{w}(\cdot)$ denotes the TF-IDF~\citep{Salton1988TFIDF} weighted token vector over the vocabulary of identifier, number, and string tokens extracted via code tokenization.

\paragraph{Contextual Accuracy.}
$F_{\textit{codebert}}$ measures contextual similarity between $\hat{c}$ and $c$ via CodeBERT~\citep{feng2020codebert}, capturing token-level contextual code representations by following the normalized embedding similarity formulation of $\texttt{sim}_{\textit{c}}$ (Eq.~\ref{eq:unixcoder}).

\paragraph{Semantic Accuracy.}
$F_{\textit{unixcoder}}$ measures semantic similarity between $\hat{c}$ and $c$ via the unified cross-modal encoder of UniXcoder~\citep{guo2022unixcoder}, capturing deeper code-to-code semantic correspondence by following the normalized embedding similarity 
formulation of $\texttt{sim}_{\textit{c}}$ (Eq.~\ref{eq:unixcoder}).

\paragraph{Final Score.}
The final Rule-as-Judge code evaluation score $F_{\textit{code}}$ is the weighted combination across five dimensions:
\begin{equation}
    F_{\textit{code}}(\hat{c}, c) = \sum_{m} \gamma_{m} \cdot F_{m}(\hat{c}, c)
    \label{eq:rule_as_judge:code_avg}
\end{equation}
where $m \in \{\textit{codebleu}, \textit{ast}, \textit{cosine}, \textit{codebert}, \textit{unixcoder}\}$, with $F_{\textit{exec}}$ (Eq.~\ref{eq:rule_as_judge:code:exec}) reported separately as a standalone executability score (Tab.~\ref{tab:exp:exp_main_code_evaluation}). Our detailed configurations are summarized in \S\ref{appendix:subsec:configuration} and Tab.~\ref{tab:configuration}.

\subsubsection{Rule-as-Judge Table Evaluation}
\label{subsubsec:appendix:ours_eval_suite:rule_judge:table}

Our rule-as-judge table evaluation assesses two complementary dimensions: schema accuracy and value accuracy. Concretely, given a model-predicted table $\hat{t}$ and a reference table $t$, each metric targets a distinct dimension of table quality:

\paragraph{Schema Accuracy.}
$F_{\textit{schema}}$ measures column-level F1 between $\hat{t}$ and $t$, directly instantiating the schema accuracy formulation of $\mathbf{F}_{\textit{t}}$ (Eq.~\ref{eq:table_reward}).

\paragraph{Value Accuracy.}
$F_{\textit{value}}$ measures cell-level F1 over all $(\textit{row}, \textit{column}, \textit{value})$ triples between $\hat{t}$ and $t$, directly instantiating the value accuracy formulation of $\mathbf{F}_{\textit{t}}$ (Eq.~\ref{eq:table_reward}).

\paragraph{Final Score.}
The final rule-as-judge table evaluation score $F_{\textit{table}}$ is the weighted combination across two dimensions:
\begin{equation}
    F_{\textit{table}}(\hat{t}, t) = \sum_{m} \gamma_{m} \cdot F_{m}(\hat{t}, t)
    \label{eq:rule_as_judge:table_avg}
\end{equation}
where $m \in \{\textit{schema}, \textit{value}\}$. Our detailed configurations are summarized in \S\ref{appendix:subsec:configuration} and Tab.~\ref{tab:configuration}.

\subsubsection{Rule-as-Judge Constraint Evaluation}
\label{subsubsec:appendix:ours_eval_suite:rule_judge:constraint}

Our rule-as-judge constraint evaluation assesses two complementary dimensions: semantic accuracy and lexical accuracy. Given a model-predicted constraint $\hat{s}$ and a reference constraint $s_{gt}$, each metric targets a distinct dimension of constraint quality:

\paragraph{Semantic Accuracy.}
$F_{\textit{ssem}}=\mathbf{F}_{\textit{s}}$ measures semantic similarity between $\hat{s}$ and $s_{gt}$ via sentence-level cosine similarity, directly instantiating the formulation of $\mathbf{F}_{\textit{s}}$ (Eq.~\ref{eq:constraint_reward}).

\paragraph{Lexical Accuracy.}
$F_{\textit{rouge}}$ measures lexical similarity between $\hat{s}$ and $s_{gt}$ via ROUGE-L~\citep{lin2004rouge}:
\begin{equation}
    F_{\textit{rouge}}(\hat{s}, s_{gt}) = 
    F_{\textit{LCS}}(\hat{s}, s_{gt}) \in [0, 1]
    \label{eq:rule_as_judge:constraint:rouge}
\end{equation}
where $F_{\textit{LCS}}$ denotes the F1 score computed over the longest common subsequence of tokens between the predicted and reference constraints.

\paragraph{Final Score.}
The final rule-as-judge constraint evaluation score $F_{\textit{constraint}}$ is the weighted combination across two dimensions:
\begin{equation}
    F_{\textit{constraint}}(\hat{s}, s_{gt}) = \sum_{m} \gamma_{m} \cdot F_{m}(\hat{s}, s_{gt})
    \label{eq:rule_as_judge:constraint_avg}
\end{equation}
where $m \in \{\textit{ssem}, \textit{rouge}\}$. Our detailed configurations are summarized in \S\ref{appendix:subsec:configuration} and Tab.~\ref{tab:configuration}.

\subsection{LLM-as-Judge Evaluation}
\label{subsec:appendix:ours_eval_suite:llm_judge}

Existing LLM-as-judge approaches for code evaluation collapse complex semantic and structural fidelity into a single and/or wide-range score, which LLMs are known to judge inconsistently and unreliably (\S\ref{subsec:method:evaluation_metrics}). To address these limitations, our \ourseval proposes LLM-as-judge evaluation that decomposes code quality into five fine-grained, independently scored dimensions, each targeting a distinct and non-overlapping dimension of visualization code quality (Fig.~\ref{fig:eval_prompt:code_llm_as_judge}). Concretely, given the ground-truth code $c$ and predicted code $\hat{c}$, the judge reasons about the expected rendered output via static code analysis, scoring each dimension on a $0$--$5$ integer scale, and then normalizing to $[0, 1]$.

\begin{oursenumerate}
    \item \textbf{Data Correctness}: correctness of underlying data values, categories, transformations, and their correspondence to the intended chart structure.
    
    \item \textbf{Chart Type Accuracy}: correctness of chart type, subtype, dimensionality, and geometry (\textit{e.g.}, grouped vs.\ stacked bars, single vs.\ multiple lines).
    
    \item \textbf{Structural Fidelity}: correctness of layout, subplot configuration, axes, labels, titles, annotations, scales, ticks, legends, and grid configuration.
    
    \item \textbf{Visual Accuracy}: correctness of data-to-visual encoding mappings, including positions, heights, trends, grouping, stacking, and category alignment.
    
    \item \textbf{Style Accuracy}: correctness of colors, colormaps, marker types, sizes, transparency, line widths, fills, and other aesthetic details of each chart element.
\end{oursenumerate}

\noindent The final LLM-as-Judge code evaluation score $J_{\textit{code}}$ is the weighted combination across five dimensions:
\begin{equation}
    J_{\textit{code}}(\hat{c}, c) = \sum_{m} \gamma_{m} \cdot J_{m}(\hat{c}, c)
    \label{eq:llm_judge:code_avg}
\end{equation}
where $m \in \{\textit{Cdata}$$, $$\textit{Ctype}$, $\textit{Cstruct}$, $\textit{Cvisual}$, $\textit{Cstyle}\}$. Our detailed configurations are summarized in \S\ref{appendix:subsec:configuration} and Tab.~\ref{tab:configuration}.

\subsection{MLLM-as-Judge Evaluation}
\label{subsec:appendix:ours_eval_suite:mllm_judge}

Similar to LLM-as-judge, existing MLLM-as-judge approaches collapse complex visual and structural fidelity into a single wide-range score, which MLLMs are known to judge unreliably and inconsistently (\S\ref{subsec:method:evaluation_metrics}).For more reliable MLLM judgments, our \ourseval decomposes each evaluation into five fine-grained, independently scored dimensions, each targeting a distinct and non-overlapping aspect of model prediction quality, scoring each dimension on a $0$--$5$ integer scale normalized to $[0, 100]$. \ourseval includes three MLLM-as-judge evaluators: chart evaluation (\S\ref{subsubsec:appendix:ours_eval_suite:mllm_judge:chart}), table evaluation (\S\ref{subsubsec:appendix:ours_eval_suite:mllm_judge:table}), and constraint evaluation (\S\ref{subsubsec:appendix:ours_eval_suite:mllm_judge:constraint}), respectively.

\subsubsection{MLLM-as-Judge Chart Evaluation}
\label{subsubsec:appendix:ours_eval_suite:mllm_judge:chart}

Given the ground-truth chart image $v$ and predicted chart image $\hat{v}$, the judge directly compares the two rendered images across five dimensions, as detailed in Fig.~\ref{fig:eval_prompt:chart_mllm_as_judge}:

\begin{oursenumerate}
    \item \textbf{Chart Type Accuracy}: correctness of chart type and subtype, including exact match of chart geometry and dimensionality (e.g., grouped vs.\ stacked bars, pie vs.\ donut).

    \item \textbf{Data Accuracy}: accuracy of data values, proportions, magnitudes, and precisions as visually perceived in the predicted chart.

    \item \textbf{Text Accuracy}: correctness of all text elements, including titles, axis labels, tick values, legend entries, and annotations, together with their positions and formatting.

    \item \textbf{Style Accuracy}: correctness of visual styles of chart elements, including colors, colormaps, markers, line styles, fills, and their assignment to the correct data series.

    \item \textbf{Visual Accuracy}: correctness of overall visual layout, spatial relationships, and structural details, including figure proportions, subplot arrangement, spacing, stacking order, category ordering, trends, and decorative details such as gridlines, error bars, and reference lines.
\end{oursenumerate}

\noindent The final MLLM-as-judge chart evaluation score $J_{\textit{chart}}$ is the weighted combination across five dimensions:
\begin{equation}
    J_{\textit{chart}}(\hat{v}, v) = \sum_{m} \gamma_{m} \cdot J_{m}(\hat{v}, v)
    \label{eq:mllm_judge:chart_avg}
\end{equation}
where $m \in \{\textit{Vtype}, \textit{Vdata}, \textit{Vtext}, \textit{Vstyle}, \textit{Vvisual}\}$. Our detailed configurations are summarized in \S\ref{appendix:subsec:configuration} and Tab.~\ref{tab:configuration}.

\subsubsection{MLLM-as-Judge Table Evaluation}
\label{subsubsec:appendix:ours_eval_suite:mllm_judge:table}

Given the ground-truth table $t$, predicted table $\hat{t}$, ground-truth code $c$, and ground-truth chart image $v$ as reference context, the judge evaluates the predicted table across five dimensions, as detailed in Fig.~\ref{fig:eval_prompt:table_mllm_as_judge}:

\begin{oursenumerate}
    \item \textbf{Schema Accuracy}: correctness of column headers and table structure, penalizing missing, extra, or incorrectly named columns.

    \item \textbf{Data Accuracy}: correctness of all cell values compared to ground truth, penalizing hallucinated, incorrect, or miscomputed values.

    \item \textbf{Data Coverage}: completeness of extracted data relative to ground truth, penalizing missing rows or partially captured data.

    \item \textbf{Row Alignment}: correctness of row ordering and alignment of values across all columns.

    \item \textbf{Format Correctness}: validity of table structure, including correct column-keyed format, string-typed values, equal-length columns, and correct representation of missing values.
\end{oursenumerate}

\noindent The final MLLM-as-judge table evaluation score $J_{\textit{table}}$ is the weighted combination across five dimensions:
\begin{equation}
    J_{\textit{table}}(\hat{t}, t) = \sum_{m} \gamma_{m} \cdot J_{m}(\hat{t}, t)
    \label{eq:mllm_judge:table_avg}
\end{equation}
where $m \in \{\textit{Tschema}$, $\textit{Tdata}$, $\textit{Tcoverage}, \textit{Trow}$, $\textit{Tformat}\}$. Our detailed configurations are summarized in \S\ref{appendix:subsec:configuration} and Tab.~\ref{tab:configuration}.

\subsubsection{MLLM-as-Judge Constraint Evaluation}
\label{subsubsec:appendix:ours_eval_suite:mllm_judge:constraint}

Given the ground-truth constraint $s_{gt}$, predicted constraint $\hat{s}$, ground-truth code $c$, and ground-truth chart image $v$ as reference context, the judge evaluates the predicted constraint across five dimensions, as detailed in Fig.~\ref{fig:eval_prompt:constraint_mllm_as_judge}:

\begin{oursenumerate}
    \item \textbf{Visual Fidelity}: correctness of all key visual contents and data values described in $\hat{s}$ relative to $s_{gt}$, penalizing missing, incorrect, or hallucinated contents and values.

    \item \textbf{Visual Completeness}: completeness of all important visual elements described in $\hat{s}$, including axes, labels, legends, trends, categories, and colors, penalizing missing key components needed for chart reconstruction.

    \item \textbf{Structural Clarity}: quality of organization of $\hat{s}$ in support of chart reproduction, including clarity of relationships between visual elements.

    \item \textbf{Trend and Relationship Accuracy}: correctness of described trends, comparisons, and relationships between data elements, penalizing incorrect interpretations such as wrong directional trends.

    \item \textbf{Style Accuracy}: correctness and completeness of described plotting colors and styles, penalizing incorrect or missing style descriptions.
\end{oursenumerate}

\noindent The final MLLM-as-judge constraint evaluation score $J_{\textit{constraint}}$ is the weighted combination across five dimensions:
\begin{equation}
    J_{\textit{constraint}}(\hat{s}, s_{gt}) = \sum_{m} \gamma_{m} \cdot J_{m}(\hat{s}, s_{gt})
    \label{eq:mllm_judge:constraint_avg}
\end{equation}
where $m \in \{\textit{Sfidelity}$, $\textit{Scompleteness}$, $\textit{Sclarity}$, $\textit{Strend}$, $\textit{Sstyle}\}$. Our detailed configurations are summarized in \S\ref{appendix:subsec:configuration} and Tab.~\ref{tab:configuration}.

\section{Analysis on Cross-Representation Failures}
\label{appendix:sec:failure_analysis}

\subsection{Quantitative Analysis on Cross-Representation Failures}
\label{appendix:subsec:failure_analysis:quantitative}

Leveraging five-dimensional evaluation for each representation (\S\ref{subsec:method:evaluation_metrics}), Fig.~\ref{fig:failure_analysis} presents the failure distribution of cross-representation understanding for the baseline, \ourstrain, and \ourstrain{}\textbf{\textcolor{oursttcolor}{+@Test}}. As the results show, \ours substantially outperforms the non-finetuned base model, achieving an overall improvement of up tp $\Delta=\uparrow35.10\%$. This gain is primarily accompanied by notable reductions in data-related errors across different representations, while the remaining categories of failures are also consistently alleviated. This finding underscores the importance of data accuracy and information completeness for cross-representation understanding success. Furthermore, applying \ourstest leads to additional reductions across all five failure dimensions of each representation. Its resulting correctness improvement of up to $\Delta=\uparrow41.38\%$ further validates the effectiveness of \ours in optimizing model cross-representation understanding during both training and testing.

\subsection{Qualitative Analysis on Cross-Representation Failures}
\label{appendix:subsec:failure_analysis:qualitative}

\begin{figure*}[!t]
    \vspace{0pt}

    \small
    \centering
    \includegraphics[width=1.0\textwidth]{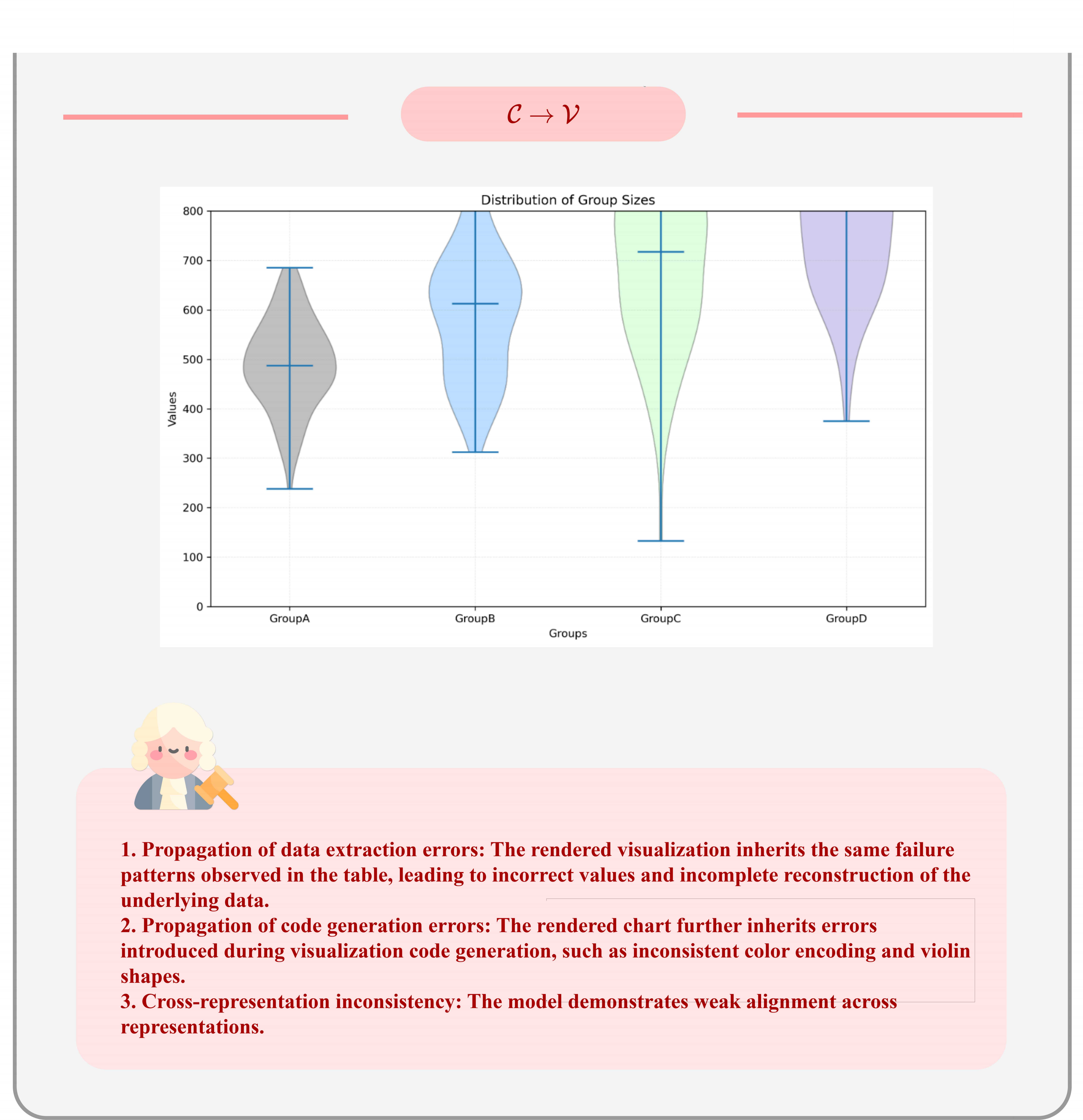}
    
    \caption{\textbf{Cross-Representation Understanding Failure Example of \texttt{Qwen3-4B-VL}.} A case study on \texttt{Qwen3-4B-VL} cross-representation understanding failures, illustrating how small-size models produce data recognition errors that accumulate and propagate across representations, resulting in inaccurate data extraction, incomplete information transfer, and significant cross-representation inconsistencies. Constraints are omitted for clarity.}
    \label{fig:failure1_qwen3}

    \vspace{0pt}

\end{figure*}

\begin{figure*}[!t]
    \vspace{-48pt}

    \small
    \centering
    \includegraphics[width=1.0\textwidth]{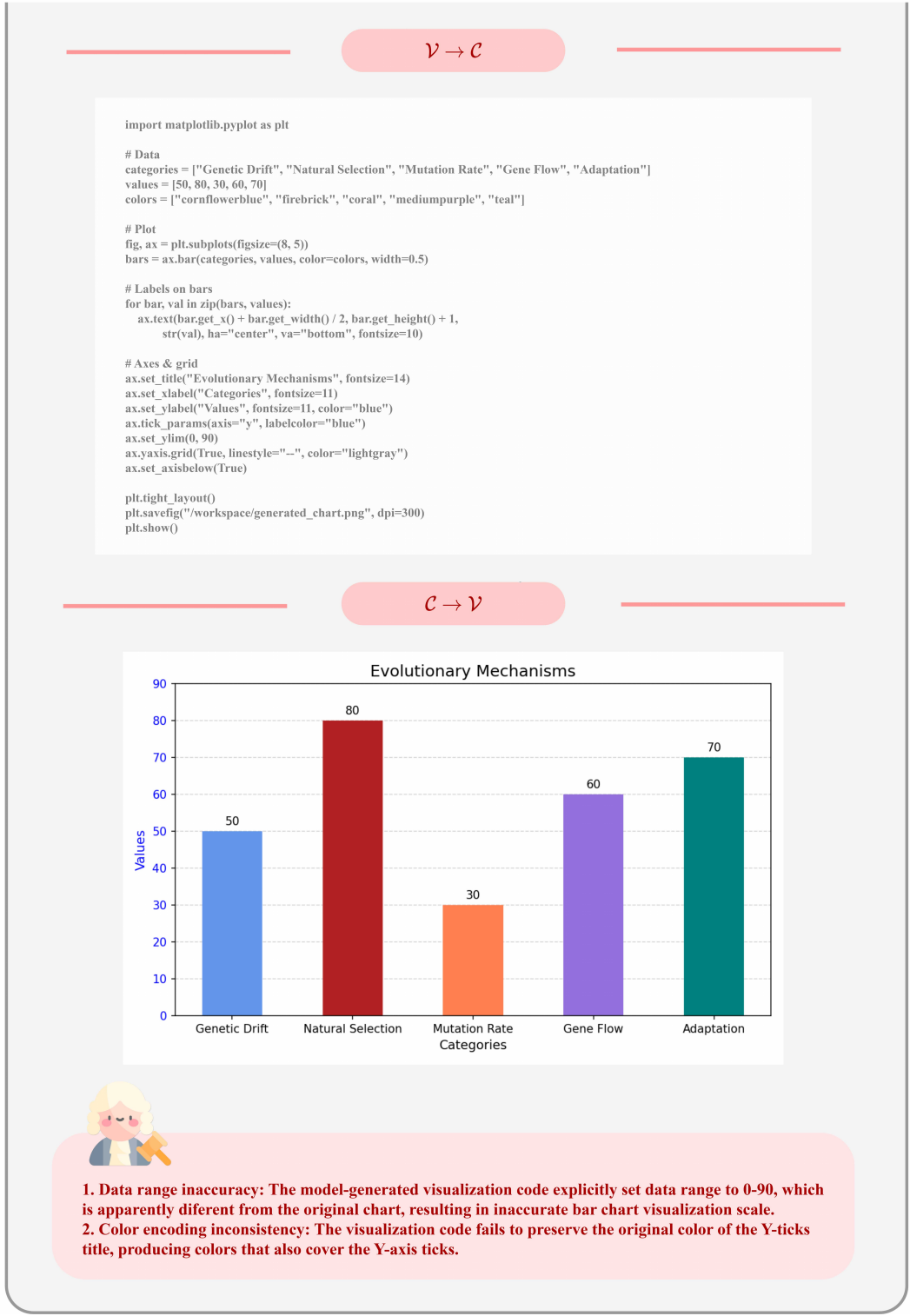}
    
    \caption{\textbf{Cross-Representation Understanding Failure Example of \texttt{Claude-4.6-Sonnet}.} A case study on \texttt{Claude-4.6-Sonnet} cross-representation understanding failures. This example is a simpler bar chart, with value labels to support accurate data recognition and extraction. Nevertheless, small errors still exist even for larger-size models, such as inaccurate data range and color encoding, resulting in data visualization inaccuracies and cross-representation inconsistencies. Constraints are omitted for clarity.}
    \label{fig:failure2_claude}

    \vspace{0pt}

\end{figure*}

To qualitatively analyze cross-representation failures of different models, we manually inspect model outputs and evaluate how each model performs. Overall, larger models showcase stronger cross-representation understanding capabilities than smaller-scale models. Through detailed examination, Fig.~\ref{fig:failure2_claude} illustrates a relatively simple case in which chart elements are explicitly labeled, mitigating the negative impact of inaccurate data extraction. However, \texttt{Claude-4.6-Sonnet} still exhibits style-related errors, failing to preserve the correct value ranges and color encodings. On the other hand, Fig.~\ref{fig:failure1_qwen3} shows the cross-representation outputs of \texttt{Qwen3-4B-VL}, revealing that its primary failures stem from data recognition and extraction errors. These inaccuracies also exist across representations and modalities, leading to inconsistent outputs. Its failures in color encoding and cross-representation consistency further highlight the importance of systematic evaluation signals for both accurately assessing cross-representation understanding and guiding model learning and optimization.

\clearpage

\begin{table}[t]
\centering
\small
\setlength{\tabcolsep}{4pt}

\begin{tabularx}{\textwidth}{lX}
\toprule
\textbf{Notation} & \textbf{Definition} \\
\midrule

\rowcolor{STUDENT_BG}
\multicolumn{2}{c}{\textit{Representation Spaces \& Data}} \\
\noalign{\vspace{2pt}}

$\mathcal{V}, \mathcal{T}, \mathcal{C}, \mathcal{S}$ & Spaces of chart images, tabular data, visualization code, and descriptive constraints \\
$v, t, c, s$ & A chart image, table, visualization code, and constraint \\
$\hat{v}, \hat{t}, \hat{c}, \hat{s}$ & Model-predicted chart image, table, code, and constraint \\
$s_{\textit{gt}}, t_{\textit{gt}}$ & Ground-truth constraint and table annotations \\[2pt]

\rowcolor{STUDENT_BG}
\multicolumn{2}{c}{\textit{Models, Policies \& Task Mappings}} \\
\noalign{\vspace{2pt}}

$M_{\theta}, M_{\psi}$ & Perform $\mathcal{V}\rightarrow\mathcal{C}$ and $\mathcal{T}\rightarrow\mathcal{C}$, respectively, with constraints \\
$\pi_{\theta}, \pi_{\psi}$ & Policies parameterizing $M_\theta$ and $M_\psi$ during RL training \\
$\pi_{\theta}^{\textit{ref}}, \pi_{\psi}^{\textit{ref}}$ & Frozen reference policies for KL regularization \\
$f_{\theta}$, $g_{\psi}$, $h$ & Task mappings instantiated by $M_\theta$, $M_\psi$, and sandbox executer, respectively  \\[2pt]

\rowcolor{STUDENT_BG}
\multicolumn{2}{c}{\textit{Teacher Module \& Training Configuration}} \\
\noalign{\vspace{2pt}}

$\mathcal{M}$ & Optional teacher module providing ground-truth conditioning context \\
$\alpha, \alpha_{\textit{start}}, \alpha_{\textit{end}}$ & Teacher-guidance fraction and its initial/final values under linear annealing \\
$\phi(\cdot)$ & Sentence embedding function via SentenceBERT \\
$N_{\textit{train}}, N_{\textit{epoch}}, \textit{lr}$ & Total training steps, epochs, and initial learning rate for \texttt{AdamW} \\[2pt]

\rowcolor{STUDENT_BG}
\multicolumn{2}{c}{\textit{Rollouts \& Sampling}} \\
\noalign{\vspace{2pt}}

$o_{\theta}^{(i)}, o_{\psi}^{(i,k)}$ & $i$-th rollout $(\hat{s}^{(i)}, \hat{t}^{(i)}, \hat{c}_{\theta}^{(i)})$ from $\pi_\theta$; $k$-th rollout from $\pi_\psi$ conditioned on $o_\theta^{(i)}$ \\
$K_{\theta}, K_{\psi}$ & Number of rollouts sampled per chart from $\pi_\theta$; per $\pi_\theta$ rollout from $\pi_\psi$ \\
$B$ & Global batch size \\
$\tau$ & Token position index within a rollout sequence \\
$\rho_{\theta,\tau}^{(i)}, \rho_{\psi,\tau}^{(i,k)}$ & Per-token importance ratios for $\pi_\theta$ and $\pi_\psi$ at token position $\tau$ \\
$\epsilon, \beta, \delta$ & Clipping threshold, KL penalty coefficient, and numerical stabilizer \\[2pt]

\rowcolor{STUDENT_BG}
\multicolumn{2}{c}{\textit{Rewards \& Advantage Estimation}} \\
\noalign{\vspace{2pt}}

$\mathbf{F}_c, \mathbf{F}_v, \mathbf{F}_f$ & Code consistency, visual consistency, and format reward components \\
$\mathbf{F}_s, \mathbf{F}_t$ & Constraint and table grounding rewards (teacher mode only) \\
$\texttt{sim}_{\textit{c}}(c_a, c_b)$ & Normalized code embedding cosine similarity between $c_a$ and $c_b$ \\
$e_a, e_b, e_{\textit{base}}$ & Code embeddings of $c_a$, $c_b$, and the baseline anchor in $\texttt{sim}_{\textit{c}}$ \\
$e_{\textit{clip}}(\cdot), e_{\textit{dino}}(\cdot)$ & Visual embedding functions of CLIP and DINOv2 \\
$R_{\theta}, R_{\psi}, R_{\theta}^{+}$ & Hierarchical rewards for $\pi_\theta$, $\pi_\psi$, and $\pi_\theta$ under teacher-guided mode \\
$r_\theta, r_\psi$ & Paired reward functionals depending on both models' outputs \\
$\bar{r}_{\theta}^{(i)}$ & $\pi_\theta$ reward marginalized over $K_\psi$ children rollouts \\
$A_{\theta}^{(i)}, A_{\psi}^{(i,k)}$ & Normalized advantages for $\pi_\theta$ rollout $i$ and $\pi_\psi$ rollout $(i,k)$ \\
$\mu_\theta, \sigma_\theta, \mu_\psi^{(i)}, \sigma_\psi^{(i)}$ & Mean and std for $\pi_\theta$ and $\pi_\psi$ advantage normalization \\[2pt]

\rowcolor{STUDENT_BG}
\multicolumn{2}{c}{\textit{Rule-as-Judge Evaluation Metrics}} \\
\noalign{\vspace{2pt}}

$F_{\textit{clip}}, F_{\textit{ssim}}, F_{\textit{ocr}}, F_{\textit{dino}}$ & Chart evaluation: semantic, structural, textual, and perceptual accuracy \\
$F_{\textit{exec}}$ & Code executability: binary sandbox execution success \\
$F_{\textit{codebleu}}, F_{\textit{ast}}, F_{\textit{cosine}}, F_{\textit{codebert}}, F_{\textit{unixcoder}}$ & Code evaluation: quality, structural, lexical, contextual, and semantic accuracy \\
$F_{\textit{ngram}}, F_{\textit{wngram}}, F_{\textit{syntax}}, F_{\textit{dataflow}}$ & Four sub-dimensions of CodeBLEU comprising $F_{\textit{codebleu}}$ \\
$F_{\textit{schema}}, F_{\textit{value}}$ & Table evaluation: column-level schema F1 and cell-level value F1 \\
$F_{\textit{ssem}}, F_{\textit{rouge}}, F_{\textit{LCS}}$ & Constraint evaluation: semantic accuracy, ROUGE-L, and LCS F1 \\
$F_{\textit{chart}}$, $F_{\textit{code}}$, $F_{\textit{table}}$, $F_{\textit{constraint}}$ & Final rule-as-judge score for chart, code, table, and constraint evaluation \\[2pt]

\rowcolor{STUDENT_BG}
\multicolumn{2}{c}{\textit{LLM-as-Judge Evaluation Metrics (prefix $C$: code)}} \\
\noalign{\vspace{2pt}}

$J_{\textit{Cdata}}, J_{\textit{Ctype}}, J_{\textit{Cstruct}}, J_{\textit{Cvisual}}, J_{\textit{Cstyle}}$ & LLM-as-judge code dimensions: data correctness, chart type, structural fidelity, visual and style accuracy \\
$J_{\textit{code}}$ & Final LLM-as-judge code evaluation score \\[2pt]

\bottomrule

\end{tabularx}

\vspace{12pt}

\end{table}

\begin{table}[t]
\centering
\small
\setlength{\tabcolsep}{4pt}

\vspace{-16pt}

\begin{tabularx}{\textwidth}{lX}
\toprule
\textbf{Notation} & \textbf{Definition} \\
\midrule

\rowcolor{STUDENT_BG}
\multicolumn{2}{c}{\textit{MLLM-as-Judge Evaluation Metrics (prefixes $V$: chart, $T$: table, $S$: constraint)}} \\
\noalign{\vspace{2pt}}

$J_{\textit{Vtype}}, J_{\textit{Vdata}}, J_{\textit{Vtext}}, J_{\textit{Vstyle}}, J_{\textit{Vvisual}}$ & MLLM-as-judge chart dimensions: type, data, text, style, and visual accuracy \\
$J_{\textit{Tschema}}, J_{\textit{Tdata}}, J_{\textit{Tcoverage}}, J_{\textit{Trow}}, J_{\textit{Tformat}}$ & MLLM-as-judge table dimensions: schema, data, coverage, row alignment, and format correctness \\
$J_{\textit{Sfidelity}}, J_{\textit{Scompleteness}}, J_{\textit{Sclarity}}, J_{\textit{Strend}}, J_{\textit{Sstyle}}$ & MLLM-as-judge constraint dimensions: fidelity, completeness, clarity, trend accuracy, and style \\
$J_{\textit{chart}}$, $J_{\textit{table}}$, $J_{\textit{constraint}}$ & Final MLLM-as-judge score for chart, table, and constraint evaluation \\[2pt]

\rowcolor{STUDENT_BG}
\multicolumn{2}{c}{\textit{Weighting Coefficient Families}} \\
\noalign{\vspace{2pt}}

$\omega_*^{(\pi)}$ & Sub-weights within a reward component \\
$\lambda_*^{(\pi)}$ & Weighting coefficients combining reward components during training \\
$\gamma_*$ & Weighting coefficients combining metrics during evaluation \\[2pt]

\rowcolor{STUDENT_BG}
\multicolumn{2}{c}{\textit{Dataset \& Filtering Thresholds}} \\
\noalign{\vspace{2pt}}

$l_{\textit{input}}, l_{\textit{code}}, l_{\textit{table}}, l_{\textit{constraint}}, l_{\textit{timeout}}$ & Token thresholds for input, code, table, and constraint; sandbox execution timeout \\

\bottomrule
\end{tabularx}

\caption{\textbf{Notation and Symbol Reference.} We summarize core symbols and notations used in this paper, grouped by functional role. The three weighting coefficient families are intentionally distinct: $\omega_*$ denotes sub-weights within a reward component, $\lambda_*$ denotes weights combining reward components during training, and $\gamma_*$ denotes weights combining metrics during evaluation. LLM-as-judge and MLLM-as-judge dimension scores use prefixes $C$, $V$, $T$, $S$ to disambiguate dimensions sharing the same name across evaluators.}

\vspace{16pt}

\label{tab:notation}
\end{table}

\begin{table*}[t]
\centering
\small
\renewcommand{\arraystretch}{1.4}
\begin{tabularx}{\textwidth}{p{0.35\textwidth}>{\centering\arraybackslash}p{0.16\textwidth}>{\centering\arraybackslash}p{0.2\textwidth}>{\centering\arraybackslash}p{0.16\textwidth}}
\toprule
\textbf{Method} & \textbf{GPU} & \textbf{API Cost} & \textbf{Time} \\
\midrule

\rowcolor{STUDENT_BG}
\multicolumn{4}{c}{\textit{Training}} \\
\noalign{\vspace{3pt}}
Baseline & 4$\times$ H100 96GB & -- & 86h / run \\
\ours~($\mathcal{M}$ disabled) & 4$\times$ H100 96GB & -- & 77h / run \\
\ours~($\mathcal{M}$ enabled, no annealing) & 4$\times$ H100 96GB & -- & 71h / run \\
\ours~($\mathcal{M}$ enabled, with annealing) & 4$\times$ H100 96GB & -- & 72h / run \\

\noalign{\vspace{3pt}}
\rowcolor{STUDENT_BG}
\multicolumn{4}{c}{\textit{Evaluation}} \\
\noalign{\vspace{3pt}}
Baseline & 4$\times$ H100 96GB & \$0.8637 / sample & 128.72s / sample \\
\ours & 4$\times$ H100 96GB & \$0.5674 / sample & 77.63s / sample \\
\ours~(w/ \ourstest) & 4$\times$ H100 96GB & \$0.2382 / sample & 71.88s / sample \\

\noalign{\vspace{1pt}}
\bottomrule
\end{tabularx}
\renewcommand{\arraystretch}{1.0}
\caption{\textbf{Computation Overhead.} We summarize GPU resources, API cost, and wall-clock time for main training configurations and evaluation. API cost is calculated for per sample averaged across all evaluation benchmarks and decreases with \ourstest. Wall-clock time for training is calculated for per full training run; for evaluation it is calculated for per sample averaged across all benchmarks. ``--'' indicates no API cost is incurred during training.}
\label{tab:computation_overhead}
\end{table*}


\begin{table*}[t]
\centering
\small
\setlength{\tabcolsep}{4pt}
\begin{tabularx}{\textwidth}{>{\centering\arraybackslash}X>{\centering\arraybackslash}p{5.0cm}>{\centering\arraybackslash}p{5.0cm}}

\toprule
\textbf{Parameter} & \textbf{$M_\theta$} & \textbf{$M_\psi$} \\
\midrule

\rowcolor{STUDENT_BG}
\multicolumn{3}{c}{\textit{Dataset Construction}} \\
\noalign{\vspace{2pt}}

$l_{\textit{input}}$ & 8,192 & 8,192 \\
$l_{\textit{code}}$ & 8,192 & 8,192 \\
$l_{\textit{table}}$ & 8,192 & 8,192 \\
$l_{\textit{constraint}}$ & 8,192 & 8,192 \\
$l_{\textit{timeout}}$ & 60 & 60 \\[2pt]

\rowcolor{STUDENT_BG}
\multicolumn{3}{c}{\textit{Main Training ($\mathcal{M}$ Disabled)}} \\
\noalign{\vspace{2pt}}

$N_{\textit{epoch}}$ & 2 & 2 \\
$N_{\textit{train}}$ & 200 & 200 \\
$B$ & 8 & 8 \\
$K_{\theta}$, $K_{\psi}$ & 4, 4 & 4, 4 \\
$\textit{lr}$ & $1\times10^{-6}$ & $1\times10^{-6}$ \\
$\alpha$ & 0 & 0 \\
$\lambda_{f}^{(\pi)}$ & 0.1 & 0.1 \\
$\lambda_{c}^{(\pi)}$ & 0.05 & 0.2 \\
$\lambda_{\pi}$ & 0.7 & 0.7 \\
$\lambda_{v}^{(\pi)}$ & 0.05 & 0.2 \\
$\omega_{e}^{(\pi)}$ & 0.9 & 0.7 \\
$\omega_{s}^{(\pi)}$ & 0.1 & 0.3 \\
$\omega_{\textit{clip}}^{(\pi)}$ & 0.25 & 0.25 \\
$\omega_{\textit{ssim}}^{(\pi)}$ & 0.25 & 0.25 \\
$\omega_{\textit{ocr}}^{(\pi)}$ & 0.25 & 0.25 \\
$\omega_{\textit{dino}}^{(\pi)}$ & 0.25 & 0.25 \\[2pt]

\rowcolor{STUDENT_BG}
\multicolumn{3}{c}{\textit{Ablation Study: Teacher Module ($\mathcal{M}$ Enabled, No Annealing)}} \\
\noalign{\vspace{2pt}}

$\alpha_{\textit{start}}$ & 1.0 & 1.0 \\
$\alpha_{\textit{end}}$ & 1.0 & 1.0 \\
$\lambda_{s}^{(\theta)}$ & 0.2 & -- \\
$\lambda_{t}^{(\theta)}$ & 0.3 & -- \\
$\omega_{\textit{schema}}^{(\theta)}$ & 0.5 & -- \\
$\omega_{\textit{value}}^{(\theta)}$ & 0.5 & -- \\[2pt]

\rowcolor{STUDENT_BG}
\multicolumn{3}{c}{\textit{Ablation Study: Teacher Module ($\mathcal{M}$ Enabled, With Annealing)}} \\
\noalign{\vspace{2pt}}

$\alpha_{\textit{start}}$ & 1.0 & 1.0 \\
$\alpha_{\textit{end}}$ & 0.0 & 0.0 \\[2pt]

\rowcolor{STUDENT_BG}
\multicolumn{3}{c}{\textit{Ablation Study: Consistency Reward Weighting}} \\
\noalign{\vspace{2pt}}

$\lambda_{c}^{(\pi)}$ & 0.2 & 0.2 \\
$\lambda_{v}^{(\pi)}$ & 0.2 & 0.2 \\
$\lambda_{\pi}$ & 0.5 & 0.5 \\
$\omega_{e}^{(\pi)}$ & 0.5 & 0.5 \\
$\omega_{s}^{(\pi)}$ & 0.5 & 0.5 \\[2pt]

\rowcolor{STUDENT_BG}
\multicolumn{3}{c}{\textit{Evaluation}} \\
\noalign{\vspace{2pt}}

$\gamma_{*}$ in $F_{\textit{chart}}$ & 0.25 & 0.25 \\
$\gamma_{*}$ in $F_{\textit{code}}$ & 0.20 & 0.20 \\
$\gamma_{*}$ in $F_{\textit{bleu}}$ & 0.25 & 0.25 \\
$\gamma_{*}$ in $F_{\textit{table}}$ & 0.50 & 0.50 \\
$\gamma_{*}$ in $F_{\textit{constraint}}$ & 0.50 & 0.50 \\
$\gamma_{*}$ in $J_{\textit{code}}$ & 0.20 & 0.20 \\
$\gamma_{*}$ in $J_{\textit{chart}}$ & 0.20 & 0.20 \\
$\gamma_{*}$ in $J_{\textit{table}}$ & 0.20 & 0.20 \\
$\gamma_{*}$ in $J_{\textit{constraint}}$ & 0.20 & 0.20 \\

\bottomrule
\end{tabularx}

\caption{\textbf{Experiment Configuration.} We summarize our core hyperparameter settings across different experimental stages. Parameters specific to teacher mode ($\lambda_s^{(\theta)}, \lambda_t^{(\theta)}, \omega_{\textit{schema}}^{(\theta)}, \omega_{\textit{value}}^{(\theta)}$) apply to $M_\theta$ only and are marked ``--'' for $M_\psi$. Ablation study rows list only parameters that differ from the corresponding main training setting. In our evaluation, we set the evaluation weighting coefficients $\gamma_*$ to be uniform within each score.}
\label{tab:configuration}

\end{table*}

\clearpage


\begin{figure}[!t]
    \vspace{-18pt}

    \small
    \centering
    \includegraphics[width=1.0\textwidth]{assets/eval_prompt_chart_mllm_as_judge.pdf}
    
    \vspace{-6pt}
    
    \caption{\textbf{MLLM-as-Judge for Chart Evaluation.} Our MLLM-as-judge for chart evaluation assesses predicted chart against ground-truth chart across five fine-grained dimensions, including chart type accuracy, data accuracy, text accuracy, style accuracy, and visual accuracy (\S\ref{subsubsec:appendix:ours_eval_suite:mllm_judge:chart}). Each dimension is scored on a $0$--$5$ integer scale with detailed instructions from reasoning to scoring, addressing the key limitations in existing MLLM-as-judge approaches (\S\ref{subsec:method:evaluation_metrics}).}
    \label{fig:eval_prompt:chart_mllm_as_judge}

    \vspace{0pt}

\end{figure}


\begin{figure*}[!t]
    \small
    \centering
    \includegraphics[width=1.0\textwidth]{assets/eval_prompt_code_llm_as_judge_1.pdf}

\end{figure*}

\begin{figure*}[!t]
    \vspace{-28pt}

    \small
    \centering
    \includegraphics[width=1.0\textwidth]{assets/eval_prompt_code_llm_as_judge_2.pdf}
    
    \vspace{16pt}
    
    \caption{\textbf{LLM-as-Judge for Code Evaluation.} Our LLM-as-judge evaluates predicted visualization code against ground-truth code across five fine-grained dimensions, including data correctness, chart type accuracy, structural fidelity, visual accuracy, and style accuracy (\S\ref{subsec:appendix:ours_eval_suite:llm_judge}). Each dimension is scored on a $0$--$5$ integer scale with detailed instructions from reasoning to scoring, addressing the key limitations in existing LLM-as-judge approaches (\S\ref{subsec:method:evaluation_metrics}).}
    \label{fig:eval_prompt:code_llm_as_judge}

    \vspace{0pt}

\end{figure*}


\begin{figure}[!t]
    \vspace{-12pt}

    \small
    \centering
    \includegraphics[width=1.0\textwidth]{assets/eval_prompt_table_mllm_as_judge.pdf}

    \vspace{-6pt}
    
    \caption{\textbf{MLLM-as-Judge for Table Evaluation.} Our MLLM-as-judge for table evaluation assesses the extracted table against the reference table across five fine-grained dimensions, including schema accuracy, data accuracy, data coverage, row alignment, and format correctness (\S\ref{subsubsec:appendix:ours_eval_suite:mllm_judge:table}). Each dimension is scored on a $0$--$5$ integer scale with detailed instructions from reasoning to scoring, addressing the key limitations in existing MLLM-as-judge approaches (\S\ref{subsec:method:evaluation_metrics}).}
    \label{fig:eval_prompt:table_mllm_as_judge}

    \vspace{0pt}

\end{figure}


\begin{figure}[!t]
    \vspace{-12pt}

    \small
    \centering
    \includegraphics[width=1.0\textwidth]{assets/eval_prompt_constraint_mllm_as_judge.pdf}
    
    \vspace{0pt}
    
    \caption{\textbf{MLLM-as-Judge for Constraint Evaluation.} Our MLLM-as-judge for constraint evaluation assesses the predicted constraint against the reference constraint across five fine-grained dimensions, including visual fidelity, visual completeness, structural clarity, trend \& relation accuracy, and style accuracy (\S\ref{subsubsec:appendix:ours_eval_suite:mllm_judge:constraint}). Each dimension is scored on a $0$--$5$ integer scale with detailed instructions from reasoning to scoring, addressing the key limitations in existing MLLM-as-judge approaches (\S\ref{subsec:method:evaluation_metrics}).}
    \label{fig:eval_prompt:constraint_mllm_as_judge}

    \vspace{0pt}

\end{figure}


\begin{figure*}[!t]
    \vspace{0pt}

    \small
    \centering
    \includegraphics[width=1.0\textwidth]{assets/case_study-failure1-qwen3-4b-1.pdf}

\end{figure*}

\begin{figure*}[!t]
    \vspace{-24pt}

    \small
    \centering
    \includegraphics[width=1.0\textwidth]{assets/case_study-failure1-qwen3-4b-2.pdf}

\end{figure*}

\begin{figure*}[!t]
    \vspace{0pt}

    \small
    \centering
    \includegraphics[width=1.0\textwidth]{assets/case_study-failure2-claude-1.pdf}

\end{figure*}

\end{document}